\documentclass[nonblindrev]{informs3noheader}

\OneAndAHalfSpacedXI

\usepackage{mathtools}
\usepackage[inline]{enumitem}
\usepackage{subcaption}
\usepackage{booktabs}
\usepackage{array}
\usepackage{multirow}
\usepackage{booktabs,caption,fixltx2e}
\usepackage[flushleft]{threeparttable}
\usepackage{lscape}
\usepackage{makecell}
\usepackage{amssymb}
\usepackage{amsmath}
\usepackage[mathscr]{euscript}
\usepackage{algorithm}
\usepackage{algpseudocode}
\usepackage{pifont}
\usepackage{caption}
\usepackage{algorithmicx}
\usepackage{hyperref}
\usepackage{mathrsfs}
\usepackage{arydshln}
\usepackage{tabularx}
\usepackage{adjustbox}
\usepackage{bm, bbm}
\usepackage{soul}
\usepackage{blindtext}
\usepackage{longtable}
\hypersetup{
colorlinks=true,
linkcolor=black,
citecolor=black,
filecolor=black,
urlcolor=black,
}
\makeatletter
\newcommand{\customlabel}[2]{%
   \protected@write \@auxout {}{\string \newlabel {#1}{{#2}{\thepage}{#2}{#1}{}} }%
   \hypertarget{#1}{#2}
}
\makeatother

\usepackage{comment}

\usepackage{tikz}
\usetikzlibrary{shapes.geometric,shapes.multipart}
\usetikzlibrary{positioning}
\usetikzlibrary{decorations.pathreplacing}
\definecolor{mygreen}    {RGB}{0,90,0}
\definecolor{myblue}     {RGB}{0,51,140}
\definecolor{myorange}   {RGB}{238,118,0}
\definecolor{myred}      {RGB}{126,0,0}
\definecolor{mygray}     {RGB}{100,100,105}
\definecolor{mygrayblue} {RGB}{0,128,128}
\definecolor{mygraygreen}{RGB}{128,128,0}
\definecolor{DarkPurple}     {RGB}{142, 36, 170}
\definecolor{LightPurple}    {RGB}{57, 130, 7}

\usepackage{natbib}
\bibpunct[, ]{(}{)}{,}{a}{}{,}%
\def\bibfont{\small}%

\usepackage[utf8]{inputenc}
\usepackage[english]{babel}
\usepackage{graphicx}

\newcommand{\Z}{\mathcal{Z}}



\newcolumntype{F}[1]{%
    >{\raggedright\arraybackslash\hspace{0pt}}p{#1}}%
\newcolumntype{T}[1]{%
    >{\centering\arraybackslash\hspace{0pt}}p{#1}}%
\DeclareMathOperator{\Bern}{Bern}
\DeclareMathOperator{\Var}{Var}

\TheoremsNumberedThrough
\EquationsNumberedThrough

\renewcommand{\d}{\mathrm{d}}
\DeclareMathOperator\Bino{Bino}
\DeclareMathOperator\Ber{Ber}

\def\R{\mathbb{R}}
\def\Z{\mathbb{Z}}

\def\E{\mathbb{E}}
\def\P{\mathbb{P}}

\def\calA{\mathcal{A}}

\def\calC{\mathcal{C}}
\def\calD{\mathcal{D}}
\def\calE{\mathcal{E}}
\def\calF{\mathcal{F}}
\def\calG{\mathcal{G}}
\def\calH{\mathcal{H}}

\def\calJ{\mathcal{J}}

\def\calL{\mathcal{L}}
\def\calM{\mathcal{M}}
\def\calN{\mathcal{N}}
\def\calO{\mathcal{O}}
\def\calP{\mathcal{P}}
\def\calQ{\mathcal{Q}}

\def\calS{\mathcal{S}}
\def\calT{\mathcal{T}}

\def\calX{\mathcal{X}}

\def\bA{\boldsymbol{A}}
\def\bB{\boldsymbol{B}}

\def\bi{\boldsymbol{i}}

\def\bs{\boldsymbol{s}}

\def\bx{\boldsymbol{x}}
\def\by{\boldsymbol{y}}

\def\bone{\boldsymbol{1}}

\def\bSigma{\boldsymbol{\Sigma}}

\def\st{\text{s.t.}}

\definecolor{mygreen}    {RGB}{0,90,0}
\definecolor{myblue}     {RGB}{0,51,140}
\definecolor{myorange}   {RGB}{238,118,0}
\definecolor{myred}      {RGB}{126,0,0}
\definecolor{mygray}     {RGB}{100,100,105}
\definecolor{mygrayblue} {RGB}{0,128,128}
\definecolor{mygraygreen}{RGB}{128,128,0}
\definecolor{DarkPurple}     {RGB}{142, 36, 170}
\definecolor{LightPurple}    {RGB}{57, 130, 7}

\begin{document}

%\RUNAUTHOR{Jacquillat and Li}
\RUNAUTHOR{}

\RUNTITLE{Learning to cover}

\TITLE{Learning to cover: online learning and optimization with irreversible decisions}

\ARTICLEAUTHORS{%
\AUTHOR{Alexandre Jacquillat and Michael Lingzhi Li}
\AFF{
Sloan School of Management, Massachusetts Institute of Technology, Cambridge, MA 02142, \EMAIL{alexjacq@mit.edu}\\
Technology and Operations Management, Harvard Business School, Cambridge, MA 02163, \EMAIL{mili@hbs.edu}}
}

\ABSTRACT{
We define an online learning and optimization problem with discrete and irreversible decisions contributing toward a coverage target. In each period, a decision-maker selects facilities to open, receives immediate feedback on their success, and updates a classification model to guide future decisions. The goal is to minimize facility openings under a chance constraint reflecting the coverage target, in an asymptotic regime with a large target number of facilities $m\to\infty$ but a finite horizon $T\in\Z_+$. We prove that, under statistical conditions, the online classifier converges to the Bayes-optimal classifier at a rate of at best $\calO(1/\sqrt n)$. We use this result to formulate the online learning and optimization problem, with an error decay rate $r>0$ and a residual error $1-p$. We derive asymptotically tight regret bounds against a fully-learned benchmark, in $\Theta\left(m^{\frac{1-r}{1-r^T}}\right)$ if $p=1$ (perfect learning) or $\Theta\left(\max\left\{m^{\frac{1-r}{1-r^T}},\sqrt{m}\right\}\right)$ otherwise. In particular, the regret rate is sub-linear and converges exponentially fast to its infinite-horizon limit. We establish the robustness of this result in a dynamic environment and extend it to a more complicated bipartite facility-customer graph with a target on customer coverage. Throughout, constructive proofs identify asymptotically optimal algorithms featuring limited exploration initially and fast exploitation later on. These results uncover the benefits of even limited online learning and optimization through pilot programs prior to full-fledged expansion.}


\maketitle

\vspace{-12pt}
\section{Introduction}

Organizations often face challenging strategic decisions when determining where to open facilities, which programs to launch, or which technologies to invest in. These decisions need to be made early on under uncertainty about which options will succeed in practice. When ample historical data exist, predictive models can be trained offline to guide these planning decisions. In many important contexts, however, comparable data are scarce—especially when decisions involve new geographies, new technologies, or new operational processes. In those settings, decision makers must learn from early deployments about successful vs. unsuccessful outcomes to guide subsequent choices.

This setting gives rise to an online learning and optimization paradigm, in which each decision both commits resources and reveals information that can improve later decisions. Early deployments provide information for learning purposes, but also risk costly missteps due to the lack of accurate predictions on the success of each investment. Vice versa, delaying decisions until more information is available can preserve early resources but also hinders data acquisition. This trade-off is most acute when organizations must make many decisions over a few rounds of iteration to quickly achieve broad coverage of a population, market, or region, and when each action involves substantial, irreversible investment. This paper studies this class of large-scale, finite-horizon deployment problems with intertwined learning and optimization, in which feedback from early outcomes can guide subsequent decisions. The following examples illustrate this structure.

\begin{example}[Clinical trials]
In late-stage randomized clinical trials, multiple sites are responsible for recruiting local patients to evaluate the efficacy of a treatment and monitor side effects. Because of the increasing complexity of these trials, pharmaceutical companies routinely open 70-100 sites per trial globally \citep{brogger2022changes}. Each one involves substantial and irreversible investments to coordinate with public health agencies, build infrastructure, and recruit staff. Yet, trials sites face uncertainty: about 80\% fail to meet recruitment timelines \citep{johnson2015evidence}, and up to 20\% are abandoned due to inadequate enrollment \citep{carlisle2015unsuccessful}. Site activation teams monitor enrollment from existing sites to guide subsequent rounds of site openings. The objective is to meet recruitment targets in a limited number of activation rounds.
\end{example}
\begin{example}[Vaccination planning]
To execute a mass vaccination campaign, public health authorities deploy a large number of vaccination clinics within a short timeframe. During the COVID-19 pandemic, for example, hundreds of vaccination centers were established nationwide to ensure broad and equitable population coverage \citep{bertsimas2022locate,jacquillat2024branch}. Opening and operating each center requires major financial and logistical commitments. However, not all sites succeed due to demand, operational or administrative bottlenecks. Planners therefore learn from initial deployments which types of sites perform well and adapt subsequent waves of openings to achieve broad population coverage within a few deployment rounds.
\end{example}
\begin{example}[Humanitarian logistics]
Following natural disasters, humanitarian organizations must rapidly establish temporary centers to provide food, water, medical supplies and shelter to displaced populations. Each site opening requires substantial irreversible commitments to secure supplies and mobilize personnel. Yet, many facilities fail to operate effectively due to logistical, coordination, and infrastructure difficulties. In practice, agencies deploy initial sites quickly, monitor their performance in the field, and use these observations to refine where and how to open subsequent facilities. The overarching objective is to reach comprehensive population coverage across affected regions over a small number of deployment waves|within days or weeks.
\end{example}
\begin{example}[Technology investing]
Venture funds or internal corporate venture programs often manage large portfolios of projects, each requiring substantial upfront capital commitment. Yet, many of ventures fail to produce viable products or gain sustainable market traction \citep{challapally2025genai}. Fund partners closely monitor the performance of early investments to refine their subsequent investing strategy, for instance by concentrating capital in sectors, technologies, or business models that show early promise. The challenge is to meet a target number of successful ``wins'' in the portfolio within a few funding stages, while containing the cost of failed investments.
\end{example}
Although these problems arise in distinct industries, they share a common structure: We need to make a large number of discrete, costly, and irreversible deployment decisions over a few periods, by leveraging observations on successful and unsuccessful outcomes from previous stages. We refer to this problem as \emph{learning to cover}. Throughout, we adopt a facility selection nomenclature and refer to the decisions as ``facility openings'', although the results equally apply to other settings (e.g., investments). This paper optimizes these decisions to meet a coverage target at minimal cost, by balancing the need to improve future predictions by generating more data with the success of facilities in the current period. Theoretically, the problem features the following elements:
\begin{itemize}
    \item[--] Online learning and optimization under endogenous uncertainty with immediate feedback. A decision-maker observes the success or failure of each selected facility or project prior to the next round of decisions. The reliance on online data arises in the case of new technologies or processes (e.g., clinical trials for a new drug, clinics for the COVID-19 vaccine in 2021, investment in AI projects, etc.). This creates a trade-off between learning-based exploration (acquiring more data to train a machine learning model and mitigate the prediction error) and optimization-based exploitation (postponing decisions until the error gets smaller).
    \item[--] Target coverage. The goal is to achieve a coverage level within a finite horizon. We formalize a chance constraint imposing a probabilistic target on the number of successes (e.g., clinical trials sites, technology ventures), or on population coverage (e.g., shelters, vaccination centers).
    \item[--] Discrete, irreversible decisions. Early facility deployment decisions incur high, irreversible costs. This commitment creates inter-temporal dependencies in the system's physical state.
    \item[--] A new asymptotic regime. In practice, each round can take days (e.g., humanitarian logistics) up to months (e.g., clinical trials), but target coverage may need to be achieved quickly, within a few rounds of iterations. We define an asymptotic regime with a finite planning horizon (e.g., 2 to 5 periods) but a large coverage target---either a large supply-side target on the number of successful facilities or demand-side target on the number of served customers.
\end{itemize}

\subsubsection*{Contributions.}

This paper develops an end-to-end approach to \textit{online learning and optimization with discrete and irreversible decisions toward target coverage}. We first establish the convergence of the online classifier in this environment to the Bayes-optimal classifier, which we use to formulate the online learning and optimization problem. We then derive tight upper and lower bounds on the optimal regret in this problem against a fully-learned benchmark, along with an interpretable policy that achieves asymptotic optimality up to the second leading order. We establish the robustness of these results in static vs. dynamic environments, and with supply-side vs. demand-side coverage targets. These results uncover a sub-linear regret rate, which highlights the benefits of even a few rounds of online learning and optimization in strategic planning under uncertainty, via pilot deployments for initial exploration followed by rapid exploitation. These contributions offer a unified theoretical and practical foundation for multi-stage planning under uncertainty, blending tools from statistical learning, online optimization, and decision theory.

Specifically, we formalize a finite-horizon optimization problem with online learning to open facilities over $T$ periods. Each facility is associated with a vector of covariates but its success is random. At each period, the decision-maker builds a machine learning model to predict facility success; attempts to open facilities; receives immediate feedback on their success of each one; and updates the machine learning model for the next time period. The objective is to minimize facility openings subject to the chance constraint reflecting target coverage by the end of the horizon.

We first characterize the statistical properties of the online classifier in this environment (Section~\ref{sec:setting}). Due to intertwined learning and optimization, the data sample is biased toward facilities that are likely to succeed, hence not independent and identically distributed (i.i.d.). Still, we prove that, under certain statistical conditions, the online classifier converges to the Bayes-optimal classifier at a rate of at best $\calO(1/\sqrt n)$, where $n$ is the number of accrued data points. We use this result to model, conservatively, the number of successful facilities at each iteration via a binomial distribution with a probability of success that decays with the size of the data sample based on a generalized learning rate $r>0$ and an irreducible error term $1-p$.

Our main result is that the optimal regret grows at a sub-linear rate, as compared to a fully-learned benchmark that has access to the Bayes-optimal classifier. When $p=1$ (perfect learning), the regret grows in $\Theta\left(m^{\frac{1-r}{1-r^T}}\right)$ if $r\neq1$ and in $\Theta\left(m^{1/T}\right)$ if $r=1$, where $m$ denotes a coverage target (on facilities or customers); when $p<1$ and $r\neq1$, the regret grows in $\Theta\left(\max\left\{m^{\frac{1-r}{1-r^T}},\sqrt{m}\right\}\right)$, where the $\sqrt{m}$ factor comes from the residual learning error. This expression provides an asymptotically tight regret as the coverage target $m$ grows but the planning horizon~$T$ remains finite. Through constructive proofs, we provide interpretable deterministic-approximation algorithms that yield asymptotically optimal regret without a priori knowledge of the machine learning model.

We first derive our result in a setting with a \textit{target on facilities}, in which a decision-maker aims to open $m$ facilities (Section~\ref{sec:base}--\ref{sec:implications}). We establish a lower bound from a deterministic approximation, using a leading-order analysis. We construct an algorithm that adds a buffer to the deterministic approximation to handle uncertainty. We prove that its solution is asymptotically feasible and optimal as $m$ grows to infinity, using concentration inequalities---Hoeffding's inequality, Bernstein's inequality and the Berry-Esseen theorem---across periods and across facilities. The algorithm gives rise to a policy where the number of facility openings is concave in $m$, which combines a limited-exploration strategy initially for learning purposes and an exploitation strategy later on by opening most facilities once uncertainty is mitigated. Empirical evidence using real-world data suggests the strong performance of this online learning and optimization algorithm even in finite samples.

We establish the robustness of the solution with offline data and against an adaptive re-optimization procedure. In particular, we prove that the static and adaptive solutions achieve the same asymptotic rates for the first $T-1$ periods. This result suggests limited benefits of adaptiveness in the online learning and optimization problem. We also propose a ``semi-adaptive'' policy that relies on the static solution for the first $T-1$ periods but implements a final-period adjustment to \textit{exactly} satisfy the chance constraint, without the buffer from our theoretical algorithm. Computational results show that this simple and interpretable algorithm yields high-quality solutions.

We extend our result to the networked setting with a \textit{target on customers}, where $m$ customers need to be served with connected facilities (Section~\ref{sec:network}). We assume that most customers can only be served by facilities of bounded degree to retain a similar asymptotic regime with many facilities. We show that the optimal regret still grows in $\Theta\left(m^{\frac{1-r}{1-r^T}}\right)$, $\Theta(m^{1/T})$ or $\Theta(\sqrt m)$ under perfect learning ($p=1$). The proof is further complicated by two factors. First, customer coverage depends on the set (vs. the number) of successful facilities; in turn, we use the concentration inequality in dependency graphs from \cite{janisch2024berry} to relate the stochastic solution to its deterministic approximation. Second, the fully-learned benchmark is no longer known due to the combinatorial complexity of facility location; in turn, we derive sensitivity analysis results to bound the difference between the fully-learned benchmark and the stochastic solution. Another complexity in the networked setting is that the deterministic-approximation algorithm relies on mixed-integer non-convex optimization; we solve it computationally with a procedure that opens facilities in increasing order of expected incremental coverage. Computational results extend our theoretical insights on learning-based exploration and optimization-based exploitation, and sub-linear regret.

These results underscore the benefits of \textit{even limited online learning and optimization}. The sub-linear regret rate achieves a middle ground between the fully-learned benchmark (which achieves zero regret by definition) and a no-learning baseline (which, we prove, achieves linear regret). Moreover, the regret rate converges exponentially to its infinite-horizon limit, thereby highlighting the benefits of even a few iterations of learning and optimization. These findings caution against both all-at-once decision-making and prolonged rounds of learning. Instead, the limited learning approach can take the form of a small pilot program to acquire online data prior to full-fledged expansion, which can be practical to implement and, as our results show, yield significant benefits.

\section{Literature Review}

Within the facility location literature, our problem of strategic planning under supply-side uncertainty relates to the maximum expected covering location problem \citep{daskin1983maximum} and the reliable facility location problem \citep{snyder2005reliability}. Such problems have been tackled via continuous approximations \citep{cui2010reliable}, stochastic programming \citep{laporte1994exact,shen2011reliable}, robust optimization \citep{cheng2021robust}, and distributionally robust optimization \citep{lu2015reliable}. \cite{lim2013facility} showed that facility location decisions are sensitive to errors in failure probabilities, motivating a machine learning approach or predict facility success.

Our problem features an online optimization structure with temporal interdependencies. This setting relates to Pandora's problem \citep{brown2024sequential}, online revenue management \citep{jasin2012re,bumpensanti2020re}, the online multi-secretary problem \citep{arlotto2019uniformly}, and online resource allocation \citep{vera2021bayesian,banerjee2019good,baxi2024online}. In these problems, inter-temporal coupling stems from an optimal stopping structure or a shared budget under exogenous demand uncertainty. In our setting, coupling arises from decision-dependent facility uncertainty resulting from intertwined learning and optimization.

In a related area, the online facility location problem assigns incoming customers to existing or new facilities \citep{fotakis2011online}. Approximation algorithms have achieved constant to logarithmic regret \citep[e.g.,][]{meyerson2001online,fotakis2008competitive,guo2020facility}. Recent extensions have incorporated predictions on facility-customer assignments to design algorithms that are $\calO(1)$-competitive when predictions are good and $\calO(\log n)$-competitive or $\calO\left(\frac{\log n}{\log\log n}\right)$-competitive otherwise \citep{jiang2021online,fotakis2021learning}. \cite{almanza2021online} obtained similar results with predictions on the set of facilities to open. The diminishing returns in facility location also link to submodular maximization, which seeks a subset of items to maximize a submodular function. \cite{lattanzi2020fully} and \cite{monemizadeh2020dynamic} derived $(1/2-\varepsilon)$-approximations; \cite{agarwal2023learning} proposed an algorithm that uses predictions on insertions and deletions. Similarly, the stochastic submodular cover problem seeks a subset that covers a submodular function under supply-side uncertainty. \cite{goemans2006stochastic} characterized the benefits of adaptivity; \cite{agarwal2019stochastic} and \cite{ghuge2022power} showed that limited adaptivity can yield most of the benefits. Within this literature, our paper defines a new problem with a chance constraint on the number of successful facilities or connected customers; moreover, it embeds an online learning paradigm in which data samples are accrued dynamically, as opposed to relying on offline advice.

This distinction links our problem to the exploration-exploitation trade-off in multi-armed bandits. Our problem exhibits a contextual Bernoulli bandit structure with i.i.d. features, and simultaneous arm pulls. An extensive literature seeks regret bounds in multi-armed contextual bandits over an asymptotic planning horizon \citep[see, e.g.,][]{lu2010contextual,rusmevichientong2010linearly,slivkins2011contextual,agarwal2012contextual,magureanu2014lipschitz,combes2017minimal,neu2022lifting,van2024optimal}. \cite{cao2021online} cast an online facility location problem in mobile retail as a contextual bandit with continuous routing approximations. \cite{farias2011irrevocable} considered a problem with irrevocable decisions, in which each discarded arm cannot be pulled again. In contrast, the irreversibility in our setting materializes through the objective of minimizing the number of arms pulls---due to large cost commitments---to achieve a probabilistic coverage target. This setting creates inter-temporal dependencies that augment the belief state with a physical state, leading to a challenging dynamic programming structure \citep{ryzhov2012knowledge,russo2018learning,osband2019deep}. Moreover, we seek regret bounds in a novel asymptotic regime with a large coverage target but a finite horizon, which challenge the use of traditional explore-then-commit algorithms from the multi-armed bandit literature. This regime relates to the adaptive experimentation problem from \cite{che2023adaptive} in a finite horizon with asymptotically large batches. Ultimately, our paper formalizes a new online learning and optimization problem with target coverage and irreversible decisions, and derives asymptotically tight bounds and asymptotically optimal algorithms in a finite-horizon regime with a large coverage target.

\section{The online learning and optimization setting}
\label{sec:setting}

We define an online learning and optimization problem in which a decision-maker opens facilities sequentially to meet a target level of coverage by the end of a finite planning horizon. Each facility opening is costly and irreversible, and its success is uncertain at the time of the decision. To guide these decisions, the decision-maker relies on a machine learning algorithm trained on the outcomes of previous opening attempts. Within each period, decisions and feedback unfold according to the sequence illustrated in Figure~\ref{fig:timeline}. Specifically, each period consists of the following steps:
\begin{enumerate}
    \item (\textbf{Realization}) By the start of each period, the decision-maker observes facilities' successes and failures from the previous period, and assembles a labeled dataset of all past outcomes.
    \item (\textbf{Prediction}) The decision-maker updates a machine learning model that predicts the success or failure of the remaining facilities, inducing a set of ``whitelisted'' facilities.
    \item (\textbf{Decision}) The decision-maker attempts to open a subset of facilities among the whitelist. Successful facilities remain opened throughout and contribute toward the coverage target.
\end{enumerate}

\begin{figure}[h!]
    \centering
    \begin{tikzpicture}
        \draw[->,thick] (-1.5,1) -- (14.5,1) node[yshift=-6pt,right]{$\cdots$};
        \node at (0,1.5) {1. Realization, time $t$};
        \node at (4,1.5) {2. Prediction, time $t$};
        \node at (8,1.5) {3. Decision, time $t$};
        \node at (12,1.5) {1. Realization, time $t+1$};
        \draw[thick] (-1.25,2.2) rectangle (1.25,4.7);
        \foreach \i in {2,...,4}{\foreach \j in {0}{
                    \node[circle,thick,draw=mygreen,fill=mygreen,minimum size=1pt] at (-0.875+.45*\j,2.575+.45*\i){};
                    }}
        \foreach \i in {0,...,1}{\foreach \j in {0}{
                    \node[circle,thick,draw=myred,fill=myred,minimum size=1pt] at (-0.875+.45*\j,2.575+.45*\i){};
                    }}
        \foreach \i in {0,...,4}{\foreach \j in {1,...,4}{
                    \node[circle,thick,draw=black,fill=white,minimum size=1pt] at (-0.875+.45*\j,2.575+.45*\i){};
                    }}
        \draw[thick] (2.75,2.2) rectangle (5.25,4.7);
        \foreach \i in {2,...,4}{\foreach \j in {0}{
                    \node[circle,thick,draw=mygreen,fill=mygreen,minimum size=1pt] at (3.125+.45*\j,2.575+.45*\i){};
                    }}
        \foreach \i in {0,...,1}{\foreach \j in {0}{
                    \node[circle,thick,draw=myred,fill=myred,minimum size=1pt] at (3.125+.45*\j,2.575+.45*\i){};
                    }}
        \foreach \i in {0,...,4}{\foreach \j in {1,...,4}{
                    \node[circle,thick,draw=black,fill=white,minimum size=1pt] at (3.125+.45*\j,2.575+.45*\i){};
                    }}
        \foreach \i in {1}{\foreach \j in {2}{
                    \node[circle,thick,draw=black,fill=black,minimum size=1pt] at (3.125+.45*\j,2.575+.45*\i){};
                    }}
        \foreach \i in {2}{\foreach \j in {3}{
                    \node[circle,thick,draw=black,fill=black,minimum size=1pt] at (3.125+.45*\j,2.575+.45*\i){};
                    }}
        \foreach \i in {4}{\foreach \j in {3}{
                    \node[circle,thick,draw=black,fill=black,minimum size=1pt] at (3.125+.45*\j,2.575+.45*\i){};
                    }}
        \foreach \i in {3}{\foreach \j in {4}{
                    \node[circle,thick,draw=black,fill=black,minimum size=1pt] at (3.125+.45*\j,2.575+.45*\i){};
                    }}
        \draw[thick] (6.75,2.2) rectangle (9.25,4.7);
        \foreach \i in {2,...,4}{\foreach \j in {0}{
                    \node[circle,thick,draw=mygreen,fill=mygreen,minimum size=1pt] at (7.125+.45*\j,2.575+.45*\i){};
                    }}
        \foreach \i in {0,...,1}{\foreach \j in {0}{
                    \node[circle,thick,draw=myred,fill=myred,minimum size=1pt] at (7.125+.45*\j,2.575+.45*\i){};
                    }}
        \foreach \i in {0,...,4}{\foreach \j in {1}{
                    \node[circle,thick,draw=myblue,fill=myblue,minimum size=1pt] at (7.125+.45*\j,2.575+.45*\i){};
                    }}
        \foreach \i in {3,...,4}{\foreach \j in {2}{
                    \node[circle,thick,draw=myblue,fill=myblue,minimum size=1pt] at (7.125+.45*\j,2.575+.45*\i){};
                    }}
        \foreach \i in {0,...,2}{\foreach \j in {2}{
                    \node[circle,thick,draw=black,fill=white,minimum size=1pt] at (7.125+.45*\j,2.575+.45*\i){};
                    }}
        \foreach \i in {0,...,4}{\foreach \j in {3,...,4}{
                    \node[circle,thick,draw=black,fill=white,minimum size=1pt] at (7.125+.45*\j,2.575+.45*\i){};
                    }}
        \foreach \i in {1}{\foreach \j in {2}{
                    \node[circle,thick,draw=black,fill=black,minimum size=1pt] at (7.125+.45*\j,2.575+.45*\i){};
                    }}
        \foreach \i in {2}{\foreach \j in {3}{
                    \node[circle,thick,draw=black,fill=black,minimum size=1pt] at (7.125+.45*\j,2.575+.45*\i){};
                    }}
        \foreach \i in {4}{\foreach \j in {3}{
                    \node[circle,thick,draw=black,fill=black,minimum size=1pt] at (7.125+.45*\j,2.575+.45*\i){};
                    }}
        \foreach \i in {3}{\foreach \j in {4}{
                    \node[circle,thick,draw=black,fill=black,minimum size=1pt] at (7.125+.45*\j,2.575+.45*\i){};
                    }}
        \draw[thick] (10.75,2.2) rectangle (13.25,4.7);
        \foreach \i in {2,...,4}{\foreach \j in {0}{
                    \node[circle,thick,draw=mygreen,fill=mygreen,minimum size=1pt] at (11.125+.45*\j,2.575+.45*\i){};
                    }}
        \foreach \i in {0,...,1}{\foreach \j in {0}{
                    \node[circle,thick,draw=myred,fill=myred,minimum size=1pt] at (11.125+.45*\j,2.575+.45*\i){};
                    }}
        \foreach \i in {3,...,4}{\foreach \j in {1}{
                    \node[circle,thick,draw=mygreen,fill=mygreen,minimum size=1pt] at (11.125+.45*\j,2.575+.45*\i){};
                    }}
        \foreach \i in {1,...,2}{\foreach \j in {1}{
                    \node[circle,thick,draw=myred,fill=myred,minimum size=1pt] at (11.125+.45*\j,2.575+.45*\i){};
                    }}
        \foreach \i in {0}{\foreach \j in {1}{
                    \node[circle,thick,draw=mygreen,fill=mygreen,minimum size=1pt] at (11.125+.45*\j,2.575+.45*\i){};
                    }}
        \foreach \i in {3,...,4}{\foreach \j in {2}{
                    \node[circle,thick,draw=mygreen,fill=mygreen,minimum size=1pt] at (11.125+.45*\j,2.575+.45*\i){};
                    }}
        \foreach \i in {0,...,2}{\foreach \j in {2}{
                    \node[circle,thick,draw=black,fill=white,minimum size=1pt] at (11.125+.45*\j,2.575+.45*\i){};
                    }}
        \foreach \i in {0,...,4}{\foreach \j in {3,...,4}{
                    \node[circle,thick,draw=black,fill=white,minimum size=1pt] at (11.125+.45*\j,2.575+.45*\i){};
                    }}
    \end{tikzpicture}
    \caption{Timeline of events. [White dots: whitelisted facilities; black dots: other facilities; blue dots: facility opening attempts; green/red dots: successfully/unsuccessfully opened facilities.]}
    \label{fig:timeline}
    \vspace{-12pt}
\end{figure}

This setting encompasses a broad class of multi-period predict-and-optimize problems in which decisions both commit resources and generate information. This creates intertwined online learning and online optimization structures, whereby decisions influence both immediate outcomes and the quality of future predictions. In particular, decisions are biased toward facilities that are likely to succeed, which leads to non-independent samples and challenges the use of standard statistical learning results to characterize the learning error as a function of past decisions.

In response, we introduce in this section the modeling setup and the key assumptions that enable a tractable analysis. Section~\ref{subsec:DMenvironment} introduces the decision-making and learning environment and formalizes the online \emph{optimization} problem governing facility selection. Section~\ref{subsec:statistical_learning} then introduces the assumptions underlying the online \emph{learning} process, including a characterization of the learning error rate as a function of the sample size. Section~\ref{subsec:setup} constructs a general statistical learning setup under which we prove that the postulated learning rate holds. We then use this error rate decay throughout the paper to formulate and solve the online learning and optimization problem. For simplicity, we focus in this section on a core setting with homogeneous facilities (Sections~\ref{sec:base} and~\ref{sec:implications}), but arguments extend to a networked setting with heterogeneous facilities (Section~\ref{sec:network}).

Throughout this paper, we adopt standard asymptotic notation: (i) $f(x)=\calO(g(x))$ if there exist $M>0$ and $\bar{x}$ such that $f(x)\leq Mg(x)$ for all $x\geq\bar{x}$; (ii) $f(x)=o(g(x))$ if for all $M>0$ there exists $\bar{x}$ such that $f(x)\leq Mg(x)$ for all $x\geq\bar{x}$; (iii) $f(x)=\Omega(g(x))$ if there exist $M>0$ and $\bar{x}$ such that $f(x)\geq Mg(x)$ for all $x\geq\bar{x}$; and (iv) $f(x)=\Theta(g(x))$ if $f(x)=\calO(g(x))$ and $f(x)=\Omega(g(x))$.

\subsection{The decision-making and learning environment}
\label{subsec:DMenvironment}

Let $\calF$ denote a set of $M$ candidate facilities. A decision-maker targets to open $m$ facilities over a finite horizon of $T\geq 2$ periods, stored in a set $\calT=\{1,\cdots,T\}$. We consider the asymptotic regime where $m\to\infty$ but $T\geq2$ is finite. This is motivated by our practical examples where the decision-maker needs to meet a large coverage target in a few rounds due to the time and cost of each deployment wave, resulting in a much higher number of deployments in each period (e.g., clinical trials sites, vaccination centers) than the number of periods $T$. We define the following variables:
\begin{align*}
    y_{it}&=\begin{cases}
        1&\text{if the decision-maker attempts to open facility $i\in\calF$ at time $t\in\calT$}\\
        0&\text{otherwise.}
    \end{cases}
\end{align*}

Let $S_{it}$ denote the binary indicator of the (uncertain) success of facility $i\in\calF$ at time $t\in\calT$. We assume that success is revealed by the next period; in the clinical trial example, this means that the recruitment potential can be determined before the next wave of site openings. We also assume that success is independent across facilities; in the example, this reflects a setting where recruits are non-overlapping across sites. We track successful facilities via the following variables:
\begin{equation}\label{eq:z}
    z_{it} = \begin{cases}
        1 & \text{if facility $i\in\calF$ is open at time $t\in\calT$, i.e. if}\ S_{it} = 1 \;\; \text{and}\;\; y_{it}=1\\ 
        0 & \text{otherwise, i.e., if}\ S_{it} = 0 \;\; \text{or}\;\; y_{it}=0
    \end{cases}
\end{equation}

Let $A_t=\sum_{i\in\calF}y_{it}$ and $B_t=\sum_{i\in\calF}z_{it}$ be the number of facilities tentatively and successfully opened at time $t\in\calT$. To characterize the facility selection process, we adopt a learning framework in which the decision-maker has access to an online classifier characterizing ``whitelisted'' facilities in each round and samples facility openings among those. We do not impose a priori distributional assumptions on facility success, but assume a declining failure rate among whitelisted facilities in Assumption~\ref{ass:classification}. In Section~\ref{subsec:setup}, we present a machine learning environment satisfying this property.

Specifically, we consider a stationary learning environment \citep[e.g.,][]{auer2002using, agrawal2012analysis}. We denote by $\eta_{i} \in [0,1]$ the (unknown) probability of success of facility $i\in\calF$, so that facility success is governed by a Bernoulli random variable $S_{it}\sim \text{Ber}(\eta_{i})$. Each facility $i\in\calF$ is associated with a feature vector $\bx_{i} \in \calX\subseteq\R^{p}$. We treat the features $\{\bx_i:i=1,\cdots,M\}$ as fixed from the set of candidate facilities $\calF$, and all probability statements are conditional on these features.

In each period $t\in\calT$, the learning population consists of all facilities that have been tentatively opened. The decision-maker has access to the following data sample, of size denoted by $N_{t-1}$:
$$\calS_t=\bigcup_{s=1}^{t-1}\bigcup_{i\in\calF:y_{is}=1}\left\{(\bx_{i},S_{is})\right\},\ \text{with}\ |\calS_t|=N_{t-1}=\sum_{s=1}^{t-1}A_s$$

The decision-maker builds a classification model for the remaining $M-N_{t-1}$ facilities. Let $\calD = \bigcup_{n=1}^\infty (\calX^n\times\{0,1\}^n)$ denote the space of all finite datasets, and $\calH \subseteq \{h:\calX\to\{0,1\}\}$ the space of feasible classification models. The classification model in period $t\in\calT$ is denoted by $h_t = F(\mathcal{S}_t)$, where $F: \calD \to \mathcal{H}$. Let $\calJ_t$ denote the set of unexplored facilities with positive classification:
\begin{equation}
    \calJ_t=\{i\in\calF\mid y_{i1}=\cdots=y_{i,t-1}=0\ ;\ h_t(\bx_{i})=1\}.\label{eq:Jt}
\end{equation}

We assume that the decision-maker selects facilities randomly with replacement among the set $\calJ_t$. This process yields the following assumption on the conditional probability of facility openings:
\begin{assumption}[Sampling]\label{ass:sampling}
    At each time period $t\in\calT$, $\P\left(y_{it}=1 \mid i \in \calJ_t\right)= 
    \frac{A_t}{|\calJ_t|}$.
\end{assumption}

This assumption shifts the decision problem from \emph{which} facilities to open (through the variables $y_{it}$) to \emph{how many} facilities to open (through the variables $A_t$), and allows us to focus on the dynamic interaction between learning and optimization through the aggregate sample size $N_{t-1}$. Specifically, the learning process (captured through the online classifier $h_t$) determines the candidate set $\calJ_t$ of ``whitelisted'' facilities, and the optimization problem determines the number $A_t$ of facilities to select from that set. One limitation of Assumption~\ref{ass:sampling} is that it does not allow the decision-maker from differentiating among whitelisted facilities based on their relative predicted probabilities of success. In this sense, this assumption captures a setting with weakly coordinated data science and optimization teams, typical of large-scale organizational process. From a technical standpoint, it captures the core statistical and optimization trade-offs, and facilitates the convergence of the statistical process by fostering exploration. Another limitation in Assumption~\ref{ass:sampling} is that it allows for the same facility to be selected multiple times in the same period. Alternatively, we could assume that facilities are sampled without replacement. This would induce negative correlation among the $y_{it}$ variables and require the use of concentration inequalities for sampling without replacement \citep{serfling1974probability, bardenet2015concentration}. Our modeling simplification facilitates the analysis without materially affecting the results, and is common in statistics—e.g., repeated sampling from an asymptotically large superpopulation \citep{Neyman1934,SarndalSwenssonWretman1992}.

We can now formally introduce our decision-making problem. The decision-maker aims to minimize the number of tentative facility openings subject to a chance constraint ensuring that $m$ facilities will be successful with sufficiently high probability $1-\delta$. The problem is given as follows:
\begin{align}\label{prob:original}
\min_{\bA,\bB}\quad \left\{\sum_{t=1}^T A_t\quad\Bigg\rvert\quad\P\left[\sum_{t=1}^TB_t\geq m\right] \geq 1-\delta ;\ B_t=\sum_{i\in\calF:y_{it}=1}S_{it}, \forall t\in\calT;\ A_t,B_t\geq0,\ \forall t\in\calT\right\}
\end{align}
where the probability $\P$ is measured over the randomness of facilities sampling ($y_{it}$) and of their success in \emph{all} previous periods ($S_{i1},\cdots,S_{it}$). The main difficulty in this formulation involves linking facility openings (through the variables $A_t$) to successful facilities (through the variables $B_t$). To this end, we introduce assumptions on the coupled learning and optimization environment.

\subsection{Statistical learning assumptions}
\label{subsec:statistical_learning}
To enable our theoretical analysis, Assumption~\ref{ass:positive} ensures that the candidate pool $M$ grows faster than the target $m$ and that the overall “mass” of success probability within $\calJ_t$ is large enough.
\begin{assumption}[Feasibility]\label{ass:positive}
    As $m \to \infty$, we have $M=\Omega(m)$. Furthermore, there exists $\xi>0$ such that the success probabilities within $\calJ_t$ satisfy $\sum_{i \in \calJ_t}\eta_i\geq \xi M$ for all $t\in\calT$.
\end{assumption}

This is a mild technical condition ensuring the feasibility of the optimization problem to select $m$ successful facilities from the eligible set $\calJ_t$. It also ensures that the size of the eligible pool $|\calJ_t|$ dominates the number of selections $A_t$. Thus, although our Assumption \ref{ass:sampling} allows the same facility to be sampled multiple times, that probability remains asymptotically vanishing per Assumption~\ref{ass:positive}.

Next, the error rate of the classification model $h_t(\cdot)$ is defined as the failure probability of a facility within the whitelisted set $\calJ_t$, i.e.: $\P(S_{it} \neq 1\mid i \in \calJ_t)=\frac{1}{|\calJ_t|} \sum_{i \in \calJ_t} \mathbf{1}\{S_{it}\neq 1\}$. This probability characterizes the average failure rate across all samples in $\calJ_t$, but different facilities within $\calJ_t$ have different probabilities of failure governed by the (unknown) parameters $\eta_i$. We make the following non-parametric assumption on the error rate, described by an initial error $\varepsilon$, a learning rate of order $\calO(1/n^{r})$ with $r>0$, and an irreducible error term $1-p$.
\begin{assumption}[Learning function] \label{ass:classification}
    There exists $0<\varepsilon\leq 1$ such that:
    \begin{align*}
        \P(S_{it}\neq1 \mid i \in \calJ_t)=\frac{\varepsilon\cdot p}{\left(N_{t-1} + 1\right)^r}+\varepsilon\cdot(1-p).
    \end{align*}
\end{assumption}

Whereas this error rate resembles classical statistical learning results \citep{boucheron2005theory, vapnik2015uniform, feldman2019high}, the key distinction lies in the dependence structure of the data. By design, the whitelisted set $\calJ_{t}$ is biased toward instances with positive classifications, and so is the next data sample $\calS_{t+1}$. This feedback loop violates i.i.d. assumptions, so standard generalization bounds cannot be applied. We introduce next a statistical learning environment that explicitly models the dependence structure and derives bounds of the same functional form. This construction serves as one instantiation|rather than a unique characterization|of statistical conditions under which Assumption \ref{ass:classification} holds; moreover, these statistical conditions are not invoked in our online learning and optimization model, so our results apply broadly as long as Assumption~\ref{ass:classification} is satisfied. In Appendix \ref{app:UCI}, we provide empirical evidence supporting Assumption~\ref{ass:classification} in our online learning and optimization setting, using real-world data.

\subsection{A statistical learning setup.}
\label{subsec:setup}

Assumption~\ref{ass:positive} states that the pool of candidate facilities is large enough to enable sampling among unexplored facilities. In this section, we operationalize that assumption by stating that candidate pool $M$ grows sufficiently faster than the target number of successful facilities $m$, in $\Omega(m^{3/2})$.
\begin{stat}[Asymptotic regime]
    \label{ass:regime}
   As $m \to \infty$, $M=\Omega(m^{3/2})$.
\end{stat}
This condition simplifies the technical exposition to effectively ignore previously selected facilities at a rate $o(m^{-1/2})$ and thus achieve the best convergence rate. Similar insights would hold for alternative scalings of the form $M=\Omega(m^{1+\omega})$ with $\omega>0$.

Next, we model success probabilities via a parametric family $f_\theta$ at unknown parameter value $\theta_0$. We do not make assumptions on the functional form of $f_\theta$.
\begin{stat}[Generative model]\label{ass:known}
The success probability for facility $i$ is $\eta_i = f_{\theta_0}(\bx_i)$, where $f_\theta:\mathcal X\to[0,1]$ is known and ${\theta_0}\in\Theta\subseteq\mathbb R^{\ell}$ is an unknown parameter.
\end{stat}

At each period $t$, we estimate the parameter $\theta_t$ using the maximum likelihood estimator (MLE): 
    \begin{equation}
        \widehat{\theta}_t= \argmax_{\theta\in\Theta} \ \prod_{s=1}^{t-1} \prod_{i\in\calJ_s:y_{is}=1} (f_\theta(\bx_{i}))^{z_{is}}(1-f_\theta(\bx_{i}))^{1-z_{is}} \label{eq:mle_estimation}
    \end{equation}
We leverage the estimator $f_{\widehat{\theta}_t}(\cdot)$ to specify the classifier $h_t(\cdot)$ guiding the set $\calJ_t$. One possibility would be to apply a greedy cutoff rule, by whitelisting the facilities $i\in\calF$ such that the probability of success $f_{\widehat{\theta}_t}(x_{i})$ exceeds a threshold $\tau$ (in particular, a value of $\tau=1/2$ would correspond to a symmetric misclassification loss between false positives and false negatives). We provide a specification that augments this rule with a (possibly empty) exploration set $\calE_t\subset \calF$:
\begin{equation}
    h_t(\bx_i)=\mathbf{1}\left\{f_{\widehat{\theta}_t}(\bx_{i})>\tau\right\}\cup \mathbf{1}\left\{i \in \calE_t\right\}\label{eq:exploration}
\end{equation}
The exploration set can be required to guarantee convergence to the optimal classifier when facilities with high success probabilities exhibit limited variability in their features (as discussed below). We impose that its size covers a sufficiently small---and vanishing---part of the set of facilities.
\begin{stat}[Exploration]\label{ass:exploration}
There exists an absolute constant $K$ such that the exploration set $\calE_t$ satisfies $\lim_{m \to \infty} \frac{|\calE_t|}{M} \leq \frac{K}{(N_{t-1})^{1/2}}$ for all periods $t\in\calT$. 
\end{stat}
This condition amounts to a mild restriction. Indeed, $M=\Omega(m^{3/2})$ per Condition~\ref{ass:regime} and as discussed below, the sample satisfies $N_{t-1}=\Theta(m)$ in our online learning and optimization process. Therefore, the exploration set can still have an asymptotically unbounded size of $|\mathcal{E}_t|=\Omega(m)$. 

We define the Bayes-optimal classifier based on the fully-learned function with perfect information on $\theta_0$ as $h^*(\bx_i)=\mathbf{1}\left\{f_{\theta_0}(\bx_{i})>\tau\right\}$. Note that the Bayes-optimal classifier is based on the ``true'' success probabilities without any exploration set. We define its error rate as follows:
$$\mu^*=\P(S_{it}\neq1 \mid h^*(\bx_i)=1)=1-\frac{\sum_{i\in\calF:\eta_{i}>\tau}f_{\theta_0}(\bx_{i})}{\left|i\in\calF:f_{\theta_0}(\bx_{i})>\tau\right|}$$

Next, we make a margin assumption to bound the concentration of $f_{\theta_0}$ around $\tau$ across all facilities $i\in\calF$.
\begin{stat}[Margin]\label{ass:margin}
Define the probability measure as the counting measure over the facilities $i\in\calF$: $\P\left(\left|f_{\theta_0}(\bx_{i})-\tau\right|\leq\nu\right)=1/M\cdot\sum_{i\in\calF} \mathbf{1}\{\left|f_{\theta_0}(\bx_{i})-\tau\right|\leq\nu\}$. Then there exist $C>0$ and $\alpha>0$ such that $\P\left(\left|f_{\theta_0}(\bx_{i})-\tau\right|\leq\nu\right)\leq C \nu^{\alpha}.$
\end{stat}
This condition is commonly used in statistical learning theory \citep{boucheron2005theory, tsybakov2004optimal}. It ensures that the true label is ``clear'' from the classification boundary for most samples.

Finally, we impose the following regularity conditions on the prediction function.

\begin{stat}[Regularity]\label{ass:regularity}
\begin{enumerate}
    \item $\Theta\subseteq\R^{l}$ and $\calX\subseteq\R^{p}$ are compact sets. 
    \item $f_\theta(\bx)$ is a twice continuously-differentiable function of $\theta$ uniformly in $\bx \in \mathcal{X}$. Furthermore, there exists $\varepsilon>0$ such that $\varepsilon \leq f_\theta(\bx) \leq 1-\varepsilon$ for all $\theta \in \Theta$ and $\bx \in \mathcal{X}$.
    \item Define $R_t(\theta\mid \calJ_t)$ as the expected log-likelihood in period $t\in\calT$: \[R_t(\theta \mid \calJ_t)=\frac{1}{|\calJ_t|}\sum_{i \in \calJ_t}\eta_i\log(f_\theta(\bx_{i}))+(1-\eta_i)\log(1-f_\theta(\bx_{i}))\]
    The realized sequence $\calJ_t$ is identifiable: $R_t(\theta \mid \calJ_t)=R_t(\theta_0\mid \calJ_t) \iff \theta = \theta_0\ \forall t\in\calT$.
    \item The Fisher information $\mathcal{I}_t(\theta \mid \calJ_t)$, is positive definite at $\theta=\theta_0$ for the realized sequence $\calJ_t$:
        \[\mathcal{I}_t(\theta_0\mid \calJ_t) := \frac{1}{|\calJ_t|} \sum_{i \in \calJ_t}\frac{1}{f_{\theta_0}(\bx_i)(1-f_{\theta_0}(\bx_i))}\frac{\partial f_{\theta_0}(\bx_i)}{\partial \theta}\frac{\partial f_{\theta_0}(\bx_i)}{\partial \theta^T} \succ 0 \qquad \forall t\in\calT\]
\end{enumerate}
\end{stat}    
These are standard regularity conditions for a well-behaved MLE estimator. The second condition ensures that the log-likelihood of the estimator does not tend toward infinity. This is satisfied by virtually all practical algorithms (e.g. logistic regression). Otherwise, we could relax it with local bracketing arguments \citep{van1996weak}. The third condition states that the features $\bx_i$ of the selected facilities in $\calJ_t$ must be rich enough to distinguish between different parameters $\theta$. The last condition ensures that the Fisher information matrix formed by these features is full rank around $\theta_0$ to ensure asymptotic normality. For instance, in a generalized linear model $f_\theta(\bx)=\sigma(\bx^\top\theta)$ with a monotone link function $\sigma$, at least $p$ feature vectors must be linearly independent among the $|\mathcal{J}_t|=\Theta(M)$ selected candidates. Recall that the set of whitelisted facilities is given by $\calJ_t=\{i\in\calF\mid y_{i1}=\cdots=y_{i,t-1}=0\ ;\ h_t(\bx_{i})=1\}$ (Equation~\eqref{eq:Jt}) with the classifier $h_t(\bx_i)=\mathbf{1}\left\{f_{\widehat{\theta}_t}(\bx_{i})>\tau\right\}\cup \mathbf{1}\left\{i \in \calE_t\right\}$ (Equation~\eqref{eq:exploration}). The exploration set $\calE_t$ may be required to capture sufficient variation in the feature space and ensure linear independence. This choice of the exploration set $\calE_t$ is problem specific, as it depends on the features $\{\bx_i:i\in\calF\}$. In an extreme example where the facilities in $\calF$ do not cover the full feature space, then identifiability would fail regardless of the facility sampling strategy. When all features $\bx_i$ are continuous, the half-space $\mathbf{1}\{f_{\widehat\theta_t}(\bx_i)>\tau\}$ typically covers a non-degenerate region of $\calX$, so linear independence can hold even with $\calE_t=\emptyset$. If $\bx$ includes categorical variables, the rule $\mathbf{1}\{f_{\widehat\theta_t}(\bx_i)>\tau\}$ may only select a subset of the categorical levels, leaving the design matrix rank-deficient, but a vanishing exploration set $\calE_t$ covering all categorical levels can retrieve full-rank coverage of the feature space for $\calJ_t$.

\subsubsection*{Rate of convergence.}

We can now prove that the error rate of the classifier $h_t(\cdot)$ converges to the Bayes-optimal classifier at a rate $\calO(n^{-\min\{\alpha,1\}\times (1/2-\gamma)})$ for any $\gamma>0$, where $n$ is the sample size accrued over time. In particular, the best possible convergence rate is $\calO(\sqrt n)$ but can get smaller in the case of a weak margin separation from the cutoff $\tau$ (Statistical Condition~\ref{ass:margin}).
\begin{proposition}\label{prop:MLE}
     Define the error rate $\mu_t = \P(S_{it}\neq 1 \mid i\in\calJ_t)$. Assume that $N_t \to \infty$ and $N_t=\calO(m)$ as $m \to \infty$ for all $t$. Under Assumptions~\ref{ass:sampling}-\ref{ass:positive} and Statistical Conditions~\ref{ass:regime}--\ref{ass:regularity}, for any $\gamma>0$, there exist $\overline\varepsilon>0$ and $m_0$ such that, for all $m\geq m_0$ and $t\in\calT$:
    \[\mu_t \leq \frac{\overline\varepsilon}{(N_{t-1}+1)^{\min\{\alpha,1\}\times (1/2-\gamma)}}+\mu^*.\]
\end{proposition}

Note that the proposition is focused on the learning process, and relies on the condition that $N_t=\calO(m)$. This condition is easily satisfied in our learning and optimization problem (Section~\ref{subsec:DMenvironment}). Indeed, Proposition~\ref{prop:regret_NL} below shows that a no-learning baseline opens on the order of $m$ facilities, so our online learning and optimization approach will open up to $\calO(m)$ facilities.

The proof (in~\ref{app:stats}) bounds the difference between the classifier's error rate $\mu_t$ and the Bayes-optimal error $\mu^*$ by: (i) the sampling probability of each facility; (ii) the probability mass around $\tau$; (iii) the classifier's error; and (iv) the mass of exploration facilities. We show that these terms are respectively bounded by $\calO(1/N_{t-1}^{1/2-\gamma})$, $\calO(1/N_{t-1}^{\alpha\cdot(1/2-\gamma)})$, $\calO(1/N_{t-1}^{1/2-\gamma})$, and $\calO(1/N_{t-1}^{1/2})$.

As expected, the classification error goes down over time as the sample becomes larger. As the sample size grows, the ``true'' success probability of each facility $\i\in\calF$ remains unchanged|equal to $\eta_i$|but the online classifier becomes stronger, so the online learning and optimization process selects facilities that are more likely to succeed. Thus, even though facility success is unaffected by the predictions, the success probabilities of the \textit{whitelisted} facilities goes up, as shown in Figure~\ref{fig:error}.

\begin{figure}[h!]
    \centering
    \subfloat[Small sample size]{\label{subfig:error1}
    \begin{tikzpicture}
        \draw[thick] (-.5,-.5) rectangle (4.5,4.5);
        \foreach \i in {0,...,3}{\foreach \j in {0,...,8}{
                    \node[circle,draw=black,fill=black!50,minimum size=1pt] at (.5*\j,.5*\i){};
                    }}
        \foreach \i in {4,...,8}{\foreach \j in {0,...,8}{
                    \node[circle,draw=black,fill=black!20,minimum size=1pt] at (.5*\j,.5*\i){};
                    }}
        \node[circle,ultra thick,draw=myred,fill=myred!80,minimum size=1pt] at (.5,.5){};
        \node[circle,ultra thick,draw=myred,fill=myred!80,minimum size=1pt] at (3.5,1){};
        \node[circle,ultra thick,draw=myred,fill=myred!20,minimum size=1pt] at (2.5,2.5){};
        \node[circle,ultra thick,draw=myred,fill=myred!20,minimum size=1pt] at (1,3.5){};
    \end{tikzpicture}
    }
    \subfloat[Medium sample size]{\label{subfig:error2}
    \begin{tikzpicture}
        \draw[thick] (-.5,-.5) rectangle (4.5,4.5);
        \foreach \i in {0,...,3}{\foreach \j in {0,...,8}{
                    \node[circle,draw=black,fill=black!50,minimum size=1pt] at (.5*\j,.5*\i){};
                    }}
        \foreach \i in {4,...,8}{\foreach \j in {0,...,8}{
                    \node[circle,draw=black,fill=black!20,minimum size=1pt] at (.5*\j,.5*\i){};
                    }}
        \node[circle,ultra thick,draw=myred,fill=myred!80,minimum size=1pt] at (1.5,.5){};
        \node[circle,ultra thick,draw=myred,fill=myred!80,minimum size=1pt] at (2.5,1.5){};
        \node[circle,ultra thick,draw=myred,fill=myred!20,minimum size=1pt] at (3.5,2.5){};
        \node[circle,ultra thick,draw=myred,fill=myred!20,minimum size=1pt] at (0.5,3){};
        \node[circle,ultra thick,draw=myred,fill=myred!20,minimum size=1pt] at (4,3.5){};
        \node[circle,ultra thick,draw=myred,fill=myred!20,minimum size=1pt] at (2,4){};
    \end{tikzpicture}
    }
    \subfloat[Large sample size]{\label{subfig:erroR2}
    \begin{tikzpicture}
        \draw[thick] (-.5,-.5) rectangle (4.5,4.5);
        \foreach \i in {0,...,3}{\foreach \j in {0,...,8}{
                    \node[circle,draw=black,fill=black!50,minimum size=1pt] at (.5*\j,.5*\i){};
                    }}
        \foreach \i in {4,...,8}{\foreach \j in {0,...,8}{
                    \node[circle,draw=black,fill=black!20,minimum size=1pt] at (.5*\j,.5*\i){};
                    }}
        \node[circle,ultra thick,draw=myred,fill=myred!80,minimum size=1pt] at (3,0){};
        \node[circle,ultra thick,draw=myred,fill=myred!20,minimum size=1pt] at (0.5,2){};
        \node[circle,ultra thick,draw=myred,fill=myred!20,minimum size=1pt] at (2,3){};
        \node[circle,ultra thick,draw=myred,fill=myred!20,minimum size=1pt] at (3,4){};
        \node[circle,ultra thick,draw=myred,fill=myred!20,minimum size=1pt] at (4,2.5){};
        \node[circle,ultra thick,draw=myred,fill=myred!20,minimum size=1pt] at (0,4){};
        \node[circle,ultra thick,draw=myred,fill=myred!20,minimum size=1pt] at (1,2.5){};
        \node[circle,ultra thick,draw=myred,fill=myred!20,minimum size=1pt] at (2.5,3.5){};
    \end{tikzpicture}
    }
    \caption{Illustration of the evolution of the online classifier over time. Light (resp.) dark dots: facilities with positive (resp. negative) realizations, i.e., $S_{it}=1$ (resp. $S_{it}=0$); red dots: selected facilities, i.e., $y_{it}=1$.}
    \label{fig:error}
    \vspace{-12pt}
\end{figure}
Importantly, the upper bound in Proposition \ref{prop:MLE} matches Assumption \ref{ass:classification} with $\varepsilon = \overline\varepsilon + \mu^*$ and $p=\frac{\overline\varepsilon}{\overline\varepsilon+\mu^*}$; the learning rate $\calO(1/n^r)$ generalizes the best-case rate of $\calO(\sqrt n)$ identified in Proposition~\ref{prop:MLE}. In particular, if $\alpha<1$ characterizes a weak margin separation, the online classifier converges at a slower rate $r<0.5$. More broadly, Assumption \ref{ass:classification} covers a range of noise structures, data-generating processes, and model misspecifications beyond the statistical conditions from this section.

Finally, Assumption \ref{ass:classification} brings an additional simplification. Under Proposition \ref{prop:MLE}, the error $\mu_t$ in period $t$ is \textit{bounded above} by a term that depends on the number of past trials; in principle, $\mu_t$ may still vary with the realized data $(y_{is},z_{is})$. Assumption \ref{ass:classification} adopts that upper bound itself as the \textit{exact} error rate, which depends solely on the count $N_{t-1}$. This simplification cleanly couples the learning and optimization parts of the model without tracking the full history. The next result confirms that this simplification is indeed conservative for our chance constraint:
\begin{corollary}[Learning environment] \label{cor:learning}
    Define independent Bernoulli variables $\widetilde{S}_{it}=\Bern\left(1-\frac{\varepsilon\cdot p}{\left(N_{t-1} + 1\right)^r}-\varepsilon\cdot(1-p)\right)$ and $\widetilde{B}_t = \sum_{i:y_{it}=1} \widetilde{S}_{it}$, with $r=\min\{\alpha,1\} \times \{1/2-\gamma\}$, $\varepsilon = \overline\varepsilon + \mu^*$, and $p = \frac{\overline\varepsilon}{\overline\varepsilon+\mu^*}$. Then, $\P\left[\sum_{t=1}^T\widetilde{B}_t\geq m\right] \leq \P\left[\sum_{t=1}^TB_t\geq m\right]$.
\end{corollary}

Consequently, any policy obtained under Assumption \ref{ass:classification} will satisfy the chance constraint under Proposition~\ref{prop:MLE}. In the remainder of this paper, we will consider the corresponding online learning and optimization problem, by modeling success probabilities conservatively according to Assumption~\ref{ass:classification} (rather than Proposition~\ref{prop:MLE}). Thus, Corollary \ref{cor:learning} therefore provides the essential link, so the rest of the analysis can proceed with Assumptions \ref{ass:sampling}--\ref{ass:classification}, rather than the full set of statistical conditions~\ref{ass:regime}–\ref{ass:regularity}.
    
\section{The Core Model: Target on Facilities}
\label{sec:base}

\subsection{Problem formulation and benchmarks}

We consider the core problem with homogeneous facilities and a supply-side target on the number of facilities defined in Section~\ref{subsec:DMenvironment}. This setting enables us to focus on the \textit{number} of facilities to open in the decision-making problem (via $A_t$ and $B_t$) as opposed to the \textit{set} of facilities (via $y_{it}$ and $z_{it}$). We characterize the number of successful facilities $B_t$ at each time period $t\in\calT$ as a binomial distribution under the online learning and optimization setting defined in Section~\ref{sec:setting}, following from the sampling process (Assumption~\ref{ass:sampling}) and the conditional success probabilities (Assumption~\ref{ass:classification}).
\begin{corollary}[Success Distribution]\label{cor:success}
Under Assumptions~\ref{ass:sampling} and \ref{ass:classification}, $B_t$ follows a binomial distribution with $A_t$ trials and success probability $1- \frac{\varepsilon\cdot p}{\left(N_{t-1} + 1\right)^r}-\varepsilon\cdot(1-p)$. 
\end{corollary}

The online learning and optimization problem given in Equation~\eqref{prob:original} can be reformulated as follows. We refer to this formulation as Problem~\eqref{prob:main}, and denote its optimal solution by $m^*(\varepsilon, p,T)$.
\begin{align}
m^*(\varepsilon, p,T)=
\min_{\bA,\bB}\quad & \sum_{t=1}^T A_t \label{prob:main}\tag{$\calP^*$}\\
\st\quad&\P\left[\sum_{t=1}^TB_t\geq m\right] \geq 1-\delta \nonumber\\
\quad& B_t\sim \Bino\left(A_t, 1- \frac{\varepsilon\cdot p}{\left(\sum_{s=1}^{t-1} A_s +1\right)^r}-\varepsilon\cdot(1-p)\right),\ \forall t\in\calT\nonumber\\
\quad& A_t,B_t\geq0,\ \forall t\in\calT\nonumber
\end{align}
Per Corollary~\ref{cor:success}, Problem~\eqref{prob:main} provides an exact reformulation of Equation~\eqref{prob:original} under Assumptions~~\ref{ass:sampling} and \ref{ass:classification}. In case the functional form given in Assumption~\ref{ass:classification} remains an upper bound of the ``true'' error rate, Problem~\eqref{prob:main} provides a conservative reformulation of Equation~\eqref{prob:original} (Corollary~\ref{cor:learning}).

Assumption~\ref{ass:positive} guarantees the asymptotic feasibility of this problem. With the chance constraints, Problem~\eqref{prob:main} amounts to a value-at-risk formulation \citep{uryasev2010var}, which is notoriously challenging to solve. In the remainder of this section, we propose a simple and interpretable algorithm, and prove the asymptotic feasibility and optimality of its solution as $m\to\infty$.

\subsubsection*{Benchmarks.}

We define two single-period benchmarks: a fully-learned benchmark to measure regret, and a no-learning baseline to evaluate the benefits of online learning and optimization.
\begin{itemize}
\item[--] A fully-learned benchmark, with error $\varepsilon (1-p)$. This is equivalent to having access to the Bayes-optimal classifier in advance; when $p=1$, it reduces to a perfect-information benchmark.
$$\widehat{m}(\varepsilon,p)=\min\quad\left\{A\ :\ \P\left[\Bino\left(A, 1-\varepsilon(1-p)\right)\geq m\right] \geq 1-\delta;\ A\geq0\right\}$$
\item[--] A no-learning baseline $m^{\text{NL}}(\varepsilon)$, with error $\varepsilon$.
$$m^{\text{NL}}(\varepsilon)=\min\quad\left\{A\ :\ \P\left[\Bino\left(A, 1-\varepsilon\right)\geq m\right] \geq 1-\delta;\ A\geq0\right\}$$
\end{itemize}

We measure regret against the fully-learned benchmark. The regret of the stochastic solution is given by $\texttt{Reg}^*(\varepsilon, p,T)=m^*(\varepsilon, p,T)-\widehat{m}(\varepsilon,p)$, and the no-learning regret by $\texttt{Reg}^{\text{NL}}(\varepsilon,p)=m^{\text{NL}}(\varepsilon)-\widehat{m}(\varepsilon,p)$. Proposition~\ref{prop:regret_NL} shows that the fully-learned and no-learning benchmarks open $\Theta\left(\frac{m}{1-\varepsilon(1-p)}\right)$ and $\Theta\left(\frac{m}{1-\varepsilon}\right)$ facilities, respectively. Therefore, the no-learning baseline leaves a linear regret.

\begin{proposition}\label{prop:regret_NL}
The no-learning baseline induces linear regret: $\texttt{Reg}^{NL}(\varepsilon)=\Omega(m)$.
\end{proposition}

\subsection{An asymptotically optimal algorithm with sub-linear regret}

To avoid case distinctions, we focus on the case where $r\neq1$; the case where $r=1$ is identical and can even be obtained by taking limits. We denote by $\Phi(\cdot)$ the cumulative distribution function of the standard normal distribution. We introduce the following parameters, for any $r\neq1$.
\begin{align*}
\alpha_T=\frac{1}{1-r^T}\qquad\qquad
\zeta_T=\frac{\frac{1}{r^T}-1}{1-r}\cdot\left(\frac{1}{r}\right)^{\frac{r}{1-r}-T\alpha_T}\cdot c_0^{\alpha_T\cdot(1-r)} \qquad\qquad c_0 = \frac{1}{1-(1-p)\varepsilon}
\end{align*}
Note that $1-r<\alpha_T<1$ if $r<1$, and that $\alpha_T$ converges exponentially fast to $1$.

Algorithm~\ref{ALG} describes a procedure that tentatively opens $A^\dagger_t$ facilities in each period $t\in\calT$, with:
\begin{equation}\label{eq:solution}
    A^\dagger_t=
        r^{-\frac{1}{1-r}\left(t-T\cdot\alpha_T\cdot(1-r^t)\right)}\cdot c_0^{\alpha_T\cdot(1-r^t)}\cdot \ell^{\alpha_T(1-r^t)},
\end{equation}
where $\ell$ is the unique positive solution to the following equation:
\begin{equation} \label{eq:solution_eq}
\sum_{t=1}^T A^\dagger_t= c_0 \left(m + \varepsilon\cdot p\cdot\zeta_T\cdot m^{\alpha_T\cdot(1-r)}+ \Delta(m)\right);\   \ \Delta(m) = \begin{cases}
\sqrt{\frac{c_0}{2}\log \frac{2}{\delta}}\sqrt{m} & \text{if }p \neq 1,\\
-\Phi^{-1}(\delta) \sqrt{2\varepsilon \zeta_T} m^{\frac{1}{2}\cdot \alpha_T \cdot (1-r)}& \text{if }p = 1.
\end{cases}
\end{equation}
\begin{algorithm} [h!]
\caption{Algorithm to solve Problem~\eqref{prob:main}.}\small
\label{ALG}
\begin{algorithmic}
\item \textbf{Initialization:} $N^\dagger_0=0$.
\item Repeat, for $t=1,\cdots,T$:
\begin{itemize}
\item[] \textbf{1. Optimize.} Tentatively open $A^\dagger_t$ facilities, according to Equations~\eqref{eq:solution}--\eqref{eq:solution_eq}.
\item[] \textbf{2. Commit.} Observe $B^\dagger_t\sim \Bino\left(A^\dagger_t, 1- \frac{\varepsilon\cdot p}{\left(\sum_{s=1}^{t-1} A_s +1\right)^r}-\varepsilon\cdot(1-p)\right)$, and update $N^\dagger_t=N^\dagger_{t-1} + B^\dagger_t$.
\end{itemize}
\end{algorithmic}
\end{algorithm}

Let $m^\dagger(\varepsilon,p,T)$ be the solution of Algorithm~\ref{ALG}, and $\texttt{Reg}^\dagger(\varepsilon,p,T)=m^\dagger(\varepsilon, p,T)-\widehat{m}(\varepsilon,p)$ its regret. Theorem~\ref{thm:main} provides the main result, showing the asymptotically tight regret expression in Problem~\eqref{prob:main}, and the asymptotic feasibility and optimality of the solution from Algorithm~\ref{ALG}. The proof is outlined below, and detailed in~\ref{app:base}; we discuss the result in the next section.

\begin{theorem}\label{thm:main}
Assume that $\delta = o\left(m^{\alpha_T\cdot(1-r)-1}\right)$. The optimal regret satisfies $\texttt{Reg}^*(\varepsilon, p,T)=h(\varepsilon, p,T)$ and Algorithm~\ref{ALG} yields an asymptotically feasible solution to Problem~\eqref{prob:main} with the same regret  $\texttt{Reg}^\dagger(\varepsilon, p,T)=h(\varepsilon, p,T)$, where:
$$h(\varepsilon, p,T)=\begin{cases}
    c_0\cdot\varepsilon\cdot p\cdot\zeta_T\cdot m^{\alpha_T\cdot(1-r)}+o\left(m^{\alpha_T\cdot(1-r)}\right)&\text{if $p=1$ or if ($p\neq1$ and $\alpha_T\cdot(1-r)\geq1/2$)}\\
    \Theta(\sqrt m)&\text{if $p\neq1$ and $\alpha_T\cdot(1-r)<1/2$}
\end{cases}$$
\end{theorem}

\proof{Sketch of the proof.}

We define a deterministic approximation~\eqref{prob:main_det} of the problem where $\delta=0$ and $B_t$ always realizes at the mean value, given by:
\begin{align}\label{prob:main_det}\tag{$\calP^D$}
\min\ \left\{\ \sum_{t=1}^T A_t \ \ \bigg|\ \ \sum_{t=1}^TA_t \left(1- \frac{\varepsilon\cdot p}{\left(\sum_{s=1}^{t-1} A_s +1\right)^r}-\varepsilon\cdot(1-p)\right)\geq m\right\}.
\end{align}

Lemma~\ref{lem:main_det} elicits the asymptotic solution of Problem~\eqref{prob:main_det}. The proof leverages the Karush–Kuhn–Tucker (KKT) conditions and performs leading-order and leading-coefficient analyses to identify the asymptotic dependency on $m$ of the deterministic solution.

\begin{lemma}[Asymptotic solution of Problem~\eqref{prob:main_det}]\label{lem:main_det}
Let $m^D(\varepsilon, T)$ be the optimal objective value of Problem~\eqref{prob:main_det}, and $A^D,\cdots, A^D$ be its solution. For any $T\geq 2$, we have as $m \to \infty$: 
\begin{align}
   & A^D_t = r^{-\frac{1}{1-r}\left(t-T\cdot\alpha_T\cdot(1-r^t)\right)}\cdot c_0^{\alpha_T\cdot(1-r^t)}\cdot m^{\alpha_T\cdot(1-r^t)}+ o\left(m^{\alpha_T\cdot(1-r^t)}\right),\ \forall t=1,\cdots,T,\label{eq:deterministic_decision}\\
   &m^D(\varepsilon, T) = c_0\left( m+\varepsilon\cdot p\cdot \zeta_T \cdot m^{\alpha_T\cdot(1-r)} +  o\left(m^{\alpha_T\cdot(1-r)}\right)\right).\label{eq:deterministic_solution}
\end{align}
\end{lemma}

The solution of the deterministic problem provides insights into the structure of Algorithm~\ref{ALG}, which constructs a solution that matches Equation~\eqref{eq:deterministic_decision} with a buffer to handle uncertainty. This observation follows from the fact that $\ell$ is on the order of $\Theta(m)$, which we prove in Remark~\ref{rem:lm}.
\begin{remark}\label{rem:lm}
 For $m$ large enough, we have: $m- m^{\alpha_T \cdot (1-r)} \leq \ell\leq m + \Delta(m)$.
\end{remark}

We can then show that the solution of Algorithm~\ref{ALG} satisfies the chance constraint as $m\to\infty$, thus providing an asympotically feasible solution of the stochastic problem~\eqref{prob:main}. Algorithm~\ref{ALG} therefore yields an upper bound to Problem~\eqref{prob:main}. The proof leverages concentration inequalities across time periods $1,\cdots,T$  to bound the probability that at least $m$ facilities get successfully opened within the planning horizon, using Equations~\eqref{eq:solution}--\eqref{eq:solution_eq}. When $p\neq 1$, we leverage Hoeffding's inequality; when $p=1$, we use the Berry-Esseen theorem instead to retain asymptotically tight bounds.

\begin{lemma}[Upper bound]\label{lem:main_feasible}
The solution of Algorithm~\ref{ALG} satisfies $\lim_{m \to \infty} \P(N^\dagger_T \geq m) \geq 1-\delta$.
\end{lemma}

Finally, Lemma~\ref{lem:main_lower} establishes the lower bound on the stochastic solution in Theorem~\ref{thm:main}. When $p=1$, the proof shows that $B_t$ remains close to its expected value at each time $t=1,\cdots,T$, using Bernstein's concentration inequality across facilities; it then calls Lemma~\ref{lem:main_det} conditionally on $B_1\approx\E(B_1),\cdots,B_T\approx\E(B_T)$ to bound the total number of facilities $m^*(\varepsilon, p,T)$. In particular, Bernstein's concentration inequality exploits the small asymptotic error on facility openings, so the variance term provides a stronger bound than with Hoeffding's inequality.  When $p\neq1$, it uses the Berry-Esseen theorem to show that the lower bound is necessary to satisfy the chance constraint.

\begin{lemma}[Lower bound]\label{lem:main_lower}
Assume that $\delta = o\left(m^{\alpha_T\cdot(1-r)-1}\right)$. Let $m^*(\varepsilon, p,T)$ denote the optimal solution of Problem~\eqref{prob:main}. The following holds, as $m \to \infty$:
\[m^*(\varepsilon, p,T) \geq   \begin{cases}
c_0\left(m +\max\{-\Phi^{-1}(\delta)\sqrt{2\varepsilon(1-p)m},\varepsilon\cdot p\cdot\zeta_T\cdot m^{\alpha_T\cdot(1-r)}\}\right) & \text{if }p \neq 1\\
  c_0\left(m + \varepsilon\cdot p\cdot\zeta_T\cdot m^{\alpha_T\cdot(1-r)} +  o\left(m^{\alpha_T\cdot(1-r)}\right)\right) & \text{if }p=1
\end{cases}\]
\end{lemma}

Together, the lemmas show that the strategy detailed in Algorithm~\ref{ALG} is asymptotically optimal up to second-leading order terms, therefore completing the proof of the theorem.\hfill\Halmos

\section{Implications, applications and extensions}
\label{sec:implications}

\subsection{Discussion of results and managerial implications}
\label{subsec:insights}

We provide a theoretical and practical interpretation of Theorem~\ref{thm:main}. To guide the discussion, Table~\ref{tab:main} reports the solution of Algorithm~\ref{ALG}, the exact solution of Problem~\eqref{prob:main} (obtained via multi-dimensional grid search, and only available in small samples), and the no-learning baseline.

\begin{table}[h!]
\centering
\footnotesize\renewcommand{\arraystretch}{1.0}
\caption{Theoretical and numerical expressions of the results ($\varepsilon=0.4$, $\delta=1\%$, $r=0.5$, $p=1$).}
\label{tab:main}
\begin{tabular}{llccccc}
\toprule
Expression	&	Metric			&	$T=2$						&	$T=3$						&	$T=4	$						&	$T=5$						&	$T=6$						\\
\hline
Theory		&	No-learning regret	&	$\Theta(m)$			&	$\Theta(m)$			&	$\Theta(m)$			&	$\Theta(m)$			&	$\Theta(m)$			\\
$(m\to\infty)$	&	Online regret ($A^\dagger$)		&	$\Theta(0.75m^{0.67})$	&	$\Theta(1.0m^{0.57})$	&	$\Theta(1.2m^{0.53})$	&	$\Theta(1.4m^{0.52})$	&	$\Theta(1.5m^{0.51})$	\\
\cmidrule{2-7}
        &	Policy $A^\dagger_t$ ($t=1$)	&	$\Theta(0.63 m^{0.67})$	&	$\Theta(0.37 m^{0.57})$	&	$\Theta(0.21 m^{0.53})$	&	$\Theta(0.11 m^{0.52})$	&	$\Theta(0.06 m^{0.51})$	\\
        &	Policy $A^\dagger_t$ ($t=2$)	&	$\Theta(m)$					&	$\Theta(0.45 m^{0.86})$	&	$\Theta(0.19 m^{0.80})$	&	$\Theta(0.07 m^{0.77})$	&	$\Theta(0.03 m^{0.76})$	\\
        &	Policy $A^\dagger_t$ ($t=3$)	&	---							&	$\Theta(m)$					&	$\Theta(0.36 m^{0.93})$	&	$\Theta(0.12 m^{0.90})$	&	$\Theta(0.04 m^{0.89})$	\\
        &	Policy $A^\dagger_t$ ($t=4$)	&	---							&	---							&	$\Theta(m)$					&	$\Theta(0.31 m^{0.97})$	&	$\Theta(0.09 m^{0.95})$	\\
        &	Policy $A^\dagger_t$ ($t=5$)	&	---							&	---							&	---							&	$\Theta(m)$					&	$\Theta(0.29 m^{0.98})$	\\
        &	Policy $A^\dagger_t$ ($t=6$)	&	---							&	---							&	---							&	---							&	$\Theta(m)$					\\
\hline
Numerical	&	No-learning regret	&	93		&	93		&	93		&	93		&	93		\\
$(m=100)$	&	Online regret ($A^\dagger$)		&	30		&	28		&	27		&	28		&	28		\\
        &	Online regret ($A^*$)		&	26		&	20		&	19		&	18		&	18		\\
\cmidrule{2-7}
        &	Policy $A^\dagger_t$ ($t=1$)	&	15		&	5		&	2		&	1		&	1		\\
        &	Policy $A^\dagger_t$ ($t=2$)	&	116		&	23		&	7		&	2		&	1		\\
        &	Policy $A^\dagger_t$ ($t=3$)	&	---		&	99		&	25		&	7		&	2		\\
        &	Policy $A^\dagger_t$ ($t=4$)	&	---		&	---		&	93		&	25		&	7		\\
        &	Policy $A^\dagger_t$ ($t=5$)	&	---		&	---		&	---		&	92		&	25		\\
        &	Policy $A^\dagger_t$ ($t=6$)	&	---		&	---		&	---		&	---		&	92		\\
\cmidrule{2-7}
        &	Policy $A^*_t$ ($t=1$)	&	13		&	6		&	1		&	1		&	0		\\
        &	Policy $A^*_t$ ($t=2$)	&	113		&	25		&	7		&	3		&	1		\\
        &	Policy $A^*_t$ ($t=3$)	&	---		&	89		&	23		&	8		&	2		\\
        &	Policy $A^*_t$ ($t=4$)	&	---		&	---		&	88		&	22		&	9		\\
        &	Policy $A^*_t$ ($t=5$)	&	---		&	---		&	---		&	84		&	24		\\
        &	Policy $A^*_t$ ($t=6$)	&	---		&	---		&	---		&	---		&	82		\\
\hline
Numerical	&	No-learning regret	&	746		&	746		&	746		&	746		&	746		\\
$(m=1,000)$	&	Online regret ($A^\dagger$)		&	104		&	79		&	73		&	72		&	73		\\
        &	Online regret ($A^*$)		&	96		&	63		&	55		&	51		&	50		\\
\cmidrule{2-7}
        &	Policy $A^\dagger_t$ ($t=1$)	&	65		&	18		&	8		&	4		&	2		\\
        &	Policy $A^\dagger_t$ ($t=2$)	&	1,040	&	155		&	41		&	13		&	5		\\
        &	Policy $A^\dagger_t$ ($t=3$)	&	---		&	906		&	192		&	51		&	15		\\
        &	Policy $A^\dagger_t$ ($t=4$)	&	---		&	---		&	832		&	202		&	54		\\
        &	Policy $A^\dagger_t$ ($t=5$)	&	---		&	---		&	---		&	802		&	204		\\
        &	Policy $A^\dagger_t$ ($t=6$)	&	---		&	---		&	---		&	---		&	794		\\
\cmidrule{2-7}
        &	Policy $A^*_t$ ($t=1$)	&	60		&	22		&	5		&	3		&	1		\\
        &	Policy $A^*_t$ ($t=2$)	&	1,036	&	187		&	42		&	14		&	3		\\
        &	Policy $A^*_t$ ($t=3$)	&	---		&	854		&	212		&	57		&	14		\\
        &	Policy $A^*_t$ ($t=4$)	&	---		&	---		&	796		&	206		&	65		\\
        &	Policy $A^*_t$ ($t=5$)	&	---		&	---		&	---		&	771		&	231		\\
        &	Policy $A^*_t$ ($t=6$)	&	---		&	---		&	---		&	---		&	736		\\
\bottomrule
\end{tabular}
\end{table}

\paragraph{Asymptotically-tight bounds.}

Theorem~\ref{thm:main} derives upper-bounding and lower-bounding approximations of the optimum of Problem~\eqref{prob:main}, hence asymptotically-tight regret. When $p=1$ or $\alpha_T(1-r)\geq1/2$, we obtain matching bounds up to the second leading-order rate and coefficients; otherwise, we obtain matching bounds up to the second leading-order rate.

Whereas the asymptotic feasibility guarantee and upper bound are always valid, the lower bound is valid for a sufficiently strong chance constraint, captured by the condition $\delta = o\left(m^{\alpha_T\cdot(1-r)-1}\right)$. Algorithm~\ref{ALG} relies on a deterministic approximation with buffers to guarantee asymptotic feasibility (Lemma~\ref{lem:main_feasible}) but the second leading-order term is independent on $\delta$. In contrast, a large value of $\delta$ alleviates the impact of the chance constraint, making it harder to control the stochastic solution and leading to larger differences with the deterministic approximation. Thus, the condition ensures that the chance constraint remains significant enough in the stochastic problem~\eqref{prob:main}. Practically, this restriction is rather weak, ranging, with $r=0.5$, from $\delta=o\left(1/m^{1/3}\right)$ with $T=2$ to $\delta=o\left(1/\sqrt m\right)$ as $T\to\infty$. These conditions translate into feasibility guarantees of 80--90\% with $m=100$ and 90--97\% with $m=1,000$ facilities, consistent with typical reliability targets in practice.

\paragraph{Sub-linear regret: benefits of online learning.}

The solution to Problem~\eqref{prob:main} achieves sub-linear asymptotic regret, thus achieving an intermediate outcome between the fully-learned solution (zero regret, by definition) and the no-learning solution (linear regret, per Proposition~\ref{prop:regret_NL}). This result highlights the performance improvements of online learning and optimization. With a learning rate of $\calO(1/\sqrt n)$, the regret is $\Theta\left(m^{0.67}\right)$ with $T=2$ down to $\Theta\left(\sqrt m\right)$ as $T\to\infty$. These benefits translate into significant cost savings, estimated at 30--40\% in Table~\ref{tab:main} with $m=100$ or $m=1,000$.

\paragraph{Exponential convergence: fast learning.}

The sub-linear regret holds in an asymptotic regime with an infinite number of facilities but any finite planning horizon. When $\frac{1-r}{1-r^{T}}<1/2$ and $r<1$, the regret rate is time-independent, in $\Theta(\sqrt{m})$. Even with a smaller learning rate, the regret rate $m^{\frac{1-r}{1-r^{T}}}$ decreases exponentially fast as the horizon grows longer, down to the $\Theta(m^{1-r})$ limit as $T\to\infty$. With $r=0.5$, this translates into $\Theta(m^{0.67})$ with $T=2$, in $\Theta(m^{0.57})$ with $T=3$, and in $\Theta(m^{0.53})$ with $T=4$. We therefore obtain a regret close to the $\Theta(\sqrt{m})$ limit with as few as 3-5 iterations.

This finding underscores the benefits of even \textit{limited} online learning and optimization. Indeed, even a few stages of learning can yield significant gains, by inducing a sub-linear regret of $\Theta(m^{0.53})$ to $\Theta(m^{0.67})$ with 2--4 iterations of online learning, as opposed to a linear regret of $\Theta(m)$ under the no-learning baseline. Conversely, more iterations bring more marginal benefits, by reducing the regret rate from $\Theta(m^{0.53})$ with 4 iterations down to $\Theta(\sqrt{m})$ in an infinite horizon.

\paragraph{Robustness.}

The findings in Theorem~\ref{thm:main} are robust across a wide range of online learning regimes characterized by any initial error $\varepsilon$, residual error $1-p$, and learning rate  $r$. As expected, the regret becomes smaller as $\varepsilon$ decreases (i.e., lower uncertainty), as $p$ increases (i.e., smaller irreducible error), and as $r$ increases (i.e., faster learning). With perfect learning ($p=1$), the long-term regret converges to $\Theta\left(m^{1-r}\right)$; with imperfect learning ($p<1$), it converges to $\Theta\left(m^{1-r}\right)$ if $r\geq0.5$ but to $\Theta(\sqrt m)$ otherwise. These results underscore that, if $p=1$ (perfect learning) or $\alpha_T(1-r)\geq1/2$ (e.g., slow learning, $r\leq0.5$), the main driver of regret lies in the learning rate; when $p<1$ (imperfect learning) and $\alpha_T(1-r)<1/2$ (fast learning, $r>0.5$), the main bottleneck lies in the residual error.

\begin{figure}[h!]
\centering
\subfloat[Dependency on learning rate.]{\label{fig:learning_rate_r}\includegraphics[width=.49\textwidth]{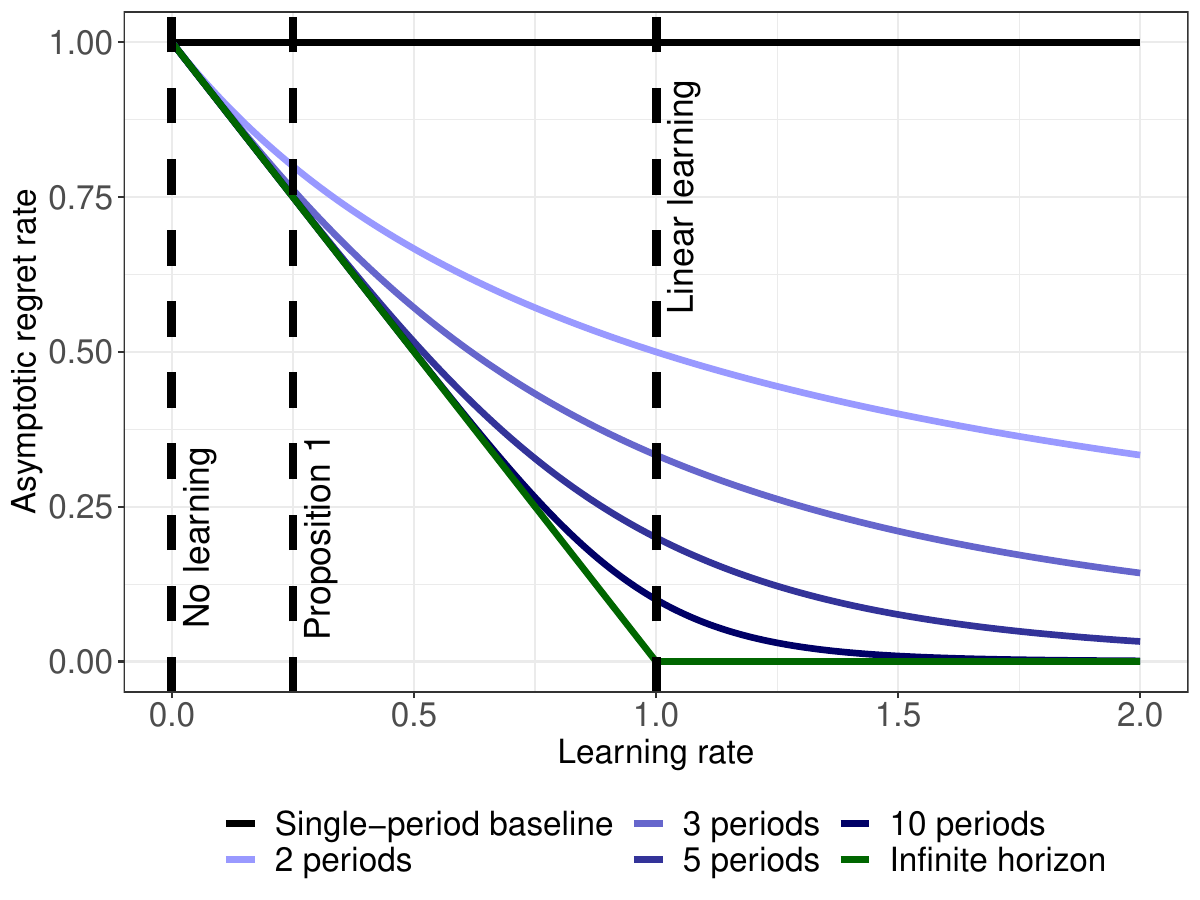}}
\hspace{2pt} 
\subfloat[Dependency on time horizon.]{\label{fig:learning_rate_T}\includegraphics[width=.49\textwidth]{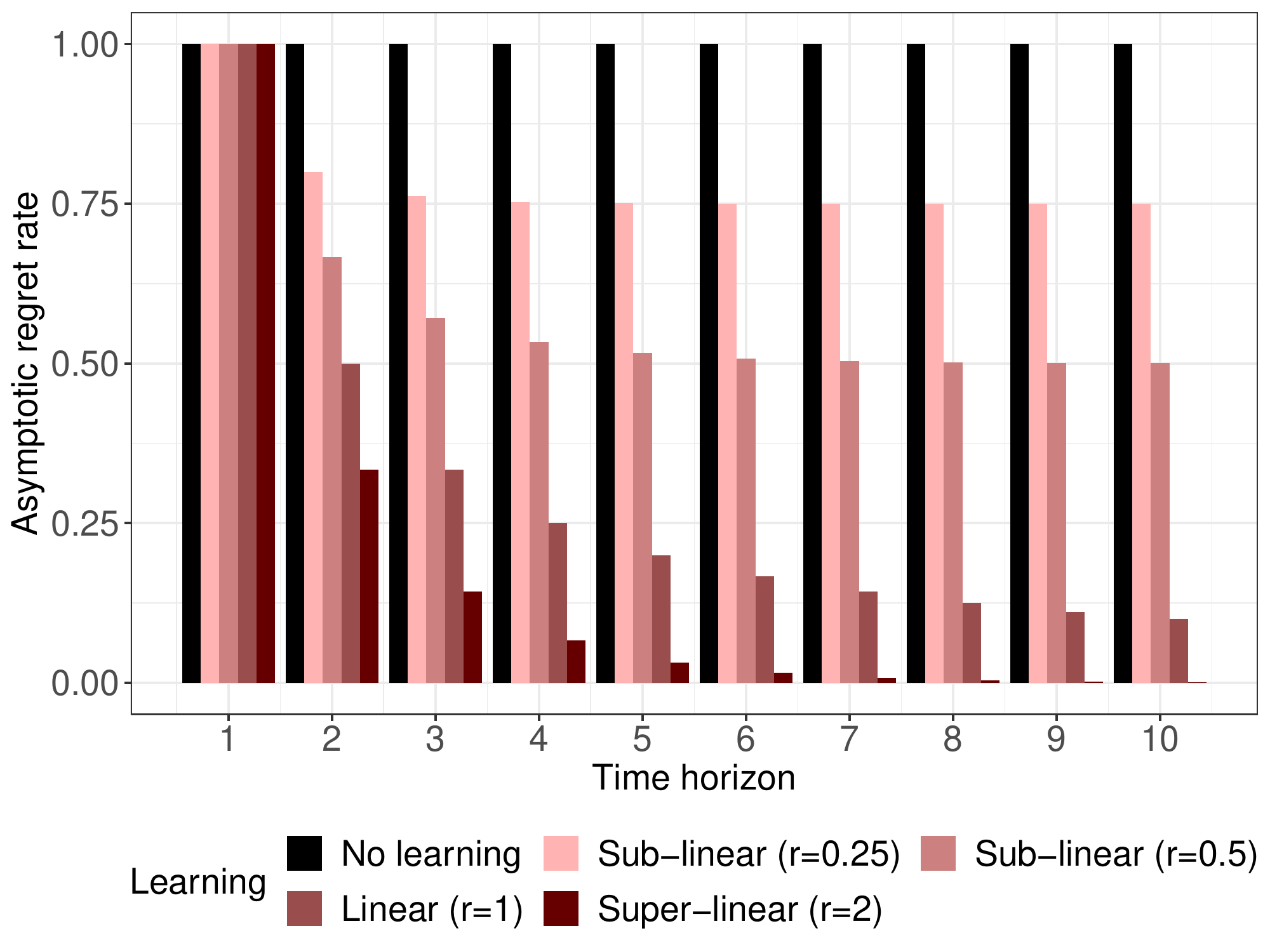}}
\caption{Impact of learning rate and time horizon on asymptotic regret ($p=1$).}
\label{fig:learning_rate}
\vspace{-12pt}
\end{figure}

Figure~\ref{fig:learning_rate} illustrates the impact of the learning rate for $p=1$. As $T$ goes to infinity (green line in Figure~\ref{fig:learning_rate_r}), the asymptotic regret converges to $1-r$ when $r<1$ and 0 when $r\geq 1$. In particular, when $r\geq1$, the algorithm achieves \textit{bounded regret} in the double-asymptotic regime where $m$ and $T$ grow at comparable rates (e.g., $m=\rho^a$ and $T=\rho^b$ with $a,b>1$). Also, the regret undergoes a phase transition in $r=1$, moving from exponential convergence when $r\neq 1$ to linear convergence when $r=1$ back to exponential convergence  when $r>1$ (Figure~\ref{fig:learning_rate_T}). When $p<1$, similar regret rates are obtained except that they are capped to 0.5 per Theorem~\ref{thm:main}. Most importantly, the figure underscores the robust benefits of online learning and optimization (sub-linear regret) against the no-learning baseline (linear regret), even with slow learning and short planning horizons.

\paragraph{An asymptotically optimal algorithm.}

The constructive proof of Theorem~\ref{thm:main} yields a simple and interpretable algorithm that attempts to open $\Theta\left(m^{\alpha_T(1-r^t)}\right)$ facilities in each period $t=1,\cdots,T$. Thus, a small number of facilities get tentatively opened in initial iterations, whereas the bulk of facilities get opened later on once uncertainty gets partially mitigated. For example, with $r=0.5$ and $T=2$, the decision-maker attempts to opens $\Theta(m^{2/3})$ facilities in the first period and $\Theta(m)$ facilities in the second period; when $T=3$, the sequence is $\Theta(m^{0.57})$, $\Theta(m^{0.86})$, and  $\Theta(m)$ facilities. As $T$ grows to $\infty$, the first-period decision is $\Theta\left(\frac{m^{1-r}}{(1/r)^{T-1/(1-r)}}\right)$. Thus, it is sufficient to consider a logarithmic horizon to achieve the full benefits of an infinite-horizon online learning algorithm (otherwise, the first-period decision involves not even building one facility). This observation further highlights the exponentially fast convergence and the benefits of even limited online learning.

In addition, the solution of Algorithm~\ref{ALG} closely mirrors the optimal solution to Problem~\eqref{prob:main}. With a target of $m=100$, the solution lies within 3--8\% of optimality; with a larger target of $m=1,000$, the gap goes down further to 1--2\%, which is consistent with its theoretical asymptotic guarantees. Moreover, both the solution $A^\dagger_1,\cdots,A^\dagger_T$ and the optimal solution $A^*_1,\cdots,A^*_T$ feature limited exploration early on and fast exploitation later on. The main difference between the two solutions lies in the final-period decision, which is inflated with a buffer of $\Delta(m)$ in Algorithm~\ref{ALG} to guarantee asymptotic feasibility in view of the chance constraint; we address this limitation in Section~\ref{subsec:robustness}. These results, however, show that all qualitative insights on the algorithm's solution do not rely on idiosyncracies of our algorithm but on the very structure of Problem~\eqref{prob:main}.

\paragraph{Managerial insights.}
These findings suggest to run a pilot program toward strategic planning under uncertainty in order to acquire a relatively small sample of online data prior to full-fledged expansion. Early iterations involve limited exploration to gather information on successful vs. unsuccessful facilities. In later iterations, the strategy shifts to fast exploitation, using the latest machine learning models to open most facilities and meet the coverage target. The rapid convergence in regret rates suggests that high-quality outcomes can be achieved in just a few iterations over a limited planning horizon, thereby realizing the benefits of online learning and optimization even when information gathering is time-consuming and coverage targets must be met rapidly.

\subsection{Application to real-world datasets}
\label{subsec:UCI}

We apply our online learning and optimization approach to four datasets from the \cite{UCI} featuring binary classification with structured multivariate data, 10--100 features, and over 1,000 records: ``bank marketing'', ``default of credit card clients'', ``occupancy detection'' and ``online shoppers purchasing intention''. We report in~\ref{app:UCI} descriptive statistics and numerical evidence of the decay in predictive error with the sample size, thereby providing empirical justification of the learning function characterized in Assumption~\ref{ass:classification}.

For each dataset, we treat each data point as a candidate ``facility'' in the set $\calF$ and we impose a target of $m=M/10$ over $T=3$ time periods. This target is large enough while maintaining a large-enough candidate pool (Assumption~\ref{ass:regime}). At each iteration, we train a random forest model based on the past samples; we select items among the sub-population with positive predictions (Equation~\eqref{eq:exploration} with $\calE_t=\emptyset$); and we observe success realizations based on the ground-truth labels. We compare the solution to the no-learning baseline that selects samples uniformly at random until $m$ positive samples are obtained. For each dataset, we replicate each simulation 100 times.

\begin{figure}[h!]
\centering
\includegraphics[width=.7\textwidth]{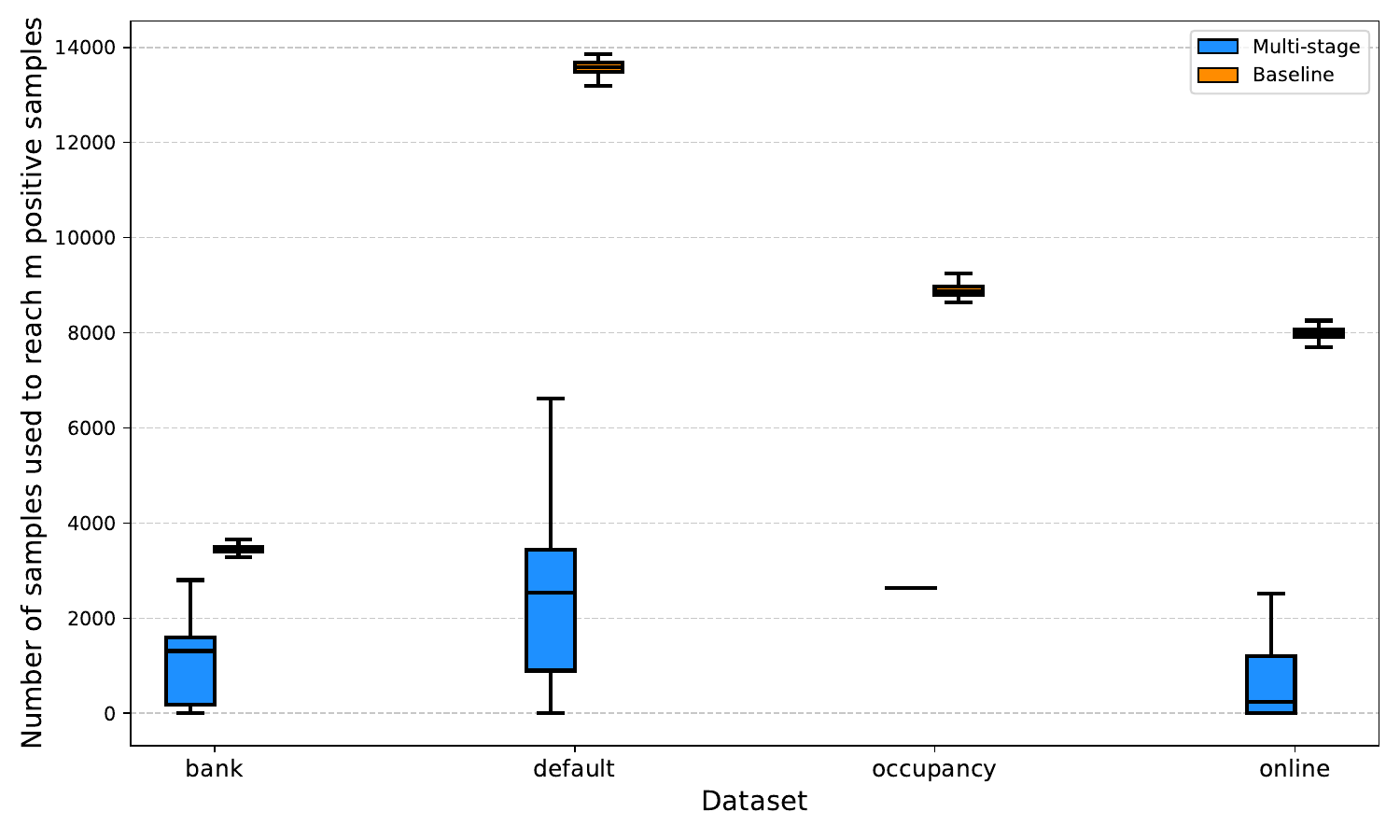}
\caption{Number of samples required to identify $m=n/10$ positive samples across datasets ($\delta=5\%$, $T=3$).}
\label{fig:ML_real}
\end{figure}

Figure~\ref{fig:ML_real} shows that the online learning and optimization procedure significantly outperforms the no-learning baseline, requiring far fewer samples to achieve the same coverage target---even over three iterations. It reduces the expected cost by over 50\% across all datasets, and by an order of magnitude for the ``online'' dataset. These benefits are robust across simulations. In particular, the ``occupancy'' dataset exhibits stable performance because it leads to uniformly strong predictability under small samples (Figure~\ref{fig:ML}). The results provide empirical evidence of the performance of our online learning and optimization approach on finite samples and real-world data.

\subsection{Robustness to the online learning and optimization environment}
\label{subsec:robustness}

In our main setting, data are \textit{only} available online and decisions are made all at once at the beginning of the planning horizon. In this section, we extend the problem in the presence of offline data and in an adaptive environment.~\ref{app:robustness} reports all formulations, algorithms and proofs. For conciseness, we assume that $p=1$ or $\alpha_T(1-r)\geq1/2$ but all arguments can be extended.

\subsubsection*{Offline data.}

We assume that the decision-maker has access to $N_0=\gamma m^\beta+o\left(m^\beta\right)$ offline data points, with $\gamma>0$ and $0\leq\beta\leq1$. Theorem~\ref{thm:offline} characterizes the regret in this environment.

\begin{theorem}\label{thm:offline}
Let $\gamma>0$ and $0\leq\beta\leq1$. Assume that $\delta = o\left(m^{(1-\beta)\cdot\alpha_T\cdot(1-r)+\beta(1-r)-1}\right)$ and $p=1$ or $\alpha_T(1-r)\geq1/2$. With $N_0=\gamma m^\beta+o\left(\gamma m^\beta\right)$ offline data points, the stochastic solution and the algorithm achieve asymptotically tight regret equal to $c_0 \cdot \varepsilon\cdot p\cdot\zeta_T\cdot \gamma^{-\alpha_T(1-r)}\cdot m^{(1-\beta)\cdot\alpha_T\cdot(1-r)+\beta(1-r)}$.
\end{theorem}

As expected, the regret decreases with more offline data points, ranging from $\Theta\left(m^{\alpha_T\cdot(1-r)}\right)$ with no offline data to the limit of $\Theta\left(m^{1-r}\right)$ with $\Theta(m)$ data points. The reliability requirement becomes slightly tighter, in $\delta = o(m^{(1-\beta)\cdot \alpha_T(1-r) + \beta(1-r) - 1})$ rather than $\delta = o(m^{\alpha_T(1-r) - 1})$. This is because larger data samples make it comparatively easier to reach the coverage target, so a looser chance constraint makes the stochastic solution comparatively stronger.

We also extend this setting to capture instances where the decision-maker can acquire offline data at cost $c\leq1$. We add an initial decision $A_0$ corresponding to the number of offline data points, so the objective function becomes $\min\ cA_0+\sum_{t=1}^T A_t$. Theorem~\ref{thm:warmstart} shows that the algorithm achieves an asymptotically optimal regret of $\Theta\left(m^{\alpha_{T+1}\cdot(1-r)}\right)$.

\begin{theorem}\label{thm:warmstart}
Assume that $\delta = o\left(m^{\alpha_{T+1}\cdot(1-r)-1}\right)$ and $p=1$ or $\alpha_T(1-r)\geq1/2$. With offline data acquisition at cost $c$, the stochastic solution and the algorithm achieve asymptotically tight regret, equal to $\left(c_0\cdot\varepsilon\cdot p\cdot\zeta_T\cdot c^{\alpha_T(1-r)}\cdot\alpha_T(1-r)\right)^{\frac{1}{\alpha_T(1-r)+1}}\cdot \left(1+\frac{1}{\alpha_T(1-r)}\right)m^{\alpha_{T+1}\cdot(1-r)}$.
\end{theorem}

This result suggests that offline learning has a limited impact, in that the regret rate with offline learning and $T$ online periods is identical to the regret rate with $T+1$ online periods. Obviously, these results only characterize the second-leading asymptotic order; in fact, the second-leading coefficient suggests that the benefits become larger as the cost $c$ of offline data acquisition gets smaller. Moreover, the benefits of offline learning may be more significant in finite samples. Still, this result reinforces the impact of online learning and optimization toward meeting the target.

\subsubsection*{Adaptive decision-making.}

Rather than relying on the static solution to Problem~\eqref{prob:main}, the decision-maker may adjust facility openings dynamically---for example, by opening fewer facilities later on if more facilities are successful than expected early on, and vice versa. The full adaptive optimization problem is formulated via dynamic programming in~\ref{app:robustness}, as a function of past facility openings and success realizations. However, this formulation is hard to analyze analytically.

We consider an adaptive re-optimization procedure that solves a variant of Problem~\eqref{prob:main} at each time period (see Algorithm~\ref{alg:adaptive} in~\ref{app:robustness}). In the first period, Problem~\eqref{prob:main} yields solution $A^\dagger_1,\cdots,A^\dagger_T$, but the procedure merely implements the first decision $A^\dagger_1$. Upon observing the realization $B_1$, it solves a corresponding variant to open $m-B_1$ facilities over a shrunken horizon of $T-1$ periods; it implements the immediate decision and proceeds. Thus, at each period $t=1,\cdots,T-1$, the decision-maker solves~$\left(\calQ^*_{T-t+1}\left(m-\sum_{s=1}^{t-1}B_s,\sum_{s=1}^{t-1} A_s\right)\right)$, where $\calQ^*_T(m,N_0)$ refers to the variant of Problem~\eqref{prob:main} with a target of $m$ facilities, a horizon of $T$ periods, and $N_0$ offline data points. In the last period $T$, the decision-maker solves a single-stage problem with a target of $m-\sum_{s=1}^{T-1}B_s$ and an initial sample size of $\sum_{s=1}^{T-1}A_s$, referred to as~$\left(\calS^*\left(m-\sum_{s=1}^{T-1}B_s,\sum_{s=1}^{T-1} A_s\right)\right)$.

Theorem~\ref{thm:adaptive} elicits the adaptive solution and its regret over the first $T-1$ periods.
\begin{theorem}\label{thm:adaptive}
Assume that $\delta = o(m^{\alpha_T(1-r)-1})$ and $p=1$ or $\alpha_T(1-r)\geq1/2$. For $t=1,\cdots,T-1$, the adaptive solution opens $A^{\textsc{ad}}_t=\Theta(m^{\alpha_T(1-r^t)})$ facilities with a regret of $\Theta\left(m^{\alpha_T\cdot(1-r)}\right)$ as $m \to \infty$.
\end{theorem}

This result holds with the same reliability level $\delta = o(m^{\alpha_T(1-r) - 1})$ as in Problem~\eqref{prob:main}. In each time period, the data sample gets larger, requiring smaller value of $\delta$, but the remaining horizon gets smaller, allowing larger values of $\delta$. We find that these two effects cancel each other out.

Theorem~\ref{thm:adaptive} shows that the adaptive solution mirrors the solution of Algorithm~\ref{ALG} for $T-1$ periods, with the same regret rate of $\Theta\left(m^{\alpha_T\cdot(1-r)}\right)$. The benefits of adaptiveness are therefore surprisingly limited as compared to the static solution of Problem~\eqref{prob:main}. Again, this result is confined to the second-leading order in the asymptotic regret rate, and it focuses on the impact of re-optimization rather than the optimal dynamic programming policy. Still, it reinforces the strong performance of the static solution to Problem~\eqref{prob:main} in the asymptotic regime with a large coverage target.

Theorem~\ref{thm:adaptive} also highlights the earlier observation that the main limitation of Algorithm~\ref{ALG} lies in the $\Delta(m)$ buffer designed to guarantee asymptotic feasibility. Indeed, the main difference between the static and adaptive solutions is that the latter adjusts the final-period decision to exactly meet the chance constraint. We therefore propose a simple ``semi-adaptive'' procedure that relies on the static solution for the first $T-1$ periods and then implements the same final-period adjustment. Specifically, the procedure implements $A^\dagger_1,\cdots,A^\dagger_{T-1}$ from Algorithm~\ref{ALG}, and then solves Problem $\left(\calS^*\left(m-\sum_{s=1}^{T-1}B_s,\sum_{s=1}^{T-1} A_s\right)\right)$ to avoid wastage from the initial buffer (see Algorithm~\ref{alg:adaptive_simple} in~\ref{app:robustness}).

\begin{figure}[h!]
\centering
\subfloat[Average number of facilities tentatively opened]{\label{subfig:adaptive_policy}\includegraphics[width=.9\textwidth]{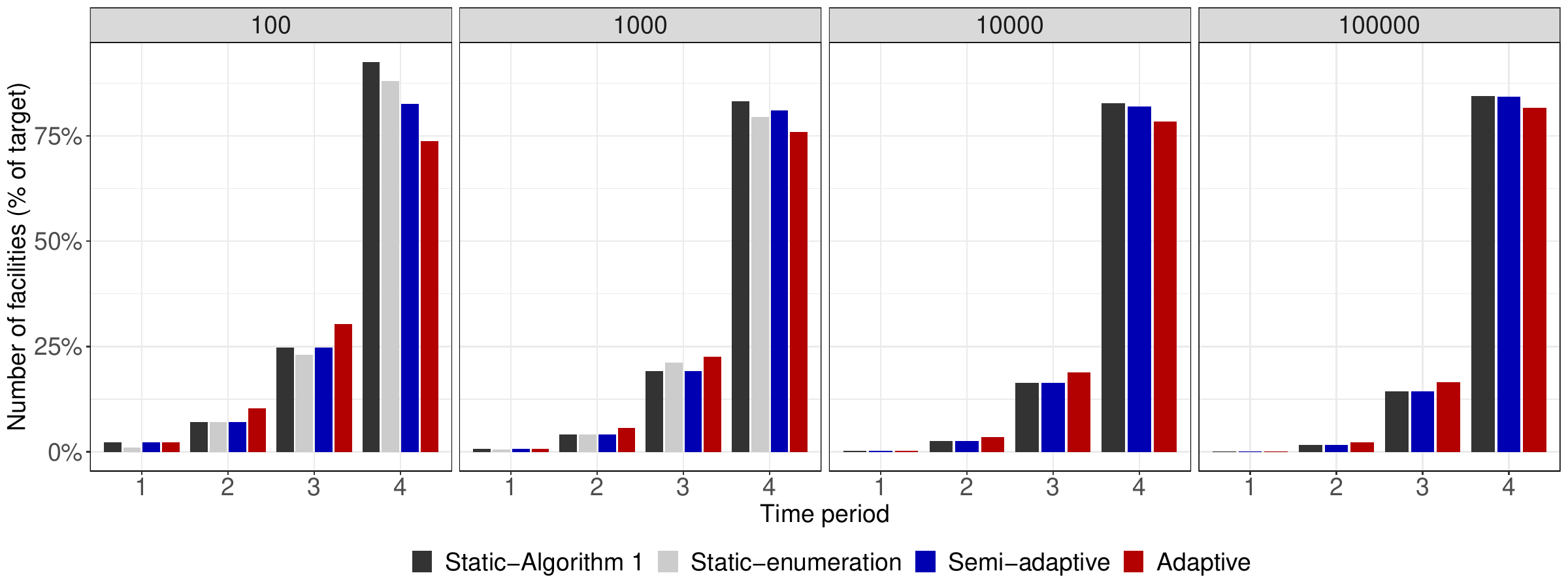}}
\hspace{0.1 cm}
\subfloat[Average regret]{\label{subfig:adaptive_regret}\includegraphics[width=.9\textwidth]{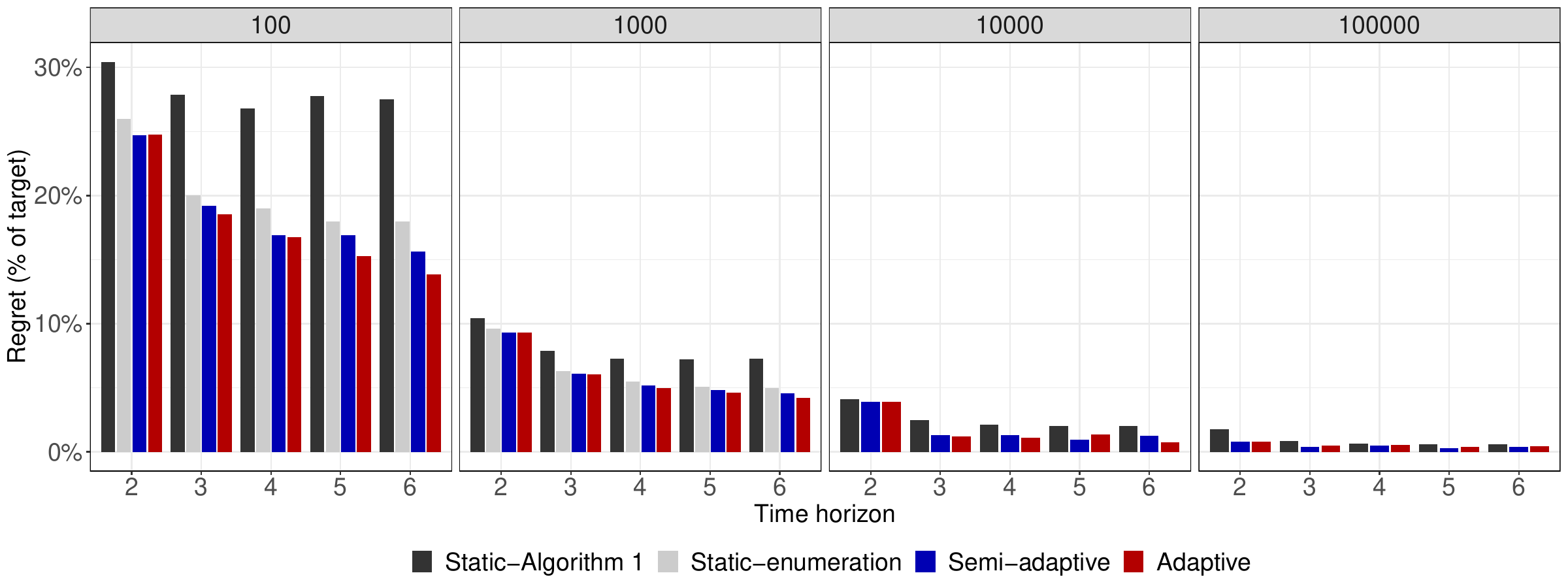}}
\caption{Comparison of static, semi-adaptive adjustment, and adaptive algorithms.}
\label{fig:adaptive}
\vspace{-12pt}
\end{figure}

Figure~\ref{fig:adaptive} compares (i) the solution of Algorithm~\ref{ALG}; (ii) the exact solution of Problem~\eqref{prob:main}, obtained via enumeration in small-scale instances ($m=100$ and $m=1,000$); (iii) the semi-adaptive solution; and (iv) the adaptive solution. The former two coincide with those reported in Table~\ref{tab:main} with $m=100$ and $m=1,000$, and the latter two come from the adaptive adjustments.

Note the large benefits of the final-period adjustment in the semi-adaptive algorithm. For instance, with $T=4$, it reduces the regret from 32 to 17 with a target of 100 facilities, and from 73 to 52 with a target of 1,000 facilities. In fact, the semi-adaptive solution outperforms the optimal solution of the static problem $(\calP^*)$. Given our previous observation that most of the approximation error in Algorithm~\ref{ALG} stems from the $\Delta(m)$ buffer, these results identify the final-period adjustment as a simple-to-implement, yet effective approach to satisfy the chance constraints based on past realizations---thus enhancing the performance of the online learning and optimization approach.

The adaptive algorithm can provide additional, but small, benefits; for instance, with $T=4$ and $m=100$, it reduces the regret from 52 to 50. In fact, it even achieves a worse regret in a few instances. One reason is that the algorithms to solve Problems~\eqref{prob:main} and~\ref{prob:QtmN} are designed to be asymptotically-optimal but do not guarantee near-optimality of early decisions, which are precisely the ones implemented in the adaptive procedure. Most importantly, the limited benefits of adaptiveness stem from limited exploration in early time periods; for instance, the first-period solution with $T=4$ and $m=1,000$ selects 8 facilities, so the exact number of successful ones (say, 5, 6 or 7 facilities) does not markedly impact the remaining target of $1,000$ facilities.

Altogether, these results confirm the strength of the static formulation in Problem~\eqref{prob:main}, as well as the effectiveness of the semi-adaptive solution in the dynamic decision-making environment. Our theoretical and numerical results underscore that Algorithm~\ref{ALG}|and its semi-adaptive extension in Algorithm~\ref{alg:adaptive_simple}|provide a simple, interpretable, yet effective approach to the online learning and optimization problem. Practically, this approach offers the advantage of providing an overall planning solution for $T-1$ periods \textit{at the beginning of the horizon}; in later stages, simple adjustments can be implemented to \textit{exactly} meet the coverage target based on latest available information.

\section{Networked Environment: Target on Customer Coverage}
\label{sec:network}

We extend our problem to heterogeneous facilities serving different customers, with a target on customer coverage. Network dependencies complicate the analysis but we prove a theorem that extends our main result. Again, a constructive proof identifies an asymptotically-optimal algorithm.

\subsection{Problem formulation}
\label{subsec:network_problem}

We formalize facility-customer interactions in a bipartite graph $\calG=(\calF,\calC, \calA)$, where $\calF=\{1, \cdots, n\}$ stores $n$ candidate facilities, $\calC=\{1, \cdots, q\}$ stores $q$ customers, and $\calA$ stores undirected edges connecting facilities and customers. We denote by $\calC_i=\left\{j\in\calC\mid(i,j)\in\calA\right\}$ the set of customers covered by facility $i\in\calF$, and by $\calF_j=\left\{i\in\calF\mid(i,j)\in\calA\right\}$ the set of facilities covering customer $j\in\calC$. The problem minimizes the number of facility openings given a target of $m$ covered customers. Our core model in Section~\ref{sec:base} is a special case with $|\calC_i|=1$ for all $i$ and $|\calF_j|=1$ for all $j$. The decision-maker optimizes the set of facilities to open via the disaggregated $y_{it}$ and $z_{it}$ variables. Moreover, we define the following variables to track customer coverage from successful facilities.
\begin{align*}
s_{jt}&=\begin{cases}
    1&\text{if customer $j \in \calC$ is served at time $t\in\calT$, i.e., if there exists $i\in\calF_j$ such that $z_{it}=1$}\\
    0&\text{otherwise.}
\end{cases}
\end{align*}

We define $g(r)$ as the regret rate from Section~\ref{sec:base}, which we extend in this section.
\[g(r) = \begin{cases}
\alpha_T\cdot (1-r) & \text{if $p=1$ or if ($p\neq1$ and $\alpha_T\cdot(1-r)\geq1/2$)}\\
1/2 & \text{if $p\neq1$ and $\alpha_T\cdot(1-r)<1/2$}
\end{cases}\] 

We introduce a degree assumption to retain a similar asymptotic regime. Definition~\ref{def:central} identifies central facilities connected to many customers, and central customers connected to central facilities. Assumption~\ref{ass:degree} states that the number of central customers grows in $\calO\left(m^{g(r)}\right)$, that the central facilities have degree $o\left(m^{g(r)}\right)$, and that the degree of each other node is bounded by~$d_\star$.

\begin{definition}\label{def:central}
Let $\deg(\cdot)$ denote the degree function in $\calG$. For $d>0$, we define central facilities $\calF^{\textsc{ctl}}(d)=\{i\in\calF \mid d< \deg(i)\}$ and central customers $\calC^{\textsc{ctl}}(d)= \{j\in\calC \mid \calF^{\textsc{ctl}}(d)\cap\calF_j\neq\emptyset\}$.
\end{definition}

\begin{assumption}[Degree]\label{ass:degree}
There exist $d_\star$ and $k <g(r)$ such that: (i) $|\calC^{\textsc{ctl}}(d_\star)|\leq \calO(m^k)$; (ii) $\deg(j)\leq m^k,\ \forall j\in\calF^{\textsc{ctl}}(d_\star)$; (iii) $\deg(j)\leq d_\star \;\;\forall j \in \calF\setminus\calF^{\textsc{ctl}}(d_\star)$; and (iv) $\deg(i)\leq d_\star \;\;\forall i \in \calC$.
\end{assumption}

Under this assumption, high-degree facilities cover a small fraction of customers, so that ``many'' customers remain to be covered even if all central facilities get successfully opened. Since every remaining facility has bounded degree, this assumption ensures that a large number of facilities will need to be opened to meet the customer coverage target, which enables us to leverage results obtained with an asymptotically large number of facilities in Section~\ref{sec:base}. In practice, this assumption corresponds to distributed settings where most customers need to be served by ``local'' facilities. In our vaccination example, most clinics cater to a limited local population; similarly, in our humanitarian logistics example, most shelters serve a limited number of displaced people. The assumption also captures broader capacitated facility location problems, which also require large numbers of facilities to satisfy customer coverage requirements.

We consider a similar learning environment as in Section~\ref{sec:setting}. Per Assumption~\ref{ass:classification}, the probability of failure of any selected facility satisfies:
\begin{equation}\label{eq:Bernoulli}
\P(z_{it}=0 \mid y_{it}=1)=\frac{\varepsilon\cdot p}{\left(\sum_{s=1}^{\tau-1} \sum_{i=1}^n y_{is} + 1\right)^r}+\varepsilon\cdot(1-p)
\end{equation}

Thus, the coverage of customer $j\in\calC$ follows a Bernoulli variable, equal to 1 if and only if at least one facility in $\calF_j$ gets successfully opened. The online learning and optimization problem, referred to as~\eqref{prob:network}, can be formulated as follows. We still denote its optimal solution by $m^*(\varepsilon, p,T)$.
\begin{align}
m^*(\varepsilon,p,T)=\min_{\by,\bs} \quad & \sum_{t=1}^T \sum_{i=1}^n y_{it} \label{prob:network}\tag{$\calP_N^*$}\\
\st\quad&\P\left[\sum_{j=1}^q s_{jT} \geq m\right] \geq 1-\delta \nonumber\\
\quad& s_{jT}\sim \Ber\left(1-\prod_{\tau=1}^T \left(\frac{\varepsilon\cdot p}{\left(\sum_{s=1}^{\tau-1} \sum_{i=1}^n y_{is} + 1\right)^r}+\varepsilon\cdot(1-p)\right)^{\sum_{i \in \calF_j} y_{i\tau}}\right),\ \forall j\in\calC\ \nonumber\\
\quad& y_{it}\in\{0,1\},\ \forall i\in \calF, \ \forall t\in\calT\nonumber
\end{align}

This problem is highly intractable, by combining the challenges of discrete optimization and stochastic non-linear optimization. Proceeding as in the core setting, we define a deterministic-approximation algorithm and establish its asymptotic optimality in Section~\ref{subsec:network_ALG}.

\subsubsection*{Benchmarks.}
We define the following two benchmarks, echoing the core setting in Section~\ref{sec:base}:
\begin{itemize}
\item[--] Fully-learned benchmark. With irreducible error ($p<1$), this benchmark is given by a similar integer non-linear optimization problem as in Problem~\eqref{prob:network}, which makes it highly challenging to study analytically. Therefore, we only define the fully-learned benchmark under perfect learning ($p=1$). In that case, the benchmark is formulated as the following deterministic facility location problem, referred to as Problem~\eqref{prob:network_perfect}. We denote by $\widehat{m}(\varepsilon)$ its optimum.
\begin{align}
    \widehat{m}(\varepsilon)=\min \quad & \sum_{t=1}^T \sum_{i=1}^n y_{it} \label{prob:network_perfect} \tag{$\widehat\calP_N$}\\
    \st\quad& \sum_{j=1}^q s_{jT} \geq  m \nonumber\\
    \quad&z_{it}=z_{i,t-1} + y_{it} \times S_i,\ \forall i\in\calF,\ \forall t\in\calT\nonumber\\
    & s_{jt} \geq z_{it} ,\ \forall j\in\calC,\ \forall i\in \calF_j, \ \forall t\in\calT\nonumber\\
    \quad& z_{i0}=0,\ \forall i\in\calF\nonumber\\
    \quad& y_{it}\in\{0,1\},\ z_{it}\in\{0,1\},\ \forall i\in \calF, \ \forall t\in\calT\nonumber\\
    \quad& s_{jt}\in\{0,1\},\ \forall j\in \calC, \ \forall t\in\calT\nonumber
\end{align}
\item[--] No-learning baseline $m^{\text{NL}}(\varepsilon)$, which solves Problem~\eqref{prob:network} with one period and error $\varepsilon$. It is given as follows, and we denote by $m^{\text{NL}}(\varepsilon)$ its optimal solution:
\begin{align}
    m^{\text{NL}}(\varepsilon)=\min \quad & \sum_{t=1}^T \sum_{i=1}^n y_{it} \label{prob:network_nolearn}\tag{$\calP_N^{NL}$}\\
    \st\quad&\P\left[\sum_{j=1}^q s_{jT} \geq m\right] \geq 1-\delta \nonumber\\
    \quad& s_{jT}\sim \Ber\left(1-\prod_{\tau=1}^T \varepsilon^{\sum_{i \in \calF_j} y_{i\tau}}\right),\quad \forall j\in\calC\ \nonumber\\
    \quad& y_{it}\in\{0,1\},\ \forall i\in \calF, \ \forall t\in\calT\nonumber
\end{align}
\end{itemize}

As the fully-learned benchmark is only available under perfect learning ($p=1$), so is any notion of regret. In that case, we omit the dependency on $p$, and we define the optimal regret as $\texttt{Reg}^*(\varepsilon,T)=m^*(\varepsilon,1,T)-\widehat{m}(\varepsilon)$ and the no-learning regret as $\texttt{Reg}^{\text{NL}}(\varepsilon)=m^{\text{NL}}(\varepsilon)-\widehat{m}(\varepsilon)$.

\subsection{An algorithm and sub-linear regret}
\label{subsec:network_ALG}

We describe our approach to solving Problem~\eqref{prob:network} in Algorithm~\ref{ALG_network}, following similar principles as in Algorithm~\ref{ALG}. Specifically, the algorithm solves a deterministic approximation of Problem~\eqref{prob:network}, with the following buffer to handle the chance constraints:
\begin{equation*}
\Delta_G(m) = \begin{cases}
    -\Phi^{-1}\left(\frac{\delta}{2}\right)d_\star^2\sqrt{2\varepsilon c_0 m}& \text{if }p\neq1,\\
        -\Phi^{-1}\left(\frac{\delta}{2}\right)d_\star^2\sqrt{\varepsilon \zeta_T m^{\alpha_T \cdot (1-r)}}& \text{if }p =1.
\end{cases}
\end{equation*}
\begin{algorithm} [h!]
\caption{Algorithm to solve Problem~\eqref{prob:network}.}\small
\label{ALG_network}
\begin{algorithmic}
\item \textbf{Initialization:} $z^\dagger_{i0}=0,\ \forall i\in\calF$.
\item \textbf{Optimize.} Solve the following optimization problem. Record solution $y^\dagger_{it}$ for $i\in\calF$ and $t=1,\cdots, T$.
\begin{align}
\min \quad & \sum_{t=1}^T \sum_{i=1}^n y_{it} \label{prob:network_ALG}\tag{$\calP_N^\dagger$}\\
\quad& \sum_{j=1}^q \left(1-\prod_{\tau=1}^T \left(\frac{\varepsilon\cdot p }{\left(\sum_{s=1}^{\tau-1} \sum_{i=1}^n y_{is} +1\right)^r}+\varepsilon\cdot(1-p)\right)^{\sum_{i \in \calF_j} y_{i\tau}}\right) \geq m+\Delta_G(m),\nonumber\\
\quad& y_{it}\in\{0,1\},\ \forall i\in \calF, \ \forall t\in\calT\nonumber
\end{align}
\item Repeat, for $t=1,\cdots,T$:
\begin{itemize}
\item[] \textbf{Commit.} Observe $z_{it}\sim \Ber\left(1-\frac{\varepsilon\cdot p}{\left(\sum_{s=1}^{\tau-1} \sum_{i=1}^n y_{is} +1\right)^r}-\varepsilon\cdot(1-p)\right)$ for all $i\in\calF$ such that $y^\dagger_{it}=1$.
\item[] \textbf{Serve.} Update variables $s_{jt}=1$ for all $j\in\calC_i$, $i\in\calF$ such that $y^\dagger_{it}=1$.
\end{itemize}
\end{algorithmic}
\end{algorithm}

Let $m^\dagger(\varepsilon,p,T)$ denote the solution of Algorithm~\ref{ALG_network}, and $\texttt{Reg}^\dagger(\varepsilon, T)=m^\dagger(\varepsilon, 1,T)-\widehat{m}(\varepsilon)$ its regret under perfect learning.

Throughout, we consider an asymptotic regime with an infinitely large target on the number of covered customers. Formally, we define a sequence of graphs $(\calG_m)$ that satisfy Assumption~\ref{ass:degree}, solve the problem on each graph $\calG_m$ with target $m$, and assess the solution as $m\to\infty$. We assume that $\calG_m$ is large enough to cover $m$ customers with an order $\Omega(m^k)$ of spare facilities and customers, for some $0<k<g(r)$; this assumption avoids corner cases so that decision-maker can always select facilities with desired coverage per Algorithm~\ref{ALG_network}. As mentioned earlier, Assumption~\ref{ass:degree} further guarantees that a large number of facilities need to be opened, thereby linking the asymptotic regime with a target on customer coverage to the one from Section~\ref{sec:base} with a target on facilities.

Theorem~\ref{thm:network} shows that Algorithm~\ref{ALG_network} yields an asymptotically feasible to Problem~\eqref{prob:network} within an optimality gap of $\calO(m^{g(r)})$. Under perfect learning ($p=1$), it extends Theorem~\ref{thm:main} by showing that the optimal regret also grows in $\Theta(m^{g(r)})$. In other words, this theorem fully characterizes the performance of Algorithm~\ref{ALG_network} in Problem~\eqref{prob:network}, and it shows that it attains asymptotically optimal regret against the fully-learned benchmark under perfect learning.

\begin{theorem}\label{thm:network}
Consider a sequence of graphs $\calG_1,\cdots,\calG_m,\cdots$ satisfying Assumption~\ref{ass:degree}, such that Problem~\eqref{prob:network_ALG} admits a feasible solution with $\Omega(m^k)$ unopened facilities and unserved customers, with $0<k<g(r)$. As $m \to \infty$, Algorithm~\ref{ALG_network} yields a feasible solution to Problem~\eqref{prob:network} with
$$m^\dagger(\varepsilon, p,T)\leq m^*(\varepsilon, p,T)+\begin{cases}
    o\left(m^{\alpha_T \cdot (1-r)}\right)& \text{if }p=1 \text{ or } (p \neq 1 \text{ and } \alpha_T\cdot(1-r)>1/2)\\
    \calO(m^{1/2}) & \text{if }p \neq 1\text{ and } \alpha_T\cdot(1-r)\leq 1/2
\end{cases}$$
Assume further that $\delta = o\left(m^{\alpha_T\cdot(1-r)-1}\right)$ and $p=1$. The optimal regret satisfies $\texttt{Reg}^*(\varepsilon, T)=\Theta\left(m^{\alpha_T(1-r)}\right)$ and the solution of Algorithm~\ref{ALG_network} achieves the same rate $\texttt{Reg}^\dagger(\varepsilon, T)=\Theta\left(m^{\alpha_T(1-r)}\right)$.
\end{theorem}

Let us outline the main arguments from the proof (in~\ref{app:network}). The networked setting creates two additional complexities: interdependencies across customer coverage variables and unknown solution of the fully-learned benchmark. The former is addressed in the proof of the first part of the theorem, whereas the latter is addressed in the proof of the second part. Specifically:
\begin{itemize}
\item[--] The networked problem optimizes over the \textit{set} of facilities to open rather than its mere \textit{number}, which prevents the application of standard concentration inequalities. In response, we use the concentration inequality from \cite{janisch2024berry}, which extends the Berry-Esseen theorem in dependency graphs. This bounds the difference between customer coverage and its deterministic approximation in Problem~\eqref{prob:network_ALG} (Lemma~\ref{lem:hoeffding_gen_bound}). This approximation motivates Algorithm~\ref{ALG_network}, which estimates customer coverage via its deterministic approximation and adds a buffer to handle uncertainty. The rest of the proof of the first part of the theorem provides asymptotic bounds on the overall number of required facilities (Lemma~\ref{lem:sol_bound}) and on the error from deterministic approximations of coverage variables (Lemma~\ref{lem:prob_bound}); it derives a result on the sensitivity of the solution of Problem~\eqref{prob:network_ALG} with respect to the right-hand side coverage target (Lemma~\ref{lem:gen_deter_sol_lower}); and it uses these results to establish the optimality gap of the solution of Algorithm~\ref{ALG_network} against the optimal solution of Problem~\eqref{prob:network_ALG} (Lemma~\ref{lem:network_deter_stochstic}).
\item[--] Under perfect learning ($p=1$), the fully-learned benchmark can be formulated via integer optimization (Problem~\eqref{prob:network_perfect}) but its solution cannot be elicited in closed-form. In response, we derive in Lemma~\ref{lem:bound_loss_gen} structural properties of its solution, which we link to the algorithm's solution in Lemma~\ref{lem:network_deter_lower} to derive asymptotically tight regret in $\Theta(m^{g(r)})$. The main observation is that the fully-learned benchmark opens each facility at most once under perfect learning, which eliminates non-linearities in customer coverage probabilities.
\end{itemize}

Finally, Proposition~\ref{prop:network_nolearn} shows that the no-learning baseline still achieves linear regret, which also extends the result from Proposition~\ref{prop:regret_NL} to the networked setting.
\begin{proposition}\label{prop:network_nolearn}
The no-learning baseline achieves a regret of $\Omega\left(m\right)$ as $m \to \infty$.
\end{proposition}

In summary, in the perfect learning case, we have derived the same regret rate of $\Theta\left(\max\left\{m^{\frac{1-r}{1-r^T}},\sqrt{m}\right\}\right)$ against the fully-learned solution and the same guarantees of asymptotic optimality via matching upper and lower bounds. Under imperfect learning, we still showed the near optimality of the solution of Algorithm~\ref{ALG_network} against the optimal solution of the stochastic problem~\eqref{prob:network}, still in $\calO\left(\max\left\{m^{\frac{1-r}{1-r^T}},\sqrt{m}\right\}\right)$, although the regret from the fully-learned benchmark remains unknown. These results extend our main takeaways to the networked setting, including the sub-linear and asymptotically tight regret rate, as well as the asymptotic optimality of Algorithm~\ref{ALG_network}. In particular, they extend the managerial insights from Section~\ref{subsec:insights} regarding the benefits of limited online learning and optimization, which arise from the sub-linear regret rate and from the exponential convergence of the regret rate as the planning horizon grows.

\subsection{A solution algorithm and computational results}
\label{subsec:solution}

One practical challenge with Algorithm~\ref{ALG_network} is that the deterministic approximation in Problem~\eqref{prob:network_ALG} still exhibits a mixed-integer non-convex optimization structure that cannot be easily solved with off-the-shelf methods. We present in~\ref{app:network_ALG} a computational algorithm (Algorithms~\ref{alg:network_simp}--\ref{alg:network_full}) that opens facilities by increasing order of expected incremental coverage, starting with moderately connected facilities for learning-based exploration and then opening more central ones for optimization-based exploitation (the least connected facilities remain unopened altogether). We show that the solution is optimal in star graphs (Proposition~\ref{prop:exact}) and it provides a principled heuristic otherwise.

\begin{figure}[h!]
\centering
\subfloat[Facility openings.]{\label{fig:facilities}\includegraphics[width=.49\textwidth]{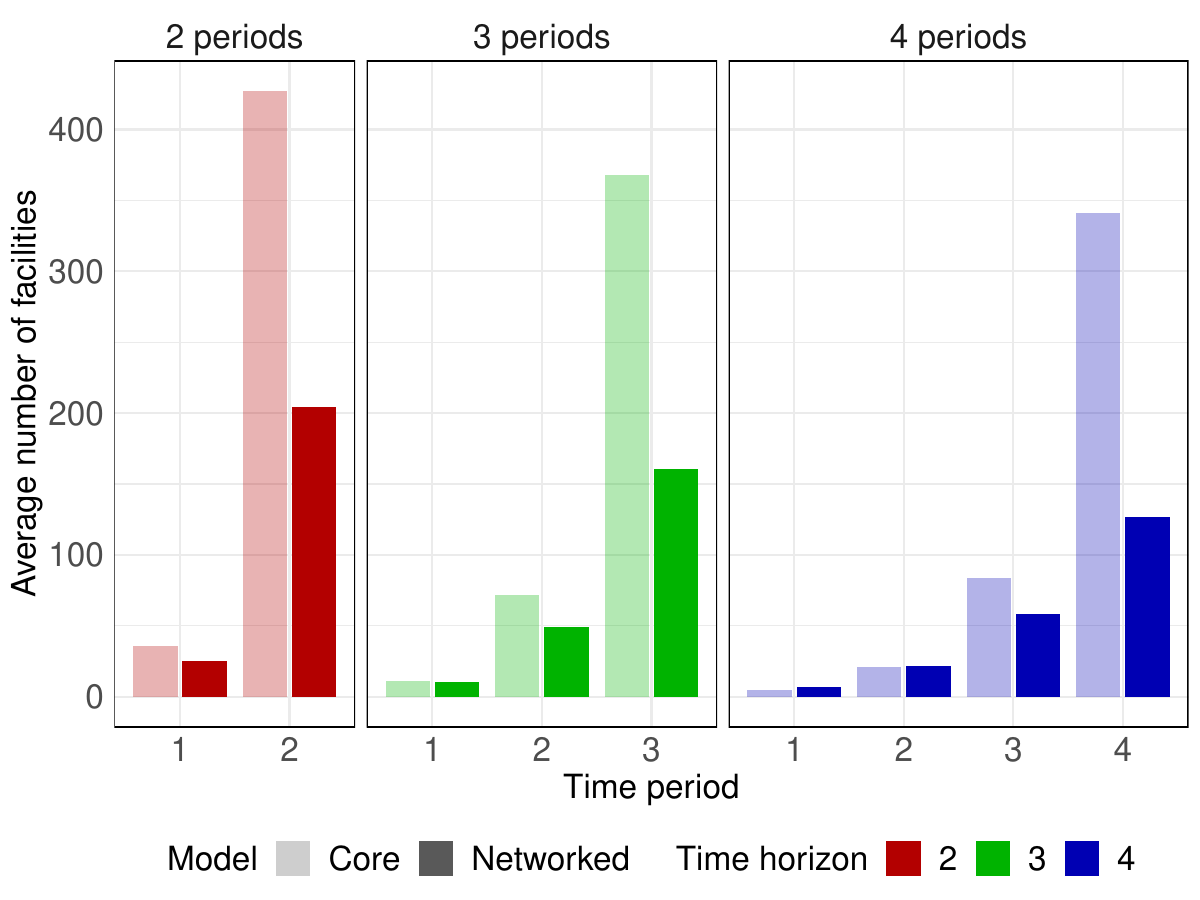}}
\hspace{2pt} 
\subfloat[Customer coverage.]{\label{fig:customers}\includegraphics[width=.49\textwidth]{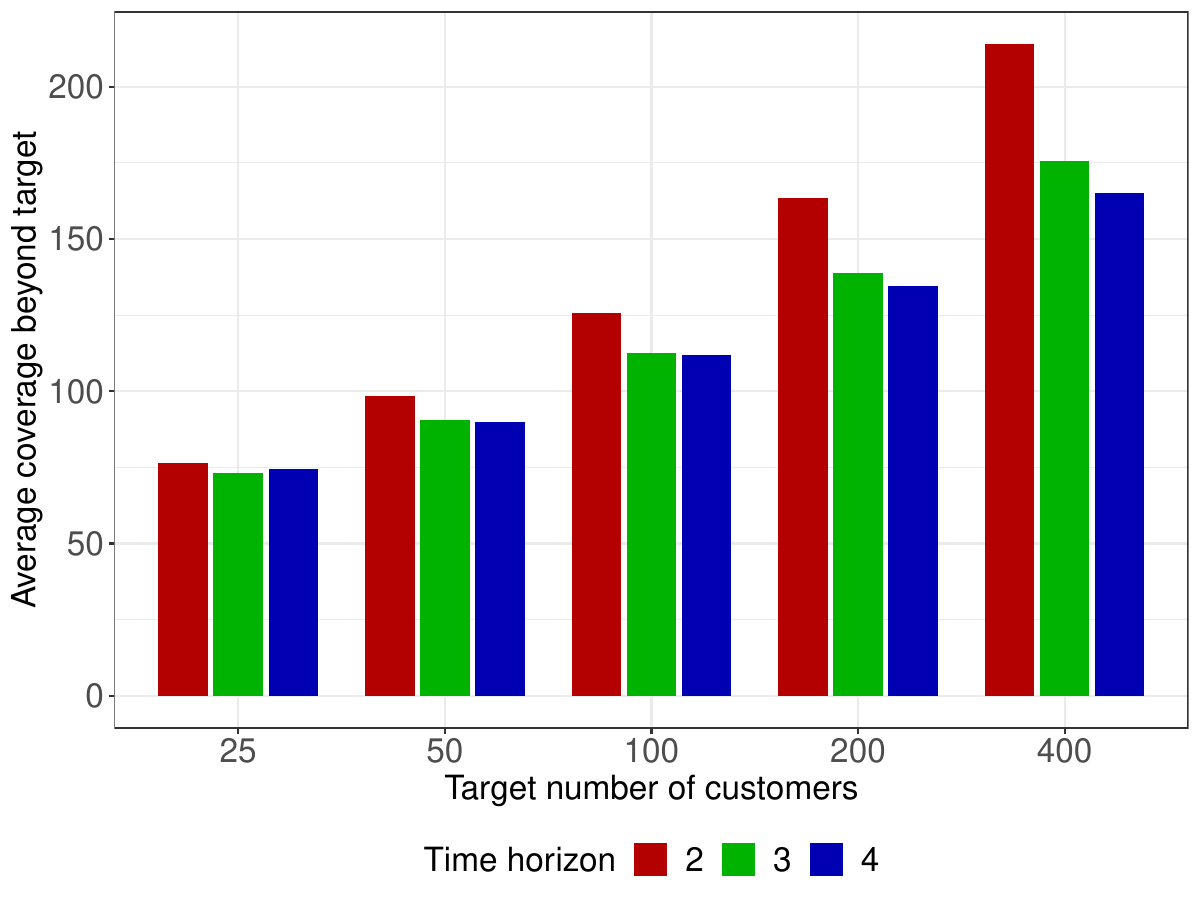}}
\caption{Computational results in the network as a function of $m$ and $T$ ($d_\star=3$, $\delta=1\%$, $\varepsilon=0.4$, $r=0.5$, $p=1$).}
\label{fig:customers_facilities}
\vspace{-12pt}
\end{figure}

We apply our algorithm on synthetic graphs with maximum degree $d_\star$, with $\eta m$ facilities, and with $\mu m$ customers for any target customer coverage $m$, where $\eta$ and $\mu$ denote scaling factors. Figure~\ref{fig:customers_facilities} reports customer coverage and facility openings, averaged across 10 randomized instances.

Figure~\ref{fig:facilities} reports the average number of facilities that get tentatively opened under the networked model and, for comparison, under the core model. As expected, fewer facilities need to be opened in the networked setting because each one can cover multiple customers. These differences are much more pronounced in later iterations than in earlier ones, which uncovers the distinct roles of exploration and exploitation. Early on, facility openings primarily aim to mitigate the learning error, so early decisions weakly exploit facility-customer connections. Later on, decision-maker selects fewer but more central facilities once the error gets mitigated, which enables to satisfy the target coverage by opening comparatively fewer facilities than in the corresponding core setting.

Figure~\ref{fig:customers} reports the average ``customer-based regret'' as a function of $m$ and $T$. Since we do not know the fully-learned solution, we visualize the difference between customer coverage and the target $m$ (the fully-learned solution covers $m$ customers within a tolerance of $d_\star$). With $r=0.5$ and $p=1$, our theoretical results guarantee that Algorithm~\ref{ALG_network} yields an asymptotically optimal solution and tight regret bounds, in $\Theta\left(m^{\alpha_T\cdot(1-r)}\right)$. The figure shows that the solution of Algorithms~\ref{alg:network_simp}--\ref{alg:network_full} also achieves sub-linear regret: as $m$ doubles, the average regret increases by 20-30\%. The $\Theta\left(m^{\alpha_T\cdot(1-r)}\right)$ result would indicate a regret growth by 45\% with $T=4$, by 49\% with $T=3$, and by 59\% with $T=2$. The computational results are consistent with---in fact, slightly better than---these predictions. Furthermore, the average regret varies only slightly as the horizon becomes longer. Altogether, these results reinforce our main takeaways regarding the benefits of even limited learning indicated by the sub-linear regret bounds and the exponential convergence of the regret rate.

\section{Conclusion}

Motivated by examples from healthcare, humanitarian operations, and technology investing, this paper studies an online learning and optimization problem under irreversible decisions and endogenous supply-side uncertainty. At each period, a decision-maker builds a classifier to predict facility success, selects facilities to open, observes the success of each one, and updates the classifier. The problem aims to minimize the number of facility openings over a finite horizon, subject to chance constraints on coverage target. We first proved a statistical learning result showing that, under certain statistical conditions, the online classifier converges to the Bayes-optimal classifier at a rate of at best $\calO(1/\sqrt n)$, where $n$ is the accrued sample size. This result led to a mathematical formulation of the online learning and optimization problem using a general-purpose characterization of the online classifier with learning rate of $r$ and irreducible error $1-p$.

Our main result is an asymptotically optimal algorithm and asymptotically tight bounds showing that the regret grows sub-linearly in the regime with a large coverage target but a finite planning horizon. In a core setting with a target $m$ on the number of successful facilities, we showed that the optimal regret grows in $\Theta\left(m^{\frac{1-r}{1-r^T}}\right)$, $\Theta\left(m^{1/T}\right)$ or $\Theta\left(\sqrt m\right)$. We established the robustness of this result in a dynamic re-optimization environment and extended it to a networked setting with a target on customer coverage. These theoretical results highlight the benefits of even \emph{limited} online learning and optimization. Specifically, the regret grows sub-linearly with the coverage targets, as compared to linearly for a no-learning baseline, and it converges exponentially fast as the horizon becomes longer. In practical terms, these results encourage to run small pilot programs for exploration purposes and to enable more effective exploitation later on as the error gets mitigated.

These theoretical and managerial insights motivate further research in online learning and optimization with discrete and irreversible decisions, using the general-purpose characterization of the learning error proposed in this paper. One opportunity lies in studying the learning-optimization trade-off under alternative specifications, such as settings where the decision-maker selects facilities with the highest estimated success probabilities or where target coverage takes other forms than the chance constraint considered in this paper. Another open question involves characterizing the regret in finite-sample regimes with a finite target on the number of facilities, or in graphs that do not satisfy the degree restriction. Yet another avenue lies in developing exact algorithms for the deterministic approximation model in the networked setting. Finally, the promising theoretical and numerical results motivate the application of the online learning and optimization approach in new empirical areas, in our motivating domains from healthcare, logistics, investing, and beyond.

\bibliographystyle{informs2014}
\bibliography{main.bib}

\newpage

\ECSwitch
\ECHead{Alexandre Jacquillat and Michael Lingzhi Li\\Learning to cover: online learning and optimization with irreversible decisions toward target coverage\\Electronic Companion}

\section{Appendix to Section~\ref{sec:setting}}
\label{app:stats}

\subsection{Proof of Proposition~\ref{prop:MLE}}
We first prove that $\widehat{\theta}_t$ is a consistent estimator.
\begin{lemma}\label{lem:mle}
For every $\varepsilon_0, \varepsilon_1,\gamma_1,\gamma_2>0$, there exists $a$ such that, for all $N_{t-1}\geq a$,
         \begin{align}
        \P\left(\|\widehat{\theta}_t-\theta_0\|_2\geq  \frac{\varepsilon_0}{(N_{t-1})^{1/2-\gamma_1}}\right)\leq\frac{\varepsilon_1}{(N_{t-1})^{1/2-\gamma_2}} \label{eq:mle_bound}
    \end{align}
\end{lemma} 
\proof{Proof of Lemma~\ref{lem:mle}}
First, we show that $\widehat{\theta}_t$ from Equation~\eqref{eq:mle_estimation} is the (conditional) maximum likelihood estimator based on independent, but not necessarily identically distributed, observations. While the facility success indicators $S_{it}$ may in general depend on prior facility openings, conditioning on the features $\bm{x}_i$ and $y_{it}=1$ fixes the success probability $\eta_i = f_{\theta_0}(\bm{x}_i)$ and the set of facilities we choose. Concretely, define $X :=\{x_i\}_{i=1}^M$, $Y_t:=\{y_{it}\}_{i=1}^M$ and $Z_t:=\{z_{it}\}_{i=1}^M$. We define the history up to period $t$ as:
\[\calH_s:=\sigma(X,\{(Y_r,Z_r)\}_{r\leq s})\]
where $\sigma(\cdot)$ is the sigma algebra induced by the randomness in the sampling of $Y_t$ and the success of $Z_t$ in our setup. By the chain rule, we have:
\begin{align*}
p_\theta(Z_{1:t-1},Y_{1:t-1} \mid X) &= \prod_{s=1}^{t-1} p_\theta(Z_s \mid \calH_{s-1}, Y_s) p(Y_s \mid \calH_{s-1})\\
p(Y_{1:t-1} \mid X) &= \prod_{s=1}^{t-1}  p(Y_s \mid \calH_{s-1})
\end{align*}
We obtain:
\begin{align*}
p_\theta(Z_{1:t-1} \mid X, Y_{1:t-1})
=&\frac{p_\theta(Z_{1:t-1},Y_{1:t-1} \mid X)}{p(Y_{1:t-1} \mid X)}
\\=&\frac{\prod_{s=1}^{t-1} p_\theta(Z_s \mid \calH_{s-1}, Y_s) p(Y_s \mid \calH_{s-1})}{\prod_{s=1}^{t-1}  p(Y_s \mid \calH_{s-1})}
\\=&\prod_{s=1}^{t-1} p_\theta(Z_s \mid \calH_{s-1}, Y_s)
\\=& \prod_{s=1}^{t-1} \prod_{i: y_{is}=1} (f_\theta(\bx_{i}))^{z_{is}}(1-f_\theta(\bx_{i}))^{1-z_{is}}
\end{align*}
where the last equality follows from the independence of $z_{is}$ across different facilities given $y_{is}$. This shows that, conditioned on observed $X$ and $Y$, Equation~\eqref{eq:mle_estimation} is a (conditional) likelihood based on independent ($z_{is}$) but not necessarily identical distributions (due to the different $f_\theta(\bx_i)$), and $\hat{\theta}_t$ is the corresponding maximum likelihood estimator (MLE).

\cite{hoadley1971asymptotic} defines the following regularity conditions for MLE estimators based on independent but not necessarily identical distributions:
\begin{enumerate}[label=(\alph*)]
\item $\Theta$ is a closed set.
\item $f_\theta(\bx)$ is an upper semi-continuous function of $\theta$ uniformly in $\bm{x} \in \mathcal{X}$.
\item The log-likelihood $\log p_\theta(Z_{1:t-1} \mid X, Y_{1:t-1})$ is a measurable function of $Z_{1:t-1}$ and is twice continuously differentiable with respect to $\theta$.
\item The log-likelihood $\log p_\theta(Z_{1:t-1} \mid X, Y_{1:t-1})$ is uniformly upper bounded by an integrable function $g(X, Y_{1:t-1})$ with respect to $X$ and $Y_{1:t-1}$. 
\item The derivative of the log-likelihood $\frac{\partial \log p_\theta(Z_{1:t-1} \mid X, Y_{1:t-1})}{\partial \theta}$ has finite third moments. 
\item The expected log-likelihood is identifiable: $R_t(\theta \mid \calJ_t)=R_t(\theta_0\mid \calJ_t) \iff \theta = \theta_0\qquad \forall t$
\item The Fisher information is positive definite at $\theta=\theta_0$:
\[\mathcal{I}_t(\theta_0\mid \calJ_t) := \frac{1}{|\calJ_t|} \sum_{i \in \calJ_t}\frac{1}{f_{\theta_0}(\bx_i)(1-f_{\theta_0}(\bx))}\frac{\partial f_{\theta_0}(\bx_i)}{\partial \theta}\frac{\partial f_{\theta_0}(\bx_i)}{\partial \theta^T} \succ 0 \qquad \forall t\]
\end{enumerate}
Condition (a) is implied by the first point of Statistical Condition \ref{ass:regularity}. Conditions  (b), (c) and (d) are implied by the second point in Statistical Condition \ref{ass:regularity}. Condition (e) directly follows because all moments of the Bernoulli distribution are finite and $f_\theta(\bx_i)$ is uniformly bounded. Conditions (f) and (g) are satisfied by the last two statements in Statistical Condition \ref{ass:regularity}. 

Therefore, per the central limit theorem for MLE estimation, $\sqrt{N_{t-1}}\left(\widehat{\theta}_t -\theta_0\right)\xrightarrow{d} N(0,\bSigma)$ for some covariance matrix $\bSigma$. Furthermore, under Statistical Condition~\ref{ass:regularity}, a multivariate Berry--Esseen bound holds \citep[e.g.][]{raivc2019multivariate}, so there exist constants $c, C$ such that for all $\varepsilon$, we have:
   \begin{equation*}
\P\left(\|\widehat{\theta}_t-\theta_0\|_2\geq \varepsilon \right)\leq \P(\|W\|_2\geq c\varepsilon \sqrt{N_{t-1}})  + \frac{C}{(N_{t-1})^{1/2}}
    \end{equation*}
where $W \sim N(0,\sum_t I_t(\theta_0 \mid \calJ_t))$ follows an $\ell$-dimensional normal distribution. Then, for any $\gamma_1,\gamma_2>0$, we have that:
    \begin{equation*}
\P\left(\|\widehat{\theta}_t-\theta_0\|_2\geq  \frac{\varepsilon}{(N_{t-1})^{1/2-\gamma_1}}\right)\leq \P(\|W\|\geq c\varepsilon N_{t-1}^{\gamma_1})  + \frac{C}{(N_{t-1})^{1/2}}\leq \frac{\varepsilon_1}{(N_{t-1})^{1/2-\gamma_2}}
    \end{equation*}
    where the last inequality is true for all $N_{t-1}$ sufficiently large.
\hfill\Halmos

Next, we bound the difference in accuracy between event probabilities conditioning on the classifier $h_t(\bx_i)=1$ versus the optimal classifier $h^*(\bx_i)=1$, as a function of the functions $f_{\widehat\theta_t}(\bx)$ and $f_{\theta}(\bx)$, of the concentration of $f_{\theta_0}(\bx)$ around $\tau$, and of the mass of exploration facilities.
\begin{lemma}\label{lem:bound_classification}
Let $E$ be any probability event defined over the set of facilities $\mathcal{F}$, so that $\P(E)=\frac{1}{|\calF|} \sum_i \bm{1}\{E\}$. For any $\varepsilon>0$, the following holds:
\[\left|\P\left(E \mid h_t(\bx_i)=1\right)- \P\left(E \mid h^*(\bx_i)=1\right)\right|\leq2\cdot\frac{\P\left( \left|f_{\widehat\theta_t}(\bx_i) - f_{\theta_0}(\bx_i)\right| > \varepsilon \right)+\P\left( \left|f_{\theta_0}(\bx_i) -\tau\right| < \varepsilon \right)+\P(i \in \calE_t)}{\P(h^*(\bx_i)=1)}\]
\end{lemma}
\proof{Proof of Lemma~\ref{lem:bound_classification}}
The result is derived as follows:
\begin{align*}
& \left|\P\left(E \mid h_t(\bx_i)=1\right)- \P\left(E \mid h^*(\bx_i)=1\right)\right|\\
=& \left|\frac{\sum_{i\in\calF} \P(E)\mathbf{1}\{h_t(\bx_i)=1\}}{\sum_{i\in\calF} \mathbf{1}\{h_t(\bx_i)=1\}}-\frac{\sum_{i\in\calF} \P(E)\mathbf{1}\{h^*(\bx_i)=1\}}{\sum_{i\in\calF} \mathbf{1}\{h^*(\bx_i)=1\}}\right| \\   
\leq& \left|\sum_{i\in\calF} \P(E)\mathbf{1}\{h_t(\bx_i)=1\}\left(\frac{1}{\sum_{i\in\calF} \mathbf{1}\{h_t(\bx_i)=1\}}-\frac{1}{\sum_{i\in\calF} \mathbf{1}\{h^*(\bx_i)=1\}}\right)\right|\\&+\left|\frac{\sum_{i\in\calF}  \P(E)\left(\mathbf{1}\{h_t(\bx_i)=1\}-\mathbf{1}\{h^*(\bx_i)=1\}\right)}{\sum_{i\in\calF} \mathbf{1}\{h^*(\bx_i)=1\}}\right|\\
\leq & 2\cdot\frac{\sum_{i\in\calF} \P(E)\left|\left(\mathbf{1}\{h_t(\bx_i)=1\}-\mathbf{1}\{h^*(\bx_i)=1\}\right)\right|}{\sum_{i\in\calF} \mathbf{1}\{h^*(\bx_i)=1\}} \\
\leq & 2\cdot \frac{\sum_{i\in\calF} |\mathbf{1}\{h^*(\bx_i)=1\}- \mathbf{1}\{h_t(\bx_i)=1\}|}{\sum_{i\in\calF} \mathbf{1}\{h^*(\bx_i)=1\}}\\
= & 2\cdot \frac{\sum_{i\in\calF}\mathbf{1}\{f_{\widehat\theta_t}(\bx_i) > \tau, f_{\theta_0}(\bx_i) < \tau\}+\sum_{i\in\calF}\mathbf{1}\{f_{\widehat\theta_t}(\bx_i) <\tau, f_{\theta_0}(\bx_i) >\tau\}+\sum_{i\in\calF}\mathbf{1}\{i \in \calE_t\}}{\sum_{i\in\calF} \mathbf{1}\{h^*(\bx_i)=1\}}\\
\leq & 2\cdot \frac{\sum_{i\in\calF}\mathbf{1}\{f_{\widehat\theta_t}(\bx_i) > \tau, f_{\theta_0}(\bx_i) < \tau-\varepsilon \}+\sum_{i\in\calF}\mathbf{1}\{f_{\widehat\theta_t}(\bx_i) <\tau, f_{\theta_0}(\bx_i) >\tau+\varepsilon\}+\sum_{i\in\calF}\mathbf{1}\{\left|f_{\theta_0}(\bx_i) - \tau\right| < \varepsilon\}}{\sum_{i\in\calF} \mathbf{1}\{h^*(\bx_i)=1\}}\\
& + 2\cdot \frac{\sum_{i\in\calF}\mathbf{1}\{i \in \calE_t\}}{\sum_{i\in\calF} \mathbf{1}\{h^*(\bx_i)=1\}}\\
\leq & 2\cdot \frac{\sum_{i\in\calF}\mathbf{1}\{|f_{\widehat\theta_t}(\bx_i)-f_{\theta_0}(\bx_i)|>\varepsilon\}+\sum_{i\in\calF}\mathbf{1}\{\left|f_{\theta_0}(\bx_i) - \tau\right| < \varepsilon\}+\sum_{i\in\calF}\mathbf{1}\{i \in \calE_t\}}{\sum_{i\in\calF} \mathbf{1}\{h^*(\bx_i)=1\}}\\
=& 2\cdot\frac{\P\left( \left|f_{\widehat\theta_t}(\bx_i) - f_{\theta_0}(\bx_i)\right| > \varepsilon \right)+\P\left( \left|f_{\theta_0}(\bx_i) - \tau\right| < \varepsilon \right)+\P(i \in \calE_t)}{\P(h^*(\bx_i)=1)}.
\end{align*}
This completes the proof of the lemma.
\hfill\Halmos
\subsubsection*{Proof of Proposition~\ref{prop:MLE}}
Fix $\gamma>0$. Since $N_t=\calO(m)$ for all $t$, so we write $N_t\leq \Psi m$ for some constant $\Psi$ and for $m\geq m_A$. Fix $\widehat\varepsilon>0$ and define $p^*= \P(h^*(\bx_i)=1)$. Per Statistical Condition~\ref{ass:regime}, we can define $m_B$ such that $\frac{m}{M}\leq\frac{\widehat\varepsilon}{9\Psi(N_{t-1})^{1/2-\gamma}}$ for all $m \geq m_B$. This implies that:
\begin{equation}
    \P(y_{i\tau}=1\ \text{for $\tau\leq t-1$})\leq \frac{\Psi m}{M} \leq \frac{\widehat\varepsilon}{9(N_{t-1})^{1/2-\gamma}} \label{eq:regime}
\end{equation}
Now define $\varepsilon_0>0$ such that $\frac{C\varepsilon_0^\alpha}{9^\alpha}=\frac{\widehat\varepsilon}{9}$. We proceed to prove the main result.
\begin{align*}
1-\mu_t=&\P(S_{it}=1 \mid y_{i1}=\cdots=y_{i,t-1}=0\ ;\ h_t(\bx_i)=1)\\
=&\frac{\P(S_{it}=1 \mid h_t(\bx_i)=1)-\P(y_{is}=1\ \text{for $s\leq t-1$} \mid h_t(\bx_i)=1)\P(S_{it}=1 \mid h_t(\bx_i)=1, y_{is}=1\ \text{for $s\leq t-1$})}{\P(y_{i1}=\cdots=y_{i,t-1}=0 \mid h_t(\bx_i)=1)}\\
\geq& \P(S_{it}=1 \mid h_t(\bx_i)=1)-\P(y_{is}=1\ \text{for $s\leq t-1$} \mid h_t(\bx_i)=1)\\
\geq&\P(S_{it}=1 \mid h^*(\bx_i)=1)-\P(y_{is}=1\ \text{for $s\leq t-1$} \mid h^*(\bx_i)=1)\\
&-4\cdot\frac{\P\left( \left|f_{\widehat\theta_t}(\bx_i) - f_{\theta_0}(\bx_i)\right| > \frac{\varepsilon_0}{9(N_{t-1})^{1/2-\gamma}} \right)}{\P(h^*(\bx_i)=1)}- 4\cdot\frac{\P\left( \left|f_{\theta_0}(\bx_i) - \tau\right| < \frac{\varepsilon_0}{9(N_{t-1})^{1/2-\gamma}} \right)}{\P(h^*(\bx_i)=1)}-4\cdot \frac{\P(i \in \calE_t)}{\P(h^*(\bx_i)=1)}\\
\geq&\P(S_{it}=1 \mid h^*(\bx_i)=1)-\frac{\widehat\varepsilon}{9p^*(N_{t-1})^{1/2-\gamma}}\\
&-4\cdot\frac{\P\left( \left|f_{\widehat\theta_t}(\bx_i) - f_{\theta_0}(\bx_i)\right| > \frac{\varepsilon_0}{9(N_{t-1})^{1/2-\gamma}} \right)}{p^*}- 4\cdot\frac{\P\left( \left|f_{\theta_0}(\bx_i) - \tau\right| < \frac{\varepsilon_0}{9(N_{t-1})^{1/2-\gamma}} \right)}{p^*}-4\cdot \frac{K}{p^*(N_{t-1})^{1/2}},       
\end{align*}
where the equality comes from the law of total probabilities, the first inequality is obtained by lower bounding probabilities by 1, and the second inequality is derived from applying~\ref{lem:bound_classification} twice. The third inequality comes from Assumption \ref{ass:positive} which states that $p^*\geq \xi>0$ and Statistical Condition \ref{ass:exploration} which bounds the probability of $i$ in $\calE_t$. 
Now from Statistical Condition~\ref{ass:margin}, we know that $\P\left( \left|f_{\theta}(\bx_i) - \tau\right| \leq \frac{\varepsilon_0}{9(N_{t-1})^{1/2-\gamma}} \right)\leq C\left(\frac{\varepsilon_0}{9(N_{t-1})^{1/2-\gamma}}\right)^\alpha=\frac{\widehat\varepsilon}{9(N_{t-1})^{\alpha\cdot(1/2-\gamma)}}$. And from Statistical Condition~\ref{ass:regularity}, we know that $f$ is twice continuously differentiable in $\theta$ and that $\Theta$ is compact, so $f$ is Lipschitz continuous in $\theta\in\Theta$. Let $L$ be the Lipschitz constant. We obtain:
\begin{align*}
1-\mu_t\geq&\P(S_{it}=1 \mid h^*(\bx_i)=1)-\frac{\widehat\varepsilon}{9p^*(N_{t-1})^{1/2-\gamma}}\\&
-\frac{4}{p^*}\P\left( \|\widehat\theta_t-\theta\|_2 > \frac{\varepsilon_0}{9L(N_{t-1})^{1/2-\gamma}} \right)- \frac{4\widehat\varepsilon}{9p^*(N_{t-1})^{\alpha\cdot(1/2-\gamma)}}-\frac{4K}{p^*(N_{t-1})^{1/2}}
\end{align*}

By Lemma~\ref{lem:mle}, define $m_C$ such that for all $m\geq m_C$, $N_{t-1}$ will be sufficiently large for the following to hold:
\begin{equation}
    \P\left( \|\widehat\theta_t -\theta\|_2 > \frac{\varepsilon_0}{9L(N_{t-1})^{1/2-\gamma}} \right)\leq \frac{\widehat\varepsilon}{9(N_{t-1})^{1/2-\gamma}},\ \forall t=1,\cdots,T \label{eq:mle_bound2}
\end{equation}
Now fix $m_0=\max\{m_A,m_B,m_C\}$ and consider any $m\geq m_0$. We obtain:
\begin{align*}
1-\mu_t\geq&\P(S_{it}=1 \mid h^*(\bx_i)=1)-\frac{\widehat\varepsilon}{9p^*(N_{t-1})^{1/2-\gamma}}
-\frac{4\widehat\varepsilon }{9p^*(N_{t-1})^{1/2-\gamma}}- \frac{4\widehat\varepsilon}{9p^*(N_{t-1})^{\alpha\cdot(1/2-\gamma)}}-\frac{4K}{p^*(N_{t-1})^{1/2}}\\
\geq&1-\mu^*-\frac{\widehat\varepsilon+4K}{p^*(N_{t-1})^{\min\{\alpha,1\}\times (1/2-\gamma)}}.
\end{align*}
Therefore, we can define $\overline\varepsilon=\frac{2(\widehat\varepsilon+4K)}{p^*}$ such that:
\[1-\mu_t\geq 1-\mu^*-\frac{\overline\varepsilon}{(N_{t-1}+1)^{\min\{\alpha,1\}\times (1/2-\gamma)}}.\]
This completes the proof.\hfill\Halmos

\subsection{Proof of Corollary~\ref{cor:learning}}

Define $\widetilde\mu_t$ such that $\widetilde{S}_{it} \sim \Bern(1-\widetilde\mu_t)$. Per Proposition \ref{prop:MLE}, $\widetilde\mu_t\geq  \mu_t$, irregardless of the possible dependence of $\mu_t$ on the previous set of selected facilities. Since $B_t \mid \mu_t \sim \Bino\left(A_t, 1-\mu_t\right)$ and $\widetilde{B}_t \sim \Bino\left(A_t, 1-\widetilde\mu_t\right)$, this implies:
\[\P\left[\sum_{t=1}^T\widetilde{B}_t\geq m \right] \leq \P\left[\sum_{t=1}^TB_t\geq m \mid \mu_t \right]\]
Therefore, by the law of total expectation, taking the expectation over $\mu_t$ gives:
\[\P\left[\sum_{t=1}^T\widetilde{B}_t\geq m \right] \leq \P\left[\sum_{t=1}^TB_t\geq m \right]\]

\subsection{Proof of Corollary~\ref{cor:success}}

Recall that $B_t \;=\; \sum_{i:\,y_{it}=1} S_{it}$. Under Assumption~\ref{ass:sampling} the indices with $y_{it}=1$ are drawn
\emph{with} replacement from the set
$\calJ_t$. Therefore, Assumption~\ref{ass:classification} implies that every draw from
$\calJ_t$ has the same conditional success probability
\[
p_t \;=\; \frac{1}{|\calJ_t|} \sum_{i \in \calJ_t} \eta_i = \P(h_t(\bx_i) = S_{it} \mid i \in \calJ_t)=1-\frac{\varepsilon p}{(N_{t-1}+1)^{r}}-\varepsilon\,(1-p),
\]
so that $S_{it} \mid y_{it} = 1 \sim \text{Bernoulli}(p_t)$ for all $i\in\calJ_t$.
The model specification $S_{it} \sim \text{Bernoulli}(\eta_i)$ further
assumes independence across facilities, i.e.\ $S_{it}\;\perp \!\!\! \perp\;S_{jt}$ for
$i\neq j$.
Because sampling is with replacement, the selection indicators
$\{y_{it}\}_{i\in\calJ_t}$ are independent conditional on $i \in \calJ_t$ as well; hence, conditional on
$\calJ_t$, the random variables
$\{S_{it}:y_{it}=1\}$ are i.i.d.\ $\text{Bernoulli}(p_t)$.
Therefore
\[
B_t \;\sim\; \text{Binomial}\bigl(A_t,\,p_t\bigr),
\]
which completes the proof.
\hfill\Halmos

\section{Proofs from Section~\ref{sec:base}}
\label{app:base}

\subsection{Proof of Proposition~\ref{prop:regret_NL}}
\label{app:regret_NL}

The fully-learned benchmark and the no-learning baseline exhibit similar formulations, with differences in the error rates. We make use of the following lemma, which proves a general result on the asymptotic number of trials required for a binomial distribution to satisfy the chance constraint.
\begin{lemma}\label{lem:binom}
Consider the following problem and denote by $n^*(m)$ its optimal solution:
$$\min\quad\left\{A\ :\ \P\left[\Bino\left(n,q\right)\geq m\right] \geq 1-\delta\right\}$$
The following holds:
$$n^*=\frac{m}{q}-\frac{\Phi^{-1}(\delta)\sqrt{1-q}}{q}\sqrt{m}+\calO(1)$$
\end{lemma}

\subsubsection{Proof of Lemma~\ref{lem:binom}.}

Let us define $S_n=\Bino\left(n,q\right)$.

\underline{\textit{Claim 1:}} Denote $\overline n=\frac{m}{q}-\frac{\Phi^{-1}(\delta)\sqrt{1-q}}{q}\sqrt{m}$. The following holds:
$$\P(S_{\overline n}\leq m-1)=\delta+\calO\left(\frac{1}{\sqrt{m}}\right)$$

We write $S_n=X_1+\cdots+X_n$ with $X_1,\cdots,X_n$ defined as independent Bernoulli variables with probability of success $q$. We have:
\begin{align*}
& \E(X_i)=q\\
& \text{Var}(X_i)=q(1-q)\\
& \E(|X_i-\E(X_i)|^3)=q(1-q)(q^2+(1-q)^2)
\end{align*}
By the Berry-Esseen theorem, we obtain for some constant $C$:
$$\left|\P\left(\frac{S_n-nq}{\sqrt{nq(1-q)}}\leq a\right)-\Phi(a)\right|\leq C\cdot\frac{q(1-q)(q^2+(1-q)^2)}{q^{3/2}(1-q)^{3/2}}\cdot\frac{1}{\sqrt n}=C\cdot\frac{q^2+(1-q)^2}{\sqrt{q(1-q)}}\cdot\frac{1}{\sqrt n}$$

Let us denote by $x(n)=\frac{m-1/2-nq}{\sqrt{nq(1-q)}}$. Note that
$$\P(S_n\leq m-1)=\P(S_n\leq m-1/2)=\P\left(\frac{S_n-nq}{\sqrt{nq(1-q)}}\leq x(n)\right)$$
We derive, for a constant $c_1=C\cdot\frac{(q^2+(1-q)^2)}{\sqrt{q(1-q)}}$:
$$\left|\P(S_n\leq m-1)-\Phi(x(n))\right|\leq \frac{c_1}{\sqrt n}$$

Using the definition of $x(n)$, we obtain:
\begin{align*}
& \left|\P(S_{\overline n}\leq m-1)-\Phi\left(\frac{m-1/2-{\overline n}q}{\sqrt{{\overline n}q(1-q)}}\right)\right|\leq\frac{c_1}{\sqrt {\overline n}},
\end{align*}
that is, by construction of $\overline n$:
\begin{align*}
&\left|\P(S_{\overline n}\leq m-1)-\Phi\left(\frac{\Phi^{-1}(\delta)\sqrt{1-q}\sqrt{m}-1/2}{\sqrt{(1-q)}\cdot\sqrt{m-\Phi^{-1}(\delta)\sqrt{1-q}\sqrt{m}}}\right)\right|\leq\frac{c_1\cdot\sqrt{q}}{\sqrt{m-\Phi^{-1}(\delta)\sqrt{1-q}\sqrt{m}}}
\end{align*}
Note in particular that:
$$\Phi(x(n))=\delta+\calO\left(\frac{1}{\sqrt{m}}\right).$$
Hence:
$$\P(S_{\overline n}\leq m-1)=\delta+\calO\left(\frac{1}{\sqrt{m}}\right)$$
This completes the proof of Claim 1. We then want to show that the optimal value of $n$ is within $\calO(1)$ of $\overline{n}$. To this end, we need to study the sensitivity of the chance constraint with respect to $n$, which requires to study the sensitivity of $x(n)$ with respect to $n$.

\underline{\textit{Claim 2:}} There exists a constant $c_2$ such that, for any constant $\Delta>0$, we have:
$$x(n+\Delta)-x(n)\leq-c_2\cdot\frac{\Delta}{\sqrt{m}}$$

We write:
$$x(n)=\frac{m-1/2}{\sqrt{q(1-q)}}\cdot\frac{1}{\sqrt n}-\frac{\sqrt q}{\sqrt{1-q}}\cdot\sqrt n$$
Taking the differential with respect to $n$, we derive:
$$x'(n)=-\frac{1}{2}\frac{1}{\sqrt{1-q}}\left(\frac{m-1/2}{\sqrt q}\cdot\frac{1}{n^{3/2}}+\sqrt q\cdot\frac{1}{\sqrt n}\right)$$
In particular, for $n=\Theta(m)$, we get $x'(n)=-\Theta\left(\frac{1}{\sqrt{m}}\right)$, hence there exists a constant $c_2$ such that $x'(n)\leq-c_2\cdot\frac{1}{\sqrt{m}}$ for $m$ large enough. By the mean value theorem, we write $x(n+\Delta)-x(n)=x'(\widetilde{n})\cdot\Delta$ for some $\widetilde{n}\in[n,n+\Delta]$ hence $\widetilde n=\Theta(m)$, which completes the proof of Claim 2.

\underline{\textit{Claim 3:}} There exist a constant $\Delta$ such that $\overline{n}-\Delta\leq n^*\leq\overline{n}+\Delta$.

Let us denote by $\varphi=\Phi'$. We know that $\varphi$ is continuous and $\varphi(\Phi^{-1}(\delta))>0$, so there exist constants $\eta>0$ and $\underline{\varphi}>0$ such that $\varphi(x)\geq\underline{\varphi}$ for all $x$ satisfying $|x-\Phi^{-1}(\delta)|\leq\eta$. Recall that $\Phi(x(\overline{n}))=\delta+\calO\left(\frac{1}{\sqrt m}\right)$ (Claim 1) and that $x(\overline{n}+\Delta)-x(\overline{n})=\calO\left(\frac{1}{\sqrt m}\right)$ with $x(\overline{n}+\Delta)-x(\overline{n})<0$ (Claim 2). Therefore, for sufficiently large $m$, we have $|x(\overline{n})-\Phi^{-1}(\delta)|\leq\eta$ and $|x(\overline{n}+\Delta)-\Phi^{-1}(\delta)|\leq\eta$. By the mean value theorem and Claim 2, there exist $\widehat{x}\in[x(\overline{n}),x(\overline{n}+\Delta)]$ such that $\Phi(x(\overline{n}+\Delta))-\Phi(x(\overline{n}))=\varphi(\widehat{x})(x(\overline{n}+\Delta)-x(\overline{n}))$. Therefore, it comes:
$$\Phi(x(\overline{n}+\Delta))-\Phi(x(\overline{n}))\leq\underline{\varphi}\cdot\left(x(\overline{n}+\Delta)-x(\overline{n})\right)\leq-\underline{\varphi}\cdot c_2\cdot\frac{\Delta}{\sqrt m}$$
Similarly, by proceeding as in Claim 2 and above, we can prove that:
$$\Phi(x(\overline{n}-\Delta))-\Phi(x(\overline{n}))\geq\underline{\varphi}\cdot\left(x(\overline{n}-\Delta)-x(\overline{n})\right)\geq\underline{\varphi}\cdot c_2\cdot\frac{\Delta}{\sqrt m}$$

Using Claim 1, we can now establish that, for a sufficiently large constant $\Delta$, $\overline{n}+\Delta$ is feasible whereas $\overline{n}-\Delta$ is not. We have, for some constants $c_3$ and $c_4$ coming from Claim 1, using $n=\Theta(m)$:
\begin{align*}
\P(S_{\overline{n}+\Delta}\leq m-1)
&\leq \Phi(x(\overline{n}+\Delta))+\frac{c_3}{\sqrt m}\\
&\leq \Phi(x(\overline{n}))-\underline{\varphi}\cdot c_2\cdot\frac{\Delta}{\sqrt m}+\frac{c_3}{\sqrt m}\\
&\leq \delta+\frac{c_4}{\sqrt m}-\underline{\varphi}\cdot c_2\cdot\frac{\Delta}{\sqrt m}+\frac{c_3}{\sqrt m}\\
&\leq\delta
\end{align*}
for $\Delta$ sufficiently large. Similarly:
\begin{align*}
\P(S_{\overline{n}-\Delta}\leq m-1)
&\geq \Phi(x(\overline{n}-\Delta))-\frac{c_3}{\sqrt m}\\
&\geq \Phi(x(\overline{n}))+\underline{\varphi}\cdot c_2\cdot\frac{\Delta}{\sqrt m}-\frac{c_3}{\sqrt m}\\
&\geq \delta-\frac{c_4}{\sqrt m}+\underline{\varphi}\cdot c_2\cdot\frac{\Delta}{\sqrt m}-\frac{c_3}{\sqrt m}\\
&>\delta
\end{align*}
This proves that $\P(S_{\overline{n}+\Delta}\geq m)\geq1-\delta$ and $\P(S_{\overline{n}-\Delta}\geq m)<1-\delta$. This proves that $n^*=\overline{n}+\calO(1)$, and completes the proof of the lemma.

\subsubsection{Proof of Proposition~\ref{prop:regret_NL}}

Per Lemma~\ref{lem:binom}, we know that:
\begin{align}
\widehat{m}(\varepsilon,p)&=\frac{m}{1-\varepsilon(1-p)}-\frac{\Phi^{-1}(\delta)\sqrt{\varepsilon(1-p)}}{1-\varepsilon(1-p)}\sqrt{m}+\calO(1)\label{eq:mhat}\\
m^{\text{NL}}(\varepsilon)&=\frac{m}{1-\varepsilon}-\frac{\Phi^{-1}(\delta)\sqrt{\varepsilon}}{1-\varepsilon}\sqrt{m}+\calO(1)\label{eq:mNL}
\end{align}
Therefore:
\begin{align*}
m^{\text{NL}}(\varepsilon)-\widehat{m}(\varepsilon,p)
&=\frac{m\varepsilon p}{(1-\varepsilon)(1-\varepsilon(1-p))}+\calO\left(\sqrt m\right)
\end{align*}
This proves that $\texttt{Reg}^{\text{NL}}(\varepsilon,p)=\Theta(m)$ as $m\to\infty$.

\subsection{Proof of Theorem~\ref{thm:main}}

Throughout this proof, we denote by $A^*_{t}$ the optimal solution of the stochastic problem (Problem~\eqref{prob:main}), by $A^\dagger_{t}$ the solution of Algorithm~\ref{ALG}, and by $A^{\textsc{D}}_{t}$ the solution of the deterministic approximation (Problem~\eqref{prob:main_det}). We denote by $B^*_{t}$, $B^\dagger_{t}$, and $B^{\textsc{D}}_{t}$ the corresponding numbers of facilities successfully opened \emph{at} period $t$, and by $N^*_{t}$, $N^\dagger_{t}$, and $N^{\textsc{D}}_{t}$ the corresponding number of facilities successfully opened \emph{by} period $t$.

Lemma~\ref{lem:main_det} elicits the solution $A^{\textsc{D}}_{t}$; Lemma~\ref{lem:main_feasible} establishes the feasibility of solution $A^\dagger_{t}$ in view of the chance constraint in Problem~\eqref{prob:main}, thus providing a constructive upper bound of $m^*(\varepsilon, p,T)$; and Lemma~\ref{lem:main_lower} provides a lower bound of $m^*(\varepsilon, p,T)$. Throughout the proof, we will focus on the case where $r\neq 1$. The case where $r=1$ is treated identically.

\subsubsection{Proof of Lemma~\ref{lem:main_det}}

Note that the constraint $N^D_T\geq m$ is binding at optimality, so the problem can be re-written as:
\begin{align*}
\min\ \left\{\ \sum_{t=1}^T A^D_t \ \ \bigg|\ \ \sum_{t=1}^TA^D_t \left(1 - \frac{\varepsilon\cdot p}{\left(\sum_{s=1}^{t-1} A^D_s +1\right)^r} - \varepsilon\cdot(1-p)\ \right)=m\right\}
\end{align*}

Let $\lambda\in\R$ be the dual variable associated with the equality constraint. The Karush–Kuhn–Tucker (KKT) conditions yield the following vanishing gradient equation:
$$1+\lambda\left(1 - \frac{\varepsilon\cdot p}{\left(\sum_{s=1}^{t-1} A^D_s +1\right)^r} - \varepsilon\cdot(1-p)\right)+\lambda\sum_{\tau=t+1}^T\frac{\varepsilon\cdot p\cdot r\cdot A^D_\tau}{\left(\sum_{s=1}^{\tau-1} A^D_s +1\right)^{r+1}}=0,\ \forall t\in\calT$$

Clearly, $\lambda\neq0$. After re-arranging the terms, we obtain:
$$\frac{1}{\left(\sum_{s=1}^{t-1} A^D_s +1\right)^r}-\sum_{\tau=t+1}^T\frac{r\cdot A^D_\tau}{\left(\sum_{s=1}^{\tau-1} A^D_s +1\right)^{r+1}}=\frac{1+\lambda}{p\lambda\varepsilon} - \frac{1-p}{p},\ \forall t\in\calT$$

Note that this expression is independent of $t$. Therefore, for all $t=1,\cdots,T-1$, we have:
$$\frac{1}{\left(\sum_{s=1}^{t-1} A^D_s +1\right)^r}-\sum_{\tau=t+1}^T\frac{ r\cdot A^D_\tau}{\left(\sum_{s=1}^{\tau-1} A^D_s +1\right)^{r+1}}=\frac{1}{\left(\sum_{s=1}^{t} A^D_s +1\right)^r}-\sum_{\tau=t+2}^T\frac{r\cdot A^D_\tau}{\left(\sum_{s=1}^{\tau-1} A^D_s +1\right)^{r+1}}.$$

This system of equations gives rise to a simple recurrence relation for $A^D_{t}$:
\begin{equation*}
A^D_{t+1} = \frac{\left(\sum_{s=1}^{t} A^D_s +1\right)^{r+1}}{r\left(\sum_{s=1}^{t-1} A^D_s +1\right)^r} - \frac{1}{r}\left(\sum_{s=1}^{t} A^D_s +1\right),\ \forall t=1,\cdots,T-1.
\end{equation*}

Let us define $u_t = \sum_{s=1}^t A^D_s + 1$. By expressing $A^D_t=u_t-u_{t-1}$, we can re-write the recursion as:
\begin{equation*}
u_{t+1} - u_t = \frac{u_t^{r+1}}{ru_{t-1}^r}-\frac{u_t}{r},\ \forall t=1,\cdots,T-1,
\end{equation*}
i.e., as:
\begin{equation}
u_{t+1} + \left(\frac{1}{r}-1\right)u_t = \frac{u_t^{r+1}}{ru_{t-1}^r},\ \forall t=1,\cdots,T-1. \label{eq:recur_rel}
\end{equation}

We are interested in the behavior of $u_t$ as $m \to \infty$. We perform a leading-order analysis and a leading-coefficient analysis, assuming that $u_t = \Theta(g(t)m^{f(t)})$. Note, first, that the boundary condition of $u_0=1=\Theta(1)$ implies that $g(0)=1$ and $f(0)=0$. Moreover, we prove that $u_T = \frac{m}{1-(1-p)\varepsilon} + o(m)=\Theta(m)$, so that $g(T)=c_0$ and $f(T)=1$. Indeed, consider the solution $\bar A_1=\frac{m^{1/(r+1)}}{1-\varepsilon}$ and $\bar A_2=\frac{m}{1-(1-p)\varepsilon} $. We have:
\begin{align*}
N^D_2 &=  \bar A_1(1-\varepsilon)+\bar A_2\left(1-\frac{\varepsilon\cdot p}{(\bar A_1+1)^{r}}-(1-p)\varepsilon\right)\\
\geq& m^{1/(r+1)} + m - \frac{m}{(\bar A_1+1)^{r}}\\
\geq& m^{1/(r+1)}+m-\frac{m}{m^{r/(r+1)}}\\
=& m
\end{align*}
Therefore, $\left(\frac{m^{1/(r+1)}}{1-\varepsilon},\frac{m}{1-(1-p)\varepsilon},0,\cdots,0\right)$ is a feasible solution of Problem~\eqref{prob:main_det} with $T\geq2$, so
$$\frac{m}{1-(1-p)\varepsilon}\leq u_T\leq \frac{m^{1/(r+1)}}{1-\varepsilon}+\frac{m}{1-(1-p)\varepsilon}+1$$
This proves that $u_T = \frac{m}{1-(1-p)\varepsilon} + o(m)=\Theta(m)$, hence $g(T)=c_0$ and $f(T)=1$.

%\textbf{Case 1: $\bm{r\neq 1}$}

Let us derive the expression of $f(t)$. From Equation~\eqref{eq:recur_rel}, we obtain the following recursion:
\[f(t+1) = (r+1)f(t)-rf(t-1),\]
i.e.:
\[f(t+1)-f(t) = r\left(f(t)-f(t-1)\right).\]
We derive:
$$f(t+1)-f(t)=Kr^t,\quad\text{where $K=f(1)-f(0)=f(1)$.}$$
By summation and by telescoping the sum, this yields:
$$\sum_{s=0}^{T-1}(f(s+1)-f(s))=f(T)-f(0)=\frac{K}{1-r}\left(1-r^T\right).$$
From the boundary conditions, we have $f(T)-f(0)=1$, so that:
$$f(t+1)-f(t)=r^{t}\cdot\frac{1-r}{1-r^T}$$
Again, by summation and by telescoping the sum, the following holds:
$$f(t)=\sum_{s=0}^{t-1}(f(s+1)-f(s))=\frac{1-r}{1-r^T}\sum_{s=0}^{t-1}r^s=\alpha_T(1-r^t).$$
In conclusion, the leading-order analysis leads to:
$$u_t=\Theta\left(m^{\alpha_T(1-r^t)}\right)=\Theta\left(m^{\alpha_T\left(1-r^t\right)}\right).$$

We conduct a similar analysis for the leading coefficient of $u_t$. Equation~\eqref{eq:recur_rel} yields
\[g(t+1) = \frac{g(t)^{1+r}}{r(g(t-1))^r}\]
Define $h(t)=\log g(t)$ and we have the following recurrent relation with $h(0)=h(T)=0$:
\[h(t+1) = (r+1)h(t)-rh(t-1)-\log r\]
We proceed as above, with the added drift term. We derive:
\begin{align*}
& h(t+1)-h(t) = r\left(h(t)-h(t-1)\right)-\log(r)\\
& h(t+1)-h(t)=Kr^t-\log r\cdot\sum_{s=0}^{t-1}r^s=K'r^t-\frac{1}{1-r}\log r,\quad\text{where $K=h(1)-h(0)$ and $K'=K+\frac{\log r}{1-r}$.}\\
& \sum_{s=0}^{T-1}(h(s+1)-h(s))=\underbrace{h(T)-h(0)}_{=\log(c_0)}=K'\frac{\left(1-r^T\right)}{1-r}-\frac{T\log r}{1-r}\implies K'=\frac{T\log r}{1-r^T} + \frac{\log(c_0)(1-r)}{1-r^T}\\
& h(t+1)-h(t)=\left(\frac{T\log r}{1-r^T} + \frac{\log(c_0)(1-r)}{1-r^T}\right)r^t -\frac{\log r}{1-r} \\
& h(t)=\frac{-\log r}{1-r}\left(t-T\alpha_T(1-r^t)\right)+\log(c_0)\cdot\alpha_T(1-r^t).
\end{align*}
Bringing the two results together, we obtain:
\begin{align*}
u_t&=r^{-\frac{1}{1-r}\left(t-T\cdot\alpha_T\cdot(1-r^t)\right)}\cdot c_0^{\alpha_T\cdot(1-r^t)}m^{\alpha_T\left(1-r^t\right)}+o\left(m^{\alpha_T\left(1-r^t\right)}\right)
\end{align*}
For ease of the exposition, let us write $u_t=r^{-\frac{\overline{h}(t)}{1-r}}c_0^{f(t)}m^{f(t)} + o\left(m^{f(t)}\right)$. Note that:
\begin{align}
\overline{h}(t)-r\overline{h}(t-1)&=\left(t-T\alpha_T(1-r^t)\right)-r\left(t-1-T\alpha_T(1-r^{t-1})\right)
=(1-r)t+r-\frac{T(1-r)}{1-r^T}\label{eq:hbar_rec}\\
f(t)-rf(t-1)&=\alpha_T(1-r^t)-r\cdot\alpha_T(1-r^{t-1})
=\alpha_T\cdot(1-r)\label{eq:alpha_rec}
\end{align}

We conclude as follows:
\begin{align*}
\left(1-(1-p)\varepsilon)\right)\sum_{t=1}^T A^D_t &= m + \sum_{t=1}^T \frac{A^D_t \cdot\varepsilon\cdot p}{\left(\sum_{s=1}^{t-1} A^D_s +1\right)^r}\\
&=  m + \sum_{t=1}^T \frac{\varepsilon\cdot p\cdot(u_t-u_{t-1})}{u^r_{t-1}}\\
&= m + \varepsilon\cdot p\cdot  \sum_{t=1}^T r^{-\frac{\overline{h}(t)-r\overline{h}(t-1)}{1-r}}\cdot c_0^{f(t)-rf(t-1)}\cdot m^{f(t)-rf(t-1)}+o\left(m^{f(t)-rf(t-1)}\right)\\
&= m +\varepsilon\cdot p\cdot\left(\sum_{t=1}^T \frac{1}{r^t}\right)\cdot\left(\frac{1}{r}\right)^{\frac{r}{1-r}}\cdot r^{T\cdot\alpha_T}\cdot c_0^{\alpha_T\cdot(1-r)}\cdot m^{\alpha_T\cdot(1-r)}+ o\left(m^{\alpha_T\cdot(1-r)}\right)\\
&= m+\varepsilon\cdot p\cdot\zeta_T\cdot m^{\alpha_T\cdot(1-r)}+ o\left(m^{\alpha_T\cdot(1-r)}\right),
\end{align*}
where the fourth equality follows from Equations~\eqref{eq:hbar_rec} and~\eqref{eq:alpha_rec}. The expressions for $A^D_t$ follow immediately from the expression of $u_t$ and the fact that $A^D_t=u_t-u_{t-1}$.
\hfill\Halmos

%\textbf{Case 2: $\bm{r=1}$}

%The recursion reduces to:
%\begin{equation*}
%    u_{t+1} =\frac{u_t^2}{u_{t-1}},\ \forall t=1,\cdots,T-1.
%\end{equation*}

%The leading-order analysis indicates that $f(t+1)-f(t)$ is constant. Since $f(T)-f(0)=1$, we have $f(t+1)-f(t)=1/T$, hence $f(t)=t/T$ for all $t=1,\cdots,T$.

%Turning to the leading-coefficient analysis, we write:
%\[g(t+1) = \frac{g(t)^{2}}{g(t-1)}\implies h(t+1) = 2h(t)-h(t-1)\]
%Since $h(0)=h(T)=0$, we obtain that $h(t)=0$, hence $g(t)=1$ for all $t=1,\cdots,T$.

%Bringing the two results together, we obtain $u_t=m^{t/T}+o\left(m^{t/T}\right)$, and conclude as follows:
%\begin{align*}
%   \left(1-(1-p)\varepsilon)\right)\sum_{t=1}^T A^D_t &= m + \sum_{t=1}^T \frac{\varepsilon\cdot p(u_t-u_{t-1})}{u_{t-1}}\\
%    &= m + \sum_{t=1}^T \varepsilon\cdot pm^{1/T}+o\left(m^{1/T}\right)\\
%    &= m + T \varepsilon\cdot pm^{1/T}+o\left(m^{1/T}\right)
%\end{align*}

%This concludes the proof of the case $r=1$, hence of the lemma.

\subsubsection{Proof of Remark~\ref{rem:lm}}

Per Lemma \ref{lem:main_det}, we have:
\[\sum_{t=1}^T r^{-\frac{1}{1-r}\left(t-T\cdot\alpha_T\cdot(1-r^t)\right)}\cdot c_0^{\alpha_T\cdot(1-r^t)}\cdot m^{\alpha_T\cdot(1-r^t)}= c_0\left(m+\varepsilon\cdot p\cdot\zeta_T\cdot m^{\alpha_T\cdot(1-r)} + o\left(m^{\alpha_T \cdot (1-r)}\right)\right) \]

\[\sum_{t=1}^T r^{-\frac{1}{1-r}\left(t-T\cdot\alpha_T\cdot(1-r^t)\right)}\cdot c_0^{\alpha_T\cdot(1-r^t)}\cdot m^{\alpha_T\cdot(1-r^t)}\geq c_0\left(m+\varepsilon\cdot p\cdot\zeta_T\cdot m^{\alpha_T\cdot(1-r)}\right) \]

The derivative of the last term of the left-hand side with respect to $m$ is equal to $c_0$, so the derivative of the left-hand side is at least equal to $c_0$. Therefore, for $m$ large enough:
\begin{align*}
& \sum_{t=1}^T r^{-\frac{1}{1-r}\left(t-T\cdot\alpha_T\cdot(1-r^t)\right)}\cdot c_0^{\alpha_T\cdot(1-r^t)}\cdot (m+\Delta(m))^{\alpha_T\cdot(1-r^t)}\geq c_0\left(m+\varepsilon\cdot p\cdot\zeta_T\cdot m^{\alpha_T\cdot(1-r)} + \Delta(m)\right)\\
& \sum_{t=1}^T r^{-\frac{1}{1-r}\left(t-T\cdot\alpha_T\cdot(1-r^t)\right)}\cdot c_0^{\alpha_T\cdot(1-r^t)}\cdot \left(m-m^{\alpha_T\cdot(1-r)}\right)^{\alpha_T\cdot(1-r^t)}\leq c_0\left(m+\varepsilon\cdot p\cdot\zeta_T\cdot m^{\alpha_T\cdot(1-r)} \right)
\end{align*}

This implies that $\ell \leq m + \Delta(m)$ and $\ell\geq m-m^{\alpha_T\cdot(1-r)}$.
\hfill\Halmos

\subsubsection{Proof of Lemma~\ref{lem:main_feasible}}

We make use of the following lemma, which provides a lower bound on the expected number of facilities that get successfully opened.

\begin{lemma}\label{lem:main_bound}
We have:
\begin{align*}
    \E\left[\sum_{t=1}^T B^\dagger_t\right] \geq \begin{cases}
        m + \sqrt{\frac{c_0}{2}\log \frac{2}{\delta}}\sqrt{m} - \varepsilon\cdot p\cdot \alpha_T\cdot\zeta_T\cdot(1-r)\sqrt{\frac{c_0}{2}\log \frac{2}{\delta}}m^{\alpha_T\cdot (1-r)-\frac{1}{2}} & \text{if }p\neq1\\
         m - \Phi^{-1}(\delta) \sqrt{2\varepsilon \zeta_T} m^{1/2 \cdot \alpha_T \cdot (1-r)} + \varepsilon\cdot p\cdot \alpha_T\cdot\zeta_T\cdot(1-r)\Phi^{-1}(\delta) \sqrt{2\varepsilon \zeta_T} m^{3/2 \cdot \alpha_T \cdot (1-r)-1} & \text{if }p=1.
    \end{cases}
\end{align*}
\end{lemma}
% \begin{comment}
% Furthermore, we have:
% \begin{align*}
%     \E\left[\sum_{t=1}^T B^\dagger_t\right] \leq \begin{cases}
%         m + \sqrt{\frac{c_0}{2}\log \frac{2}{\delta}}\sqrt{m} +\calO\left(m^{\alpha_T \cdot (1-r)-1}\right) & \text{if }p\neq1\\
%          m + \Phi^{-1}(\delta) \sqrt{2\varepsilon \zeta_T} m^{1/2 \cdot \alpha_T \cdot (1-r)} + \O\left(m^{\alpha_T \cdot (1-r)-1}\right) & \text{if }p=1.
%     \end{cases}
% \end{align*}
% \end{comment}

\proof{Proof:}
By definition of $B_t$ and by construction of $A^\dagger_t$, we have:
\begin{align*}
\E\left[\sum_{t=1}^T B^\dagger_t\right] & =  \frac{1}{c_0}\sum_{t=1}^T A^\dagger_t - 
    \sum_{t=1}^T \frac{A^\dagger_t\cdot \varepsilon\cdot p }{\left(\sum_{s=1}^{t-1} A^\dagger_s +1\right)^r}\\
    & = m + \varepsilon\cdot p\cdot\zeta_T\cdot m^{\alpha_T\cdot(1-r)} + \Delta(m)
     - \sum_{t=1}^T \frac{A^\dagger_t\cdot \varepsilon\cdot p}{\left(\sum_{s=1}^{t-1} A^\dagger_s +1\right)^r}
\end{align*}

We then bound the last term as follows:
\begin{align*}
\sum_{t=1}^T \frac{A^\dagger_t}{\left(\sum_{s=1}^{t-1} A^\dagger_s +1\right)^r}
&\leq \sum_{t=1}^T \frac{r^{-\frac{1}{1-r}\left(t-T\cdot\alpha_T\cdot(1-r^t)\right)}\cdot c_0^{\alpha_T\cdot(1-r^t)}\cdot \ell^{\alpha_T(1-r^t)}}{\left(r^{-\frac{1}{1-r}\left(t-1-T\cdot\alpha_T\cdot(1-r^{t-1})\right)}\cdot c_0^{\alpha_T\cdot(1-r^{t-1})}\cdot \ell^{\alpha_T(1-r^{t-1})}\right)^r}\\
&= \left(\sum_{t=1}^T r^{-t}\right)r^{-\frac{r}{1-r}}r^{T\cdot \alpha_T}c_0^{\alpha_T\cdot (1-r)}\ell^{\alpha_T\cdot (1-r)}\\
&= \zeta_T\cdot\ell^{\alpha_T\cdot (1-r)}\\
&\leq \zeta_T \cdot \left(m+ \Delta(m)\right)^{\alpha_T\cdot (1-r)}
\\&\leq  \zeta_T \cdot m^{\alpha_T \cdot (1-r)} +  \alpha_T\zeta_T(1-r) \Delta(m)m^{\alpha_T \cdot (1-r)-1}
\end{align*}
where the first inequality comes from bounding the sum by the term associated with $s=t-1$, the next equalities follows from algebraic manipulations, the next inequality follows from Remark \ref{rem:lm}, and the last inequality stems from the fact that $(1+a)^\delta\leq1+\delta a$ for $\delta<1$.

By plugging this bound into the previous equation, we obtain:
\begin{align*}
\E\left[\sum_{t=1}^T B_t\right]
&\geq  m + \varepsilon\cdot p\cdot\zeta_T\cdot m^{\alpha_T\cdot(1-r)} + \Delta(m)-\varepsilon\cdot p \cdot \left(\zeta_T \cdot m^{\alpha_T \cdot (1-r)} + \alpha_T\zeta_T(1-r) \Delta(m)m^{\alpha_T \cdot (1-r)-1}\right)\\
 &= m + \Delta(m) - \varepsilon\cdot p\cdot \alpha_T\zeta_T(1-r) \Delta(m)m^{\alpha_T \cdot (1-r)-1}
\end{align*}
This completes the proof by plugging the values of $\Delta(m)$ into the bound.
% \begin{comment}
% The proof of of the first part of Lemma~\ref{lem:main_bound} is completed by plugging the expression of $\Delta(m)$. For the upper bounds, we can note that, for sufficiently large $m$, we have:
% \begin{align*}
%     (\star)
%     &\geq  \sum_{t=1}^T \frac{1}{2} \frac{r^{-\frac{1}{1-r}\left(t-T\cdot\alpha_T\cdot(1-r^t)\right)}\cdot c_0^{\alpha_T\cdot(1-r^t)}\cdot \ell^{\alpha_T(1-r^t)}}{\left(r^{-\frac{1}{1-r}\left(t-1-T\cdot\alpha_T\cdot(1-r^{t-1})\right)}\cdot c_0^{\alpha_T\cdot(1-r^{t-1})}\cdot \ell^{\alpha_T(1-r^{t-1})}\right)^r}\\
%     &= \frac{1}{2}\frac{\frac{1}{r^T}-1}{1-r}r^{-\frac{r}{1-r}}r^{T\cdot \alpha_T}c_0^{\alpha_T\cdot (1-r)}\ell^{\alpha_T\cdot (1-r)}\\
%     &\geq \frac{1}{2}\frac{1}{2}\zeta_T \cdot \left(m- m^{\alpha_T \cdot  (1-r)}\right)^{\alpha_T\cdot (1-r)}
%     \\&=  \frac{1}{2}\zeta_T \cdot m^{\alpha_T \cdot (1-r)} + \calO\left(m^{\alpha_T \cdot (1-r)-1}\right)
% \end{align*}
% where the second inequality comes from Remark \ref{rem:lm}. Therefore, we have that:
% \begin{align*}
%     \E\left[\sum_{t=1}^T B_t\right]
%     &\leq   m + \varepsilon\cdot p\cdot\zeta_T\cdot m^{\alpha_T\cdot(1-r)} + \Delta(m)-\varepsilon\cdot p \cdot \left(\zeta_T \cdot m^{\alpha_T \cdot (1-r)} +O\left(m^{\alpha_T \cdot (1-r)-1}\right)\right)\\
%     &=m + \Delta(m) - O\left(m^{\alpha_T \cdot (1-r)-1}\right)
% \end{align*}
% which gives the required expressions once we plug the expression of $\Delta(m)$. 
% \end{comment}
\hfill\Halmos
\endproof 

\subsubsection*{Proof of Lemma~\ref{lem:main_feasible}.}\

\underline{Case 1: $p\neq 1$.}

Note that $0\leq B^\dagger_t\leq A^\dagger_t$ for all $t=1,\cdots,T$. By Hoeffding's inequality, we have:
\begin{align*}
\P\left(N^\dagger_T<m\right)
&=\P\left(\E\left[\sum_{t=1}^T B^\dagger_t\right]-\sum_{t=1}^T B^\dagger_t < \E\left[\sum_{t=1}^T B^\dagger_t\right] - m\right)\\
&\leq\P\left(\left|\sum_{t=1}^T B^\dagger_t - \E\left[\sum_{t=1}^T B^\dagger_t\right]\right|>\left|\E\left[\sum_{t=1}^T B^\dagger_t\right]-m\right|\right)\\
&\leq2 \exp\left(-2\frac{\left(\E\left[\sum_{t=1}^T B^\dagger_t\right]-m\right)^2}{\sum_{t=1}^T A^\dagger_t}\right)
\end{align*}
By Lemma~\ref{lem:main_bound} and by construction of $A^\dagger_t$, this yields:
\begin{align*}
\P\left(N^\dagger_T<m\right)
&\leq 2 \exp\left(-2\frac{\left(\sqrt{\frac{c_0}{2}\log \frac{2}{\delta}}\sqrt{m} - \varepsilon\cdot p \alpha_T\zeta_T(1-r)\sqrt{\frac{c_0}{2}\log \frac{2}{\delta}}m^{\alpha_T\cdot (1-r)-\frac{1}{2}}\right)^2}{c_0\left(m+\varepsilon\cdot p \cdot \zeta_T \cdot m^{\alpha_T \cdot(1-r)}+ \sqrt{\frac{c_0}{2}\log \frac{2}{\delta}}\sqrt{m}\right)}\right)
\end{align*}

By taking the limit as $m$ goes to infinity, we conclude:
\begin{align*}
\lim_{m \to \infty}\P\left(N^\dagger_T<m\right)
&\leq 2 \exp\left(-2\left(\sqrt{\frac{c_0}{2}\log \frac{2}{\delta}}\right)^2\frac{1}{c_0}\right)
= \delta
\end{align*}

\underline{Case 2: $p=1$.}

Since $B_t$ follows a binomial distribution with $A_t$ trials and failure probability $\frac{\varepsilon}{\left(\sum_{s=1}^{t-1} A^\dagger_s +1\right)^r}$,
\begin{align*}
\Var(B^\dagger_t)&=A^\dagger_t \left(1 - \frac{\varepsilon}{\left(\sum_{s=1}^{t-1} A^\dagger_s +1\right)^r} \right) \frac{\varepsilon}{\left(\sum_{s=1}^{t-1} A^\dagger_s +1\right)^r}\\
\E(B^\dagger_t-E(B^\dagger_t))^3&=A^\dagger_t \left(1 - \frac{\varepsilon}{\left(\sum_{s=1}^{t-1} A^\dagger_s +1\right)^r} \right) \frac{\varepsilon}{\left(\sum_{s=1}^{t-1} A^\dagger_s +1\right)^r}\left(2\frac{\varepsilon}{\left(\sum_{s=1}^{t-1} A^\dagger_s +1\right)^r}-1\right)
\end{align*}

Moreover, since the variables $B_1,\cdots,B_T$ are independent, we have:
\begin{align*}
\Var\left(\sum_{t=1}^T B^\dagger_t\right)
=&\sum_{t=1}^T \Var(B^\dagger_t)\\
\leq&\sum_{t=1}^T A^\dagger_t \frac{\varepsilon}{\left(\sum_{s=1}^{t-1} A^\dagger_s +1\right)^r}
\\
\leq & \varepsilon\left(\zeta_T \cdot m^{\alpha_T \cdot (1-r)} -  \alpha_T\zeta_T(1-r) \Phi^{-1}(\delta) \sqrt{2\varepsilon \zeta_T} m^{3/2 \cdot \alpha_T \cdot (1-r)-1}\right)\\
\leq & 2\varepsilon\zeta_T \cdot m^{\alpha_T \cdot (1-r)},
\end{align*}
where the second inequality follows from the same procedure in the proof of Lemma \ref{lem:main_bound} and the last inequality holds for $m$ large enough. Similarly, we can also prove that:
\begin{align*}
\Var\left(\sum_{t=1}^T B^\dagger_t\right)
= &\sum_{t=1}^T \Var(B^\dagger_t)\\
\geq&\frac{1}{2}\sum_{t=1}^T A^\dagger_t \frac{\varepsilon}{\left(\sum_{s=1}^{t-1} A^\dagger_s +1\right)^r}
\\
= &\frac{1}{2}\sum_{t=1}^T   \frac{\varepsilon r^{-\frac{1}{1-r}\left(t-T\cdot\alpha_T\cdot(1-r^t)\right)}\cdot c_0^{\alpha_T\cdot(1-r^t)}\cdot \ell^{\alpha_T(1-r^t)}}{\left(\sum_{s=1}^{t-1} r^{-\frac{1}{1-r}\left(s-T\cdot\alpha_T\cdot(1-r^s)\right)}\cdot c_0^{\alpha_T\cdot(1-r^s)}\cdot \ell^{\alpha_T(1-r^s)}  +1\right)^r}\\
\geq & \frac{1}{2}   \frac{\varepsilon c_0 \ell}{\left(\sum_{s=1}^{T-1} r^{-\frac{1}{1-r}\left(s-T\cdot\alpha_T\cdot(1-r^s)\right)}\cdot c_0^{\alpha_T\cdot(1-r^s)}\cdot \ell^{\alpha_T(1-r^s)}  +1\right)^r}\\
=& \frac{1}{2}   \frac{\varepsilon c_0 \ell}{c_0^r\left(m+\varepsilon\cdot p \cdot \zeta_T \cdot m^{\alpha_T \cdot(1-r)}+\Delta(m)-\ell\right)^r}\\
\geq& \frac{1}{2}   \frac{\varepsilon c_0 \left(m-m^{\alpha_T(1-r)}\right)}{c_0^r\left(m+\varepsilon\cdot p \cdot \zeta_T \cdot m^{\alpha_T \cdot(1-r)}+\Delta(m)-\left(m-m^{\alpha_T(1-r)}\right)\right)^r}\\
=& \Omega(m^{1-\alpha_T \cdot r(1-r)}),
\end{align*}
where the last inequality comes from Remark~\ref{rem:lm}.

From the Berry-Esseen theorem, we obtain:
\begin{align*}
&\P\left(\sum_{t=1}^T B^\dagger_t - \E\left[\sum_{t=1}^T B^\dagger_t\right] \leq -k\sqrt{\sum_{t=1}^T \Var(B^\dagger_t)}\right)\\
&\leq \Phi(-k) + \calO\left(\frac{\sum_{t=1}^T A^\dagger_t \left(1 - \frac{\varepsilon}{\left(\sum_{s=1}^{t-1} A^\dagger_s +1\right)^r} \right) \frac{\varepsilon}{\left(\sum_{s=1}^{t-1} A^\dagger_s +1\right)^r} \left( \frac{2\varepsilon}{\left(\sum_{s=1}^{t-1} A^\dagger_s +1\right)^r} -1 \right)}{\left(\sum_{t=1}^T A^\dagger_t \left(1 - \frac{\varepsilon}{\left(\sum_{s=1}^{t-1} A^\dagger_s +1\right)^r} \right) \frac{\varepsilon}{\left(\sum_{s=1}^{t-1} A^\dagger_s +1\right)^r} \right)^{3/2}}\right) \\
&= \Phi(-k) + \calO\left(\left(\sum_{t=1}^T A^\dagger_t \left(1 - \frac{\varepsilon}{\left(\sum_{s=1}^{t-1} A^\dagger_s +1\right)^r} \right) \frac{\varepsilon}{\left(\sum_{s=1}^{t-1} A^\dagger_s +1\right)^r} \right)^{-1/2}\right) \\
&= \Phi(-k) + \calO\left(m^{-1/2\left(1-\alpha_T \cdot r(1-r)\right)}\right),
\end{align*}
where the last equality stems from the lower bound on the variance term. For sufficiently large $m$, we get, using the upper bound on the variance term:
\begin{equation}
\P\left(\sum_{t=1}^T B^\dagger_t - \E\left[\sum_{t=1}^T B^\dagger_t\right] \leq -k\sqrt{2\varepsilon \zeta_T \cdot m^{\alpha_T \cdot (1-r)}}\right) \leq \Phi(-k) \label{eq:chernoff_bound}
\end{equation}
Then, we have:
\begin{align*}
\P\left(N^\dagger_T<m\right)
&=\P\left(\sum_{t=1}^T B^\dagger_t - \E\left[\sum_{t=1}^T B^\dagger_t\right] < m - \E\left[\sum_{t=1}^T B^\dagger_t\right]\right)
\leq\Phi\left(-\frac{ \E\left[\sum_{t=1}^T B^\dagger_t\right]-m }{\sqrt{2 \varepsilon \zeta_T \cdot m^{\alpha_T \cdot (1-r)}}}\right)
\end{align*}
By Lemma~\ref{lem:main_bound}, since $\E\left[\sum_{t=1}^T B^\dagger_t\right]\geq m + \Delta(m)- \varepsilon\cdot p\cdot \alpha_T\zeta_T(1-r) \Delta(m)m^{\alpha_T \cdot (1-r)-1}$ this yields:
\begin{align*}
\P\left(N^\dagger_T<m\right)
&\leq \Phi\left(-\frac{-\Phi^{-1}(\delta)\sqrt{2 \varepsilon \zeta_T }\cdot m^{1/2\cdot \alpha_T \cdot (1-r)} + \varepsilon\cdot p\cdot \alpha_T\cdot\zeta_T\cdot(1-r)\Phi^{-1}(\delta) \sqrt{2\varepsilon \zeta_T} m^{3/2 \cdot \alpha_T \cdot (1-r)-1}}{\sqrt{2 \varepsilon \zeta_T \cdot m^{\alpha_T \cdot (1-r)}}}\right)
\end{align*}

By taking the limit as $m$ goes to infinity, we conclude: $\lim_{m \to \infty}\P\left(N^\dagger_T<m\right)\leq\delta$
\hfill\Halmos

\subsubsection{Proof of Lemma~\ref{lem:main_lower}}
\label{app:unif_prob_bin_determin_sto}

\underline{Case 1: $p=1$.}

Let $k$ be such that $\frac{1}{2}\alpha_T\cdot(1-r)<k<\alpha_T\cdot(1-r)$. Let us define the following ``good'' event:
$$\calE=\left\{\left|B_t - A_t \left(1 -\frac{\varepsilon}{\left(\sum_{s=1}^{t-1} A_s +1\right)^r}\right)\right|\leq m^k,\ \forall t \in \{1,\cdots, T\}\right\}$$

\begin{lemma}\label{lem:hoeffding_bound}
Assume that $p=1$. If $A_t\leq m+ \calO(m^{\alpha_T \cdot (1-r)})$ for all $t=1,\cdots, T$, then $\P(\calE^c) =\calO(\exp\left(-m^{2k-\alpha_T\cdot(1-r)}\right))$ as $m \to \infty$.
\end{lemma}
\proof{Proof:}
By construction, $\mathbb{E}[B_t] = A_t \left(1 -\frac{\varepsilon}{\left(\sum_{s=1}^{t-1} A_s +1\right)^r}\right)\leq A_t$. From Bernstein's inequality, using the definition of $B_t=\sum_{i:y_{it}=1}S_{it}$:
\begin{align*}
    \P(|B_t - \mathbb{E}[B_t]| \geq m^k)
    &\leq 2\exp\left(-\frac{1/2\cdot m^{2k}}{\sum_{i:y_{it}=1}\text{Var}(S_{it})+1/3\cdot m^k}\right)\\
    &= 2\exp\left(-\frac{1/2\cdot m^{2k}}{\text{Var}(B_t)+1/3\cdot m^k}\right)\\
    &\leq 2\exp\left(-\frac{m^{2k}}{4\varepsilon \zeta_T m^{\alpha_T\cdot(1-r)} + 2m^k/3}\right)\\
    &= \calO\left(\exp\left(-m^{2k-\alpha_T\cdot(1-r)}\right)\right),
\end{align*}
where the equality follows from the independence of $B_1,\cdots,B_T$ and the second inequality uses the lower bound on the variance term from the proof of Lemma~\ref{lem:main_feasible}. By assumption, $2k - \alpha_T\cdot(1-r) > 0$, so $\P\left(|B_t - \mathbb{E}[B_t]| \geq m^k\right) \to 0$ as $m \to \infty$. We obtain from the union bound:
\begin{align*}
\P(\calE^c)
&=\P\left(\exists t \in \{1,\cdots, T\} \;\; : \;\; \left|B_t - A_t \left(1 -\frac{\varepsilon}{\left(\sum_{s=1}^{t-1} A_s +1\right)^r}\right)\right|\geq m^k\right)\\
&\leq \sum_{t=1}^T \P\left(\left|B_t - A_t \left(1 -\frac{\varepsilon}{\left(\sum_{s=1}^{t-1} A_s +1\right)^r}\right)\right|\geq m^k\right)\\
& =\calO(\exp\left(-m^{2k-\alpha_T\cdot(1-r)}\right))
\end{align*}
This completes the proof of Lemma~\ref{lem:hoeffding_bound}.\hfill\Halmos\endproof

\begin{lemma}\label{lem:sum_bound}
Assume that $p=1$. Let $(B^*_1,\cdots,B^*_T)$ be a solution of Problem~\eqref{prob:main}. If $\delta = o\left(m^{\alpha_T\cdot(1-r)-1}\right)$, then as $m \to \infty$:
\begin{align*}
\sum_{\sigma=0}^\infty \sigma \P\left(\sum_{t=1}^T B_t^*=\sigma \mid \calE\right) & \geq m + o\left(m^{\alpha_T\cdot(1-r)}\right)\\
\sum_{\sigma=0}^\infty(\sigma-Tm^k)^{\alpha_T\cdot(1-r)}\P\left(\sum_{t=1}^T B_t^*=\sigma\mid \calE \right)&\geq m^{\alpha_T\cdot(1-r)}+o\left(m^{\alpha_T\cdot(1-r)}\right)
\end{align*}
\end{lemma}
\proof{Proof:}
Note that the optimal solution of Problem~\eqref{prob:main} must satisfy the constraint:
\begin{equation}
\P\left(\sum_{t=1}^T B_t^*\geq m\right)\geq 1-\delta = 1-o\left(m^{\alpha_T\cdot(1-r)-1}\right)\label{eq:Bmdelta}
\end{equation}
From the law of total probabilities, we obtain, using Equation~\eqref{eq:Bmdelta} and Lemma~\ref{lem:hoeffding_bound}:
\begin{align*}
\P\left(\sum_{t=1}^T B_t^*\geq m \mid \calE\right)
&=\frac{\P\left(\sum_{t=1}^T B_t^*\geq m\right)-\P\left(\sum_{t=1}^T B_t^*\geq m\mid \calE^c \right)\P(\calE^c)}{\P(\calE)}\\
&\geq1-o\left(m^{\alpha_T\cdot(1-r)-1}\right)-\calO\left(\exp\left(-m^{2k-\alpha_T\cdot(1-r)}\right)\right)\\
&=1-o\left(m^{\alpha_T\cdot(1-r)-1}\right)
\end{align*}
Then, we have:
\begin{equation*}
\sum_{\sigma=0}^\infty \sigma \P\left(\sum_{t=1}^T B_t^*=\sigma \mid \calE\right)\geq m\cdot \P\left(\sum_{t=1}^T B_t^*\geq m \mid \calE\right)\geq m+o\left(m^{\alpha_T\cdot(1-r)}\right)
\end{equation*}

Similarly:
\begin{align*}
&\sum_{\sigma=0}^\infty(\sigma-Tm^k)^{\alpha_T\cdot(1-r)}\P\left(\sum_{t=1}^T B_t^*
=\sigma\mid \calE \right)\\
&\geq (m-Tm^k)^{\alpha_T\cdot(1-r)}\P\left(\sum_{t=1}^T B_t^*\geq m\mid \calE \right)\\
&= m^{\alpha_T\cdot(1-r)}(1-Tm^{k-1})^{\alpha_T\cdot(1-r)}\left(1-o\left(m^{\alpha_T\cdot(1-r)-1}\right)\right)\\
&= m^{\alpha_T\cdot(1-r)}\left(1-T^{\alpha_T\cdot(1-r)}m^{(k-1)\alpha_T\cdot(1-r)}+o\left(m^{(k-1)\alpha_T\cdot(1-r)}\right)\right)\left(1-o\left(m^{\alpha_T\cdot(1-r)-1}\right)\right)\\
&=m^{\alpha_T\cdot(1-r)}+o(m^{\alpha_T\cdot(1-r)}),
\end{align*}
where the first equality results from Lemma~\ref{lem:hoeffding_bound}, the second equality follows from Taylor expansion and the last equality follows from the assumption that $k<\alpha_T(1-r)\leq1$.
This completes the proof of Lemma~\ref{lem:sum_bound}.\hfill\Halmos

Now, let $(A^*_1,\cdots,A^*_T)$ and $(B^*_1,\cdots,B^*_T)$ denote the optimal solution of Problem~\eqref{prob:main}. We obtain from Lemma~\ref{lem:hoeffding_bound}:
\begin{align*}
\E\left[\sum_{t=1}^T A^*_t\right]
&=\E\left[\sum_{t=1}^T A^*_t \mid \calE\right]\P(\calE)+\E\left[\sum_{t=1}^T A^*_t \mid \calE^c\right]\P(\calE^c)\\
&\geq \E\left[\sum_{t=1}^T A^*_t \mid \calE\right]\left(1-\calO(\exp(-m^{2k-\alpha_T\cdot(1-r)}))\right)\\
&=\E\left[\sum_{t=1}^T A^*_t \mid \calE\right] + o(1)
\end{align*}

Conditioned on $\calE$, we have:
$$\sum_{t=1}^T B^*_t=\left|\sum_{t=1}^T \E(B^*_t)-\sum_{t=1}^T (B^*_t-\E(B^*_t))\right|\leq\sum_{t=1}^T \E(B^*_t)+\sum_{t=1}^T |B^*_t-\E(B^*_t)|\leq\sum_{t=1}^T \E(B^*_t)+Tm^k$$

Therefore, the following holds:
\begin{align*}
\E\left[\sum_{t=1}^T A^*_t\mid \calE\right]
&=\sum_{\sigma=0}^\infty \E\left[\sum_{t=1}^T A^*_t \mid \calE, \sum_{t=1}^T B^*_t=\sigma\right]\P\left(\sum_{t=1}^T B^*_t=\sigma \mid \calE \right)\\
&=\sum_{\sigma=0}^\infty\E\left[\sum_{t=1}^T A^*_t \mid \calE, \sum_{t=1}^T B^*_t=\sigma, \sum_{t=1}^T A^*_t \left(1- \frac{\varepsilon}{\left(\sum_{s=1}^{t-1} A^*_s +1\right)^r}\right)\geq \sigma-Tm^k\right]\\&\qquad\qquad\cdot\P\left(\sum_{t=1}^T B^*_t=\sigma\mid \calE \right)
\end{align*}
The first expression resembles the deterministic variant of the full problem, with a right-hand side target of $\sigma-Tm^k$. From Lemma~\ref{lem:main_det}, we can lower bound it to obtain:
\begin{align*}
\E\left[\sum_{t=1}^T A^*_t\mid \calE\right]
&\geq c_0\sum_{\sigma=0}^\infty\left(\sigma-Tm^k+\varepsilon\cdot\zeta_T\cdot \left(\sigma-Tm^k\right)^{\alpha_T\cdot(1-r)}+ o\left(\left(\sigma-Tm^k\right)^{\alpha_T\cdot(1-r)}\right)\right)\P\left(\sum_{t=1}^T B^*_t=\sigma\mid \calE\right)\\
&+ o(1)\\
&\geq c_0\sum_{\sigma=0}^\infty\sigma\P\left(\sum_{t=1}^T B^*_t=\sigma\mid \calE\right)+c_0\cdot\varepsilon\cdot\zeta_T\cdot\sum_{\sigma=0}^\infty(\sigma-Tm^k)^{\alpha_T\cdot(1-r)}\P\left(\sum_{t=1}^T B^*_t=\sigma\mid \calE\right)\\
&+ o\left(m^{\alpha_T\cdot(1-r)}\right)\quad\text{(because $k<\alpha_T(1-r)$)}\\
&\geq c_0\left( m+\varepsilon\cdot \zeta_T \cdot m^{\alpha_T\cdot(1-r)} +  o\left(m^{\alpha_T\cdot(1-r)}\right)\right)
\quad\text{(Lemma~\ref{lem:sum_bound})}
\end{align*}
The proof is completed since $\E\left[\sum_{t=1}^T A^*_t\right]\geq\E\left[\sum_{t=1}^T A^*_t \mid \calE\right] + o(1)$. 

\underline{Case 2: $p\neq 1$.}

First, we could follow the same logic as in the proof for $p=1$, and get:
\[m^*(\varepsilon,p,T) \geq c_0\left( m+\varepsilon\cdot p\cdot \zeta_T \cdot m^{\alpha_T\cdot(1-r)} +  o\left(m^{\alpha_T\cdot(1-r)}\right)\right)\]

This bound is tight if $\alpha_T(1-r)\geq1/2$, since the upper-bounding solution admits a buffer of $\Theta(m^{\alpha_T(1-r)})$ in that case; however, if $\alpha_T(1-r)<1/2$, the bound may no longer be tight since the the upper-bounding buffer then grows in $\Theta(\sqrt{m})$. To improve it, we assume by contradiction that:
\[m^*(\varepsilon,p,T)=\sum_{t=1}^T A^*_t <c_0\left(m -\Phi^{-1}(\delta)\sqrt{2\varepsilon(1-p)m}\right)\]
Consider independent trials $S'_{it} \sim \Bern\left(1-\varepsilon\cdot(1-p)\right)$ and define $B'_t = \sum_{i \in \mathcal{A}_t} S'_{it}$. We have:
\[\P\left(\sum_{t=1}^T B^*_t \geq m\right)\leq \P\left(\sum_{t=1}^T B'_t \geq m\right)\]
Then by the Berry-Esseen Theorem, we have:
\begin{align*}
\P\left(\sum_{t=1}^T B'_t<m \right)
=&\P\left(\sum_{t=1}^T B'_t -\E\left[\sum_{t=1}^T B'_t \right] <m-\sum_{t=1}^T A^*_t (1-\varepsilon(1-p)) \right)
\\=&\Phi\left(\frac{m-\sum_{t=1}^T A^*_t (1-\varepsilon(1-p))}{\sqrt{\sum_{t=1}^T A^*_t (1-\varepsilon(1-p))\varepsilon(1-p)}}\right) + \calO\left(\frac{\sum_{t=1}^TA^*_t(1-\varepsilon(1-p))\varepsilon(1-p)(2\varepsilon(1-p)-1)}{\left(\sum_{t=1}^T A^*_t (1-\varepsilon(1-p))\varepsilon(1-p)\right)^{3/2}}\right)
\\=&\Phi\left(\frac{m-\sum_{t=1}^T A^*_t (1-\varepsilon(1-p))}{\sqrt{\sum_{t=1}^T A^*_t (1-\varepsilon(1-p))\varepsilon(1-p)}}\right) + o(1)
\end{align*}
We use the contradiction assumption $\sum_{t=1}^T A^*_t < c_0\left(m -\Phi^{-1}(\delta)\sqrt{2\varepsilon(1-p)m}\right)$ to bound the numerator and the fact that $\sum_{t=1}^T A^*_t<2c_0m$ to bound the denominator. We obtain
\begin{align*}
\P\left(\sum_{t=1}^T B'_t<m \right)> &\Phi\left(\frac{\Phi^{-1}(\delta)\sqrt{2\varepsilon(1-p)m}}{\sqrt{2\varepsilon(1-p)m}}\right) + o(1)   =\delta +o(1)
\end{align*}
Therefore, for sufficiently large $m$:
\[\P\left(\sum_{t=1}^T B^*_t \geq m\right)\leq \P\left(\sum_{t=1}^T B'_t \geq m\right)<1-\delta\]
This contradicts the fact that $A^*_t$ is a feasible (and optimal) solution to Problem~\eqref{prob:main}. Therefore, for sufficiently large $m$:
\[m^*(\varepsilon, p,T)=\sum_{t=1} A_t^* \geq c_0\left(m -\Phi^{-1}(\delta)\sqrt{2\varepsilon(1-p)m}\right)\]
Combining both bounds gives us:
\[m^*(\varepsilon, p,T)=\sum_{t=1} A_t^* \geq c_0\left(m +\max\{-\Phi^{-1}(\delta)\sqrt{2\varepsilon(1-p)m},\varepsilon\cdot p\cdot\zeta_T\cdot m^{\alpha_T\cdot(1-r)}\}+o\left(m^{\alpha_T(1-r)}\right)\right)\]
This completes the proof.
\hfill\Halmos

\subsubsection{Proof of Theorem~\ref{thm:main}}

The main result follows directly from the lemmas. Per Lemma~\ref{lem:main_feasible}, the solution given in Algorithm~\ref{ALG} is feasible in Problem~\eqref{prob:main}. Per Equations~\eqref{eq:solution} and~\eqref{eq:solution_eq}, it achieves an upper bound of:
$$m^\dagger(\varepsilon,p,T)=\begin{cases}
    c_0\left( m+\varepsilon\cdot p\cdot \zeta_T \cdot m^{\alpha_T\cdot(1-r)} +  \sqrt{\frac{c_0}{2}\log \frac{2}{\delta}}\sqrt{m}\right) &  \text{if }p\neq 1\\
c_0\left( m+\varepsilon\cdot p\cdot \zeta_T \cdot m^{\alpha_T\cdot(1-r)} +  o\left(m^{\alpha_T\cdot(1-r)}\right)\right) &  \text{if }p=1
\end{cases}$$
Moreover, since $\delta = o\left(m^{\alpha_T\cdot(1-r)-1}\right)$ by assumption, the optimal solution of Problem~\eqref{prob:main} admits the following lower bound, per Lemma~\ref{lem:main_lower}:
\[m^*(\varepsilon, p,T) \geq   \begin{cases}
c_0\left(m +\max\{-\Phi^{-1}(\delta)\sqrt{2\varepsilon(1-p)m},\varepsilon\cdot p\cdot\zeta_T\cdot m^{\alpha_T\cdot(1-r)}\}\right) +o\left(m^{\alpha_T(1-r)}\right)& \text{if }p \neq 1\\
  c_0\left(m + \varepsilon\cdot p\cdot\zeta_T\cdot m^{\alpha_T\cdot(1-r)} +  o\left(m^{\alpha_T\cdot(1-r)}\right)\right) & \text{if }p=1
\end{cases}\]

Recall from Equation~\eqref{eq:mhat} that
$$\widehat{m}(\varepsilon,p)=c_0m-c_0\cdot\Phi^{-1}(\delta)\sqrt{\varepsilon(1-p)m}+\calO(1)$$

Therefore, if $p=1$ or if ($p\neq1$ and $\alpha_T\cdot(1-r)>1/2$), then:
\begin{align*}
m^\dagger(\varepsilon,p,T)&=c_0m+c_0\cdot\varepsilon\cdot p\cdot \zeta_T \cdot m^{\alpha_T\cdot(1-r)} +  o\left(m^{\alpha_T\cdot(1-r)}\right)\\
m^*(\varepsilon,p,T)&=c_0m+c_0\cdot\varepsilon\cdot p\cdot \zeta_T \cdot m^{\alpha_T\cdot(1-r)} +  o\left(m^{\alpha_T\cdot(1-r)}\right)\\\
\widehat{m}(\varepsilon,p)&=c_0m+  o\left(m^{\alpha_T\cdot(1-r)}\right)
\end{align*}
This proves that:
\begin{align*}
\texttt{Reg}^\dagger(\varepsilon, p,T)&=
    c_0\cdot\varepsilon\cdot p\cdot\zeta_T\cdot m^{\alpha_T\cdot(1-r)}+o\left(m^{\alpha_T\cdot(1-r)}\right)\\
\texttt{Reg}^*(\varepsilon, p,T)&=
    c_0\cdot\varepsilon\cdot p\cdot\zeta_T\cdot m^{\alpha_T\cdot(1-r)}+o\left(m^{\alpha_T\cdot(1-r)}\right)
\end{align*}

If $p\neq1$ and $\alpha_T\cdot(1-r)<1/2$ (the case $\alpha_T\cdot(1-r)=1/2$ is treated similarly by adjusting the second-leading coefficient term of $m^\dagger(\varepsilon,p,T)$), we have:
\begin{align*}
m^\dagger(\varepsilon,p,T)&=c_0m+c_0\cdot\sqrt{\frac{c_0}{2}\log \frac{2}{\delta}}\sqrt{m} +  o\left(\sqrt{m}\right)\\\
m^*(\varepsilon,p,T)&=c_0m-c_0\cdot\Phi^{-1}(\delta)\sqrt{2\varepsilon(1-p)m} +  o\left(\sqrt{m}\right)\\
\widehat{m}(\varepsilon,p)&=c_0m-c_0\cdot\Phi^{-1}(\delta)\sqrt{\varepsilon(1-p)m}+o(\sqrt m)
\end{align*}
This proves that:
\begin{align*}
\texttt{Reg}^\dagger(\varepsilon, p,T)&=
    \Theta(\sqrt m)\\
\texttt{Reg}^*(\varepsilon, p,T)&=
    \Theta(\sqrt m)
\end{align*}

This completes the proof.\hfill\Halmos

\section{Details on the application to real-world datasets}
\label{app:UCI}

We apply the online learning and optimization methodology to the four datasets (bank, default, occupancy, and online) from the \cite{UCI}, as described in Section~\ref{subsec:UCI}. Table~\ref{tab:dataset} reports descriptive statistics on the datasets.

\begin{figure}[h!]
\centering
\subfloat['Bank' dataset.]{\label{fig:bank}\includegraphics[width=.49\textwidth]{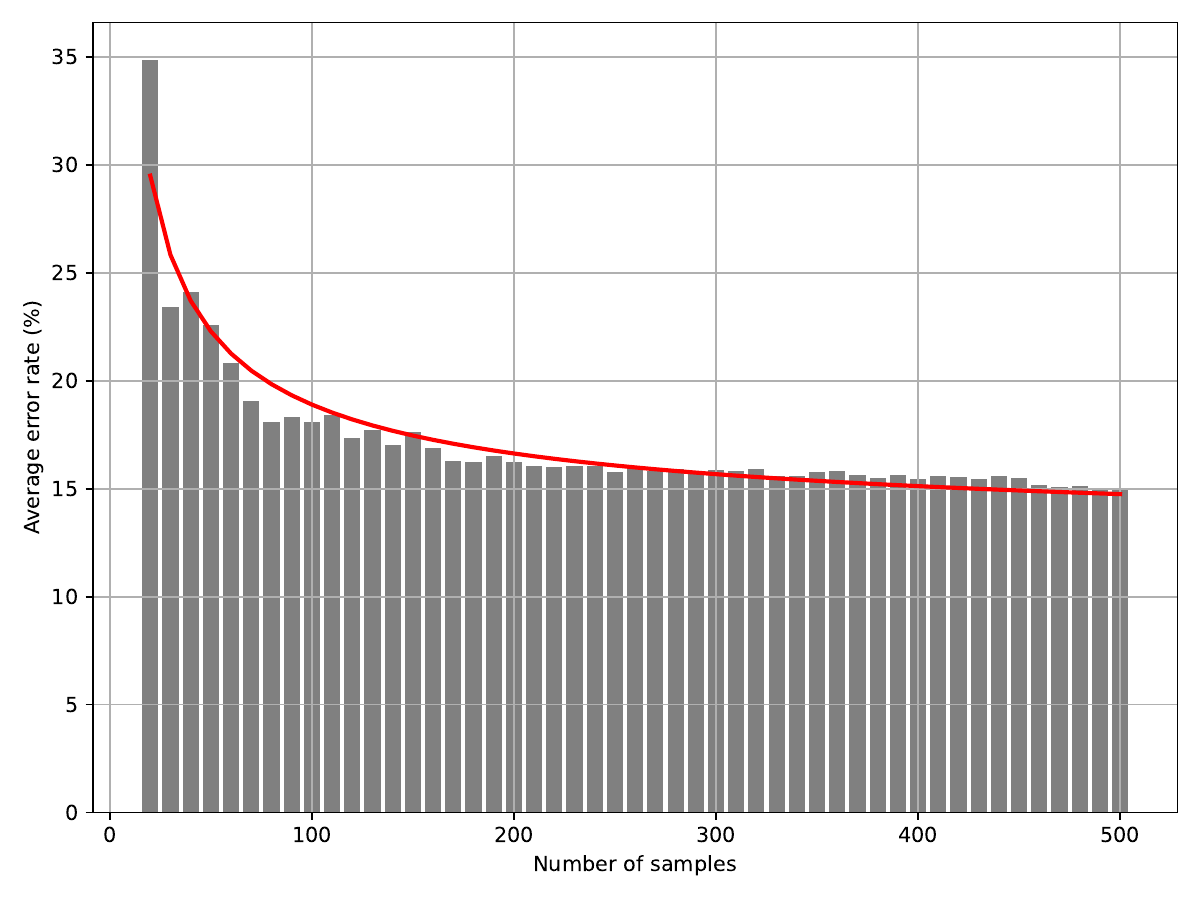}}
\hspace{2pt} 
\subfloat['Default' dataset.]{\label{fig:default}\includegraphics[width=.49\textwidth]{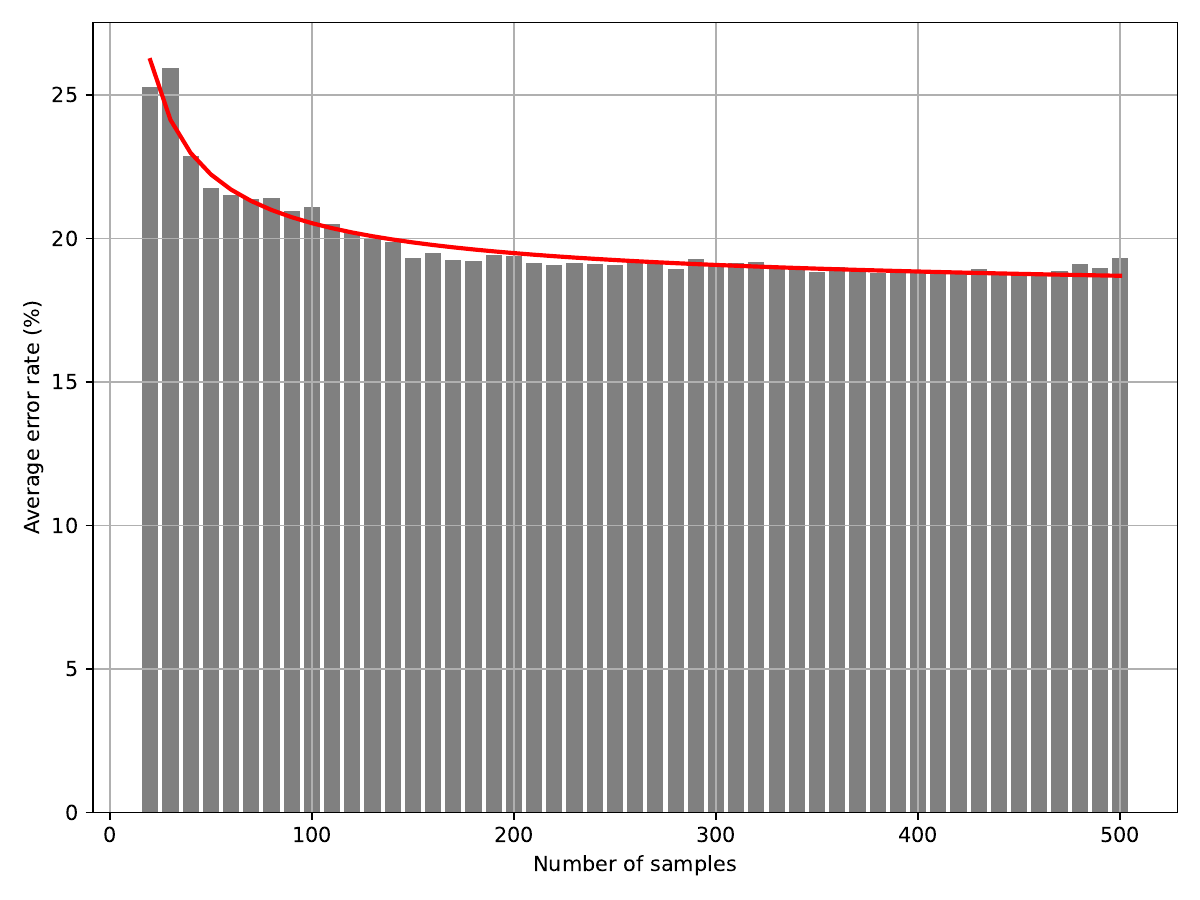}}
\hspace{2pt} 
\subfloat['Occupancy' dataset.]{\label{fig:ccupancy}\includegraphics[width=.49\textwidth]{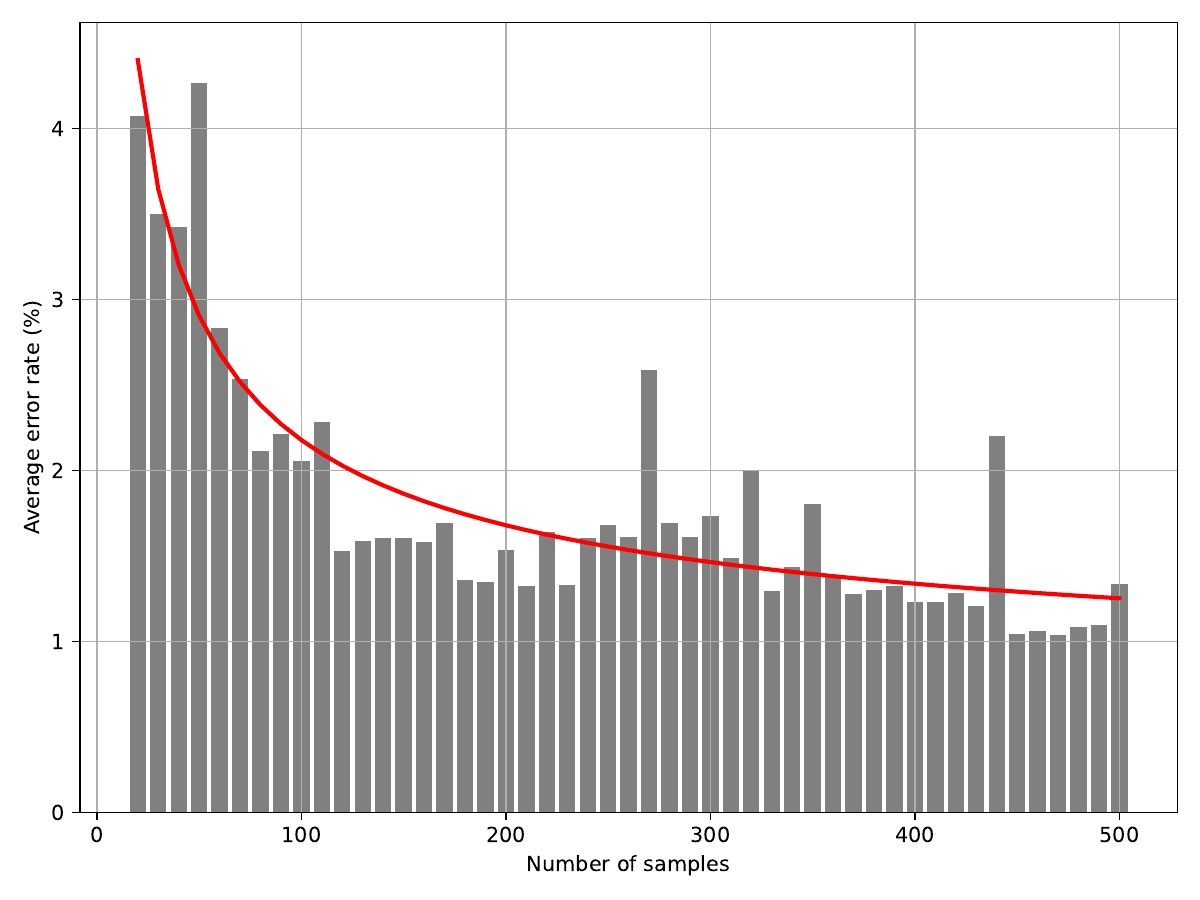}}
\hspace{2pt} 
\subfloat['Online' dataset.]{\label{fig:online}\includegraphics[width=.49\textwidth]{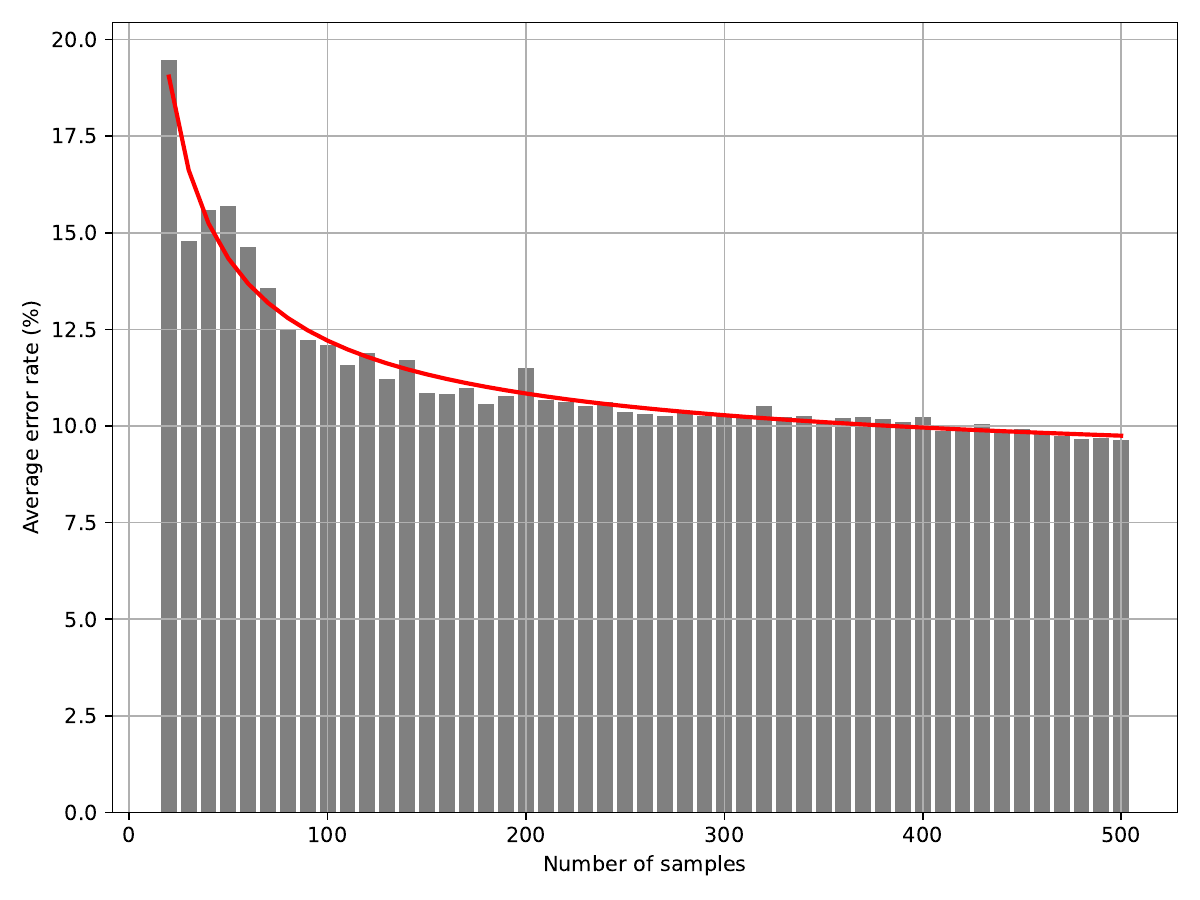}}
\caption{Average error of the classification models as a function of the accrued sample size. The red curve shows the best fit of the form $\frac{\gamma}{n^r}+\beta$.}
\label{fig:ML}
\vspace{-12pt}
\end{figure}

We show the improvement in predictive performance as a function of the accrued sample size under the online learning environment from Section \ref{subsec:DMenvironment}, thereby providing empirical evidence for the learning function in Assumption~\ref{ass:classification} from Proposition~\ref{prop:MLE} and Corollary~\ref{cor:learning}. For every dataset, we train a random forest classification model and follow the learning procedure outlined in Section \ref{subsec:DMenvironment} by iteratively randomly selecting 10 samples with positive predictions from the current classification model to serve as training samples for the next round (Assumption \ref{ass:sampling}, with $\calE=\emptyset$). We repeat the procedure 10 times for each dataset. Figure~\ref{fig:ML} plots the out-of-sample classification error as a function of the sample size, averaged across the ten instances. At each iteration, the out-of-sample classification error is calculated using all samples that have not been utilized for training. Table~\ref{tab:dataset} and Figure~\ref{fig:ML} also report the best fits of the form $\frac{\gamma}{n^r}+\beta$ (which corresponds to Assumption~\ref{ass:classification}). These results suggest a clear downward trend as the sample size increases, as well as a concave dependency indicating diminishing returns as the sample size becomes increasingly large. The fitted curves yield a good first-order approximation of the learning error, so that the assumption that the learning error decay in $\calO(1/n^r)$ characterizes fairly well the impact of additional data points on predictive performance---even in our online learning environment where data samples are not independent over time. These results complement our statistical learning results with an empirical justification of the general-purpose characterization of the machine learning error.

\begin{table}[h!]
\centering \small
\caption{Descriptive statistics of the 10 datasets and corresponding data-driven experimental setup.}
\label{tab:dataset}
\renewcommand{\arraystretch}{1.3}
\begin{tabular}{lllllll}
\toprule[1pt]
Dataset & Positive class & \# observations & \# features & Positive rate & Best fit\\ \midrule[0.5pt]
Bank & 521 & 4,521 & 16 & 11.5\% & $\frac{1}{n^{0.584}}+0.12$ \\
Default & 6,636 & 30,000 & 23 & 22.1\% & $\frac{0.69}{n^{0.703}}+0.18$ \\
Occupancy & 2,049 & 9,752 & 6 & 21.0\% & $\frac{0.19}{n^{0.544}}+0.01$\\
Online & 1,908 & 12,330 & 6 & 15.5\% & $\frac{0.71}{n^{0.635}}+0.08$ \\
\bottomrule[1pt] 
\end{tabular}
\end{table}

\section{Proof of statements from Section~\ref{subsec:robustness}}
\label{app:robustness}

\subsection{Proof of Theorem~\ref{thm:offline}}

We define Problem~\eqref{prob:QtmN} as follows, with $N_0=\gamma m^\beta+\calO\left(m^\beta\right)$:
\begin{align}
\min\quad & \sum_{t=1}^{T}A_t \label{prob:QtmN}\tag{$\calQ^*_T(m,N_0)$}\\
\st\quad&\P\left[\sum_{t=1}^TB_t\geq m\right] \geq 1-\delta \nonumber\\
\quad& B_t\sim \Bino\left(A_t, 1- \frac{\varepsilon\cdot p}{\left(N_0\sum_{s=1}^{t-1} A_s +1\right)^r}-\varepsilon\cdot(1-p)\right),\ \forall t\in\calT\nonumber\\
\quad& A_t,B_t\geq0,\ \forall t\in\calT\nonumber
\end{align}

We adjust the algorithm to account for offline data as follows:
\begin{equation}\label{eq:allocation_QtmN}
\calA(T,m,\gamma,\beta):\quad A^\dagger_t=r^{-\frac{1}{1-r}\left(t-T\cdot\alpha_T\cdot(1-r^t)\right)}\cdot \left(\frac{c_0}{\gamma}\right)^{\alpha_T\cdot(1-r^t)}\cdot \ell^{(1-\beta)\alpha_T\left(1-r^t\right) + \beta},
\end{equation}
where $\ell$ is the unique positive solution to the following equation:
\begin{align*}
\calL(T,m,\gamma,\beta):\quad&\sum_{t=1}^T A^\dagger_t=  c_0 \left(m + \varepsilon\cdot p\cdot\zeta_T\cdot \gamma^{-\alpha_T(1-r)}\cdot m^{(1-\beta)\cdot\alpha_T\cdot(1-r)+\beta(1-r)}+ \Delta(m)\right),
\end{align*}
and
$$\Delta(m) = \begin{cases}
\sqrt{\frac{c_0}{2}\log \frac{2}{\delta}}\sqrt{m} & \text{if }p \neq 1,\\
-\Phi^{-1}(\delta) \sqrt{2\varepsilon \zeta_T} m^{\frac{1}{2}\cdot \left((1-\beta)\cdot\alpha_T\cdot(1-r)+\beta(1-r)\right)}& \text{if }p = 1.\end{cases}$$

The only non-trivial change from Theorem~\ref{thm:main} comes from calculating the optimal solution of the its deterministic approximation, formulated as follows. The solution is given in Lemma~\ref{lem:QtmN_det}, which is analogous to Lemma~\ref{lem:main_det}.
\begin{align}
\min\ \left\{\ \sum_{t=1}^T A^{D}_t \ \ \bigg|\ \ \sum_{t=1}^TA^{D}_t \left(1 - \frac{\varepsilon\cdot p}{\left(N_0+ \sum_{s=1}^{t-1} A^{D}_s +1\right)^r} - \varepsilon\cdot(1-p)\ \right)=m\right\} \label{prob:QtmN_det} \tag{$\calQ^D_T(m,N_0)$}
\end{align}

\begin{lemma}\label{lem:QtmN_det}
Let $m^D(\varepsilon, T)$ be the optimal objective value of Problem~\eqref{prob:QtmN_det}, and $A^{D}_1,\cdots, A^{D}_T$ be its solution. For any $T\geq 2$, we have as $m \to \infty$: 
\begin{align*}
   & A^{D}_t = r^{-\frac{1}{1-r}\left(t-T\cdot\alpha_T\cdot(1-r^t)\right)}\cdot \left(\frac{c_0}{\gamma}\right)^{\alpha_T\cdot(1-r^t)}\cdot m^{(1-\beta)\alpha_T\left(1-r^t\right) + \beta}+o\left(m^{(1-\beta)\alpha_T\left(1-r^t\right) + \beta}\right),\ \forall t=1,\cdots,T\\
   &m^D(\varepsilon, T) = c_0 \left(m + \varepsilon\cdot p\cdot\zeta_T\cdot \gamma^{-\alpha_T(1-r)}\cdot m^{(1-\beta)\cdot\alpha_T\cdot(1-r)+\beta(1-r)}\right)+o\left(m^{(1-\beta)\cdot\alpha_T\cdot(1-r)+\beta(1-r)}\right).
\end{align*}
\end{lemma}

\subsubsection*{Proof of Lemma~\ref{lem:QtmN_det}}

Let $\lambda\in\R$ be the dual variable associated with the equality constraint. The KKT conditions yield the following vanishing gradient equation:
\begin{align*}
& 1+\lambda\left(1 - \frac{\varepsilon\cdot p}{\left(N_0+ \sum_{s=1}^{t-1} A^{D}_s +1\right)^r} - \varepsilon\cdot(1-p)\ \right)+\lambda\sum_{\tau=t+1}^T\frac{r\cdot\varepsilon \cdot p\cdot A^{D}_\tau}{\left(N_0 +\sum_{s=1}^{\tau-1} A^{D}_s +1\right)^{r+1}}=0,\ \forall t\in\calT
\end{align*}

Using the same logic as in the proof of Lemma~\ref{lem:main_det}, we can define $u_t = N_0 + \sum_{s=1}^t A^{D}_s + 1$ for $t=1,\cdots,T$, which gives rise to the same recursion:
\begin{equation*}
u_{t+1} + \left(\frac{1}{r}-1\right)u_t = \frac{u_t^{r+1}}{ru_{t-1}^r},\ \forall t=1,\cdots,T-1.
\end{equation*}
The difference comes from the boundary condition. 
Let us write $u_t = \Theta(g(t)m^{f(t)})$ toward our leading-order and leading-coefficient analyses. Since $u_0 = \gamma m^\beta + o(m^\beta)$, we now have $g(0)=\gamma$ and $f(0)=\beta$. Moreover, since the solution to Problem~\eqref{prob:QtmN_det} is upper bounded by that to Problem~\eqref{prob:main_det}, we still have $g(T)=c_0$ and $f(T)=1$.

We perform the same leading-order analysis as in Lemma~\ref{lem:main_det}, with $f(T)-f(0)=1-\beta$ instead of $f(T)-f(0)=1$. Therefore:
\begin{align*}
& f(T)-f(0)=1-\beta=\frac{K}{1-r}\left(1-r^T\right)\\
& f(t+1)-f(t)=Kr^t=(1-\beta)\cdot r^{t}\cdot\frac{1-r}{1-r^T}\\
& f(t)=\sum_{s=0}^{t-1}(f(s+1)-f(s))+\beta=(1-\beta)\cdot\frac{1-r}{1-r^T}\cdot\sum_{s=0}^{t-1}r^s+\beta=(1-\beta)\cdot\alpha_T(1-r^t)+\beta
\end{align*}

In conclusion, the leading-order analysis leads to:
$$u_t=\Theta\left(m^{(1-\beta)\alpha_T\left(1-r^t\right) + \beta}\right).$$

We now proceed with the leading-coefficient analysis. Using $h(t)=\log g(t)$, we obtain the same recursion as in Lemma~\ref{lem:main_det} except for the boundary conditions $h(T)=0$ and $h(0)=\log(\gamma)$:
\[h(t+1) = (r+1)h(t)-rh(t-1)-\log r\]

We obtain the following modified equations from the proof of Lemma~\ref{lem:main_det}:
\begin{align*}
& \sum_{s=0}^{T-1}(h(s+1)-h(s))=\underbrace{h(T)-h(0)}_{=-\log(c_0)-\log(\gamma)}=K'\frac{\left(1-r^T\right)}{1-r}-\frac{T\log r}{1-r}\\
& \implies h(t)=\frac{-\log r}{1-r}\left(t-T\alpha_T(1-r^t)\right)+\log(c_0/\gamma)\cdot\alpha_T(1-r^t).
\end{align*}

Bringing the two results together, we obtain:
\begin{equation*}
u_t=r^{-\frac{1}{1-r}\left(t-T\cdot\alpha_T\cdot(1-r^t)\right)}\cdot \left(\frac{c_0}{\gamma}\right)^{\alpha_T\cdot(1-r^t)}\cdot m^{(1-\beta)\alpha_T\left(1-r^t\right) + \beta}+o\left(m^{(1-\beta)\alpha_T\left(1-r^t\right) + \beta}\right)
\end{equation*}

For ease of the exposition, let us write $u_t=r^{-\frac{\overline{h}(t)}{1-r}}\left(\frac{c_0}{\gamma}\right)^{f(t)}m^{\overline{f}(t)} + o\left(m^{\overline{f}(t)}\right)$. As earlier, $\overline{h}(t)-r\overline{h}(t-1)=(1-r)t+r-\frac{T(1-r)}{1-r^T}$ and $f(t)-rf(t-1)=\alpha_T\cdot(1-r)$. Moreover:
\begin{align*}
\overline{f}(t)-r\overline{f}(t-1)&=(1-\beta)(f(t)-rf(t-1))+\beta(1-r)
=(1-\beta)\cdot\alpha_T\cdot(1-r)+\beta(1-r)
\end{align*}

We conclude as follows:
\begin{align*}
&\left(1-(1-p)\varepsilon)\right)\sum_{t=1}^T A^D_t\\
&= m + \sum_{t=1}^T \frac{A^D_t \cdot\varepsilon\cdot p}{\left(\sum_{s=1}^{t-1} A^D_s +1\right)^r}\\
&=  m + \sum_{t=1}^T \frac{\varepsilon\cdot p(u_t-u_{t-1})}{u^r_{t-1}}\\
&= m + \varepsilon\cdot p\cdot  \sum_{t=1}^T r^{-\frac{\overline{h}(t)-r\overline{h}(t-1)}{1-r}}\cdot \left(\frac{c_0}{\gamma}\right)^{f(t)-rf(t-1)}\cdot m^{\overline{f}(t)-r\overline{f}(t-1)}+o\left(m^{\overline{f}(t)-r\overline{f}(t-1)}\right)\\
&= m +\varepsilon\cdot p\cdot\left(\sum_{t=1}^T \frac{1}{r^t}\right)\cdot\left(\frac{1}{r}\right)^{\frac{r}{1-r}}\cdot r^{T\cdot\alpha_T}\cdot \left(\frac{c_0}{\gamma}\right)^{\alpha_T\cdot(1-r)}\cdot m^{(1-\beta)\cdot\alpha_T\cdot(1-r)+\beta(1-r)}+ o\left(m^{(1-\beta)\cdot\alpha_T\cdot(1-r)+\beta(1-r)}\right)\\
&= m+\varepsilon\cdot p\cdot\zeta_T\cdot \gamma^{-\alpha_T\cdot(1-r)}\cdot m^{(1-\beta)\cdot\alpha_T\cdot(1-r)+\beta(1-r)}+ o\left(m^{(1-\beta)\cdot\alpha_T\cdot(1-r)+\beta(1-r)}\right)
\end{align*}
This concludes the proof of Lemma~\ref{lem:QtmN_det}.
\hfill\Halmos

\subsection{Proof of Theorem~\ref{thm:warmstart}}
\label{app:warmstart}

We define Problem~\eqref{prob:WS} as follows:
    \begin{align}
        \min\quad & cA_0+\sum_{t=1}^T A_t \label{prob:WS}\tag{$\calP^*_{WS}$}\\
        \st\quad&\P\left[\sum_{t=1}^TB_t\geq m\right] \geq 1-\delta \nonumber\\
        \quad& B_t\sim \Bino\left(A_t, 1- \frac{\varepsilon\cdot p}{\left(A_0+\sum_{s=1}^{t-1} A_s +1\right)^r}-\varepsilon\cdot(1-p)\right),\ \forall t\in\calT\nonumber\\
        \quad& A_t,B_t\geq0,\ \forall t\in\calT\nonumber
    \end{align}

Again, the only non-trivial change from Theorem~\ref{thm:main} comes from calculating the optimal solution of the its deterministic approximation, formulated as follows. The solution is given in Lemma~\ref{lem:ws_det}.
{\small\begin{align}
            \min\ \left\{\ cA^D_0+ \sum_{t=1}^T A^{D}_t \ \ \bigg|\ \ \sum_{t=1}^TA^{D}_t \left(1- \frac{\varepsilon\cdot p}{\left(A^D_0+\sum_{s=1}^{t-1} A^D_s +1\right)^r}-\varepsilon\cdot(1-p)\ \right)=m\right\} \label{prob:ws_det} \tag{$\calP^{D}_{WS}$}
    \end{align}}

\begin{lemma}\label{lem:ws_det}
Let $m^D(\varepsilon, T)$ be the optimal objective value of Problem~\eqref{prob:ws_det}, and $A^{D}_1,\cdots, A^{D}_T$ be its solution. Define $\gamma_\star=\left(\frac{\alpha_T(1-r)c_0\varepsilon\cdot p \zeta_T}{c}\right)^{\frac{1}{\alpha_T(1-r)+1}}$. For any $T\geq 2$, we have as $m \to \infty$: 
\begin{align*}
   & A^{D}_t = r^{-\frac{1}{1-r}\left(t-T\cdot\alpha_T\cdot(1-r^t)\right)}\cdot \left(\frac{c_0}{\gamma_\star}\right)^{\alpha_T\cdot(1-r^t)}\cdot m^{\alpha_{T+1}(1-r^{t+1})}+ o\left(m^{\alpha_{T+1}(1-r^{t+1})}\right),\ \forall t=1,\cdots,T\\
   &m^D(\varepsilon, T) = c_0m+\left(c_0\cdot\varepsilon\cdot p\cdot\zeta_T\cdot \gamma_\star^{-\alpha_T\cdot(1-r)}+c\cdot\gamma_\star\right)m^{\alpha_{T+1}\cdot(1-r)}+ o\left(m^{\alpha_{T+1}\cdot(1-r)}\right).
\end{align*}
\end{lemma}
\subsubsection*{Proof of Lemma~\ref{lem:ws_det}}
Let us define $\gamma_\star>0$ and $\beta_\star\in[0,1)$ such that $A_0^D=\gamma_\star\cdot m^{\beta_\star}+o\left(m^{\beta_\star}\right)$. Unlike in Lemma~\ref{lem:QtmN_det}, $\gamma_\star$ and $\beta_\star$ are decision variables, reflecting the amount of offline data to be acquired. As in Lemma~\ref{lem:QtmN_det}, we can define $u_t = A^D_0 + \sum_{s=1}^t A^{D}_s + 1$ for $t=1,\cdots,T$, and derive:
\begin{equation*}
u_t=r^{-\frac{1}{1-r}\left(t-T\cdot\alpha_T\cdot(1-r^t)\right)}\cdot \left(\frac{c_0}{\gamma_\star}\right)^{\alpha_T\cdot(1-r^t)}\cdot m^{(1-\beta_\star)\alpha_T\left(1-r^t\right) + \beta_\star}+o\left(m^{(1-\beta_\star)\alpha_T\left(1-r^t\right) + \beta_\star}\right),
\end{equation*}
which achieves an objective of
$$c_0 \left(m + \varepsilon\cdot p\cdot\zeta_T\cdot \gamma_\star^{-\alpha_T(1-r)}\cdot m^{(1-\beta_\star)\cdot\alpha_T\cdot(1-r)+\beta_\star(1-r)}\right)+o\left(m^{(1-\beta_\star)\cdot\alpha_T\cdot(1-r)+\beta_\star(1-r)}\right).$$

By construction, $A^D_0 = \gamma_\star\cdot m^{\beta_\star}+ o(m^{\beta_\star})$ and therefore the full objective value is:
\[ c_0m + \Theta\left(m^{(1-\beta_\star)\cdot\alpha_T\cdot(1-r)+\beta_\star(1-r)}\right) + \Theta\left(m^{\beta_\star}\right)\]
As $m \to \infty$, the minimum is achieved when:
$$(1-\beta_\star)\cdot\alpha_T\cdot(1-r)+\beta_\star(1-r)=\beta_\star$$
Solving this equation gives $\beta_\star=\frac{\alpha_T(1-r)}{r+\alpha_T(1-r)}$. Recalling the definition of $\alpha_T=\frac{1}{1-r^T}$, we obtain: $\beta_\star=\frac{1-r}{1-r^{T+1}}=(1-r)\alpha_{T+1}$. We get $A^D_0 = \gamma_\star\cdot m^{\alpha_{T+1}\cdot(1-r)}$ and:
\begin{equation*}
u_t=r^{-\frac{1}{1-r}\left(t-T\cdot\alpha_T\cdot(1-r^t)\right)}\cdot \left(\frac{c_0}{\gamma_\star}\right)^{\alpha_T\cdot(1-r^t)}\cdot m^{\alpha_{T+1}(1-r^{t+1})}+o\left(m^{\alpha_{T+1}(1-r^{t+1})}\right)
\end{equation*}

Proceeding as in the proof of Lemma~\ref{lem:QtmN_det}, we derive:
\begin{align*}
\left(1-(1-p)\varepsilon)\right)\sum_{t=1}^T A^D_t
&= m+\varepsilon\cdot p\cdot\zeta_T\cdot \gamma_\star^{-\alpha_T\cdot(1-r)}\cdot m^{\alpha_{T+1}\cdot(1-r)}+ o\left(m^{\alpha_{T+1}\cdot(1-r)}\right)
\end{align*}

The optimal objective value is therefore:
$$c_0m+\left(c_0\cdot\varepsilon\cdot p\cdot\zeta_T\cdot \gamma_\star^{-\alpha_T\cdot(1-r)}+c\cdot\gamma_\star\right)m^{\alpha_{T+1}\cdot(1-r)}+ o\left(m^{\alpha_{T+1}\cdot(1-r)}\right)$$

Minimizing over $\gamma_\star$ yields the desired expression:
$$\gamma_\star=\left(\frac{\alpha_T(1-r)c_0\varepsilon\cdot p \zeta_T}{c}\right)^{\frac{1}{\alpha_T(1-r)+1}}$$
Then, we have:
\begin{align*}
c_0\cdot\varepsilon\cdot p\cdot\zeta_T\cdot \gamma_\star^{-\alpha_T\cdot(1-r)}+c\cdot\gamma_\star
&= c_0\cdot\varepsilon\cdot p\cdot\zeta_T\cdot \left(\frac{\alpha_T(1-r)c_0\varepsilon\cdot p \zeta_T}{c}\right)^{-\frac{\alpha_T\cdot(1-r)}{\alpha_T(1-r)+1}}+c\left(\frac{\alpha_T(1-r)c_0\varepsilon\cdot p \zeta_T}{c}\right)^{\frac{1}{\alpha_T(1-r)+1}}\\
&= \left(c_0\cdot\varepsilon\cdot p\cdot\zeta_T\cdot c^{\alpha_T(1-r)}\cdot\alpha_T(1-r)\right)^{\frac{1}{\alpha_T(1-r)+1}}\cdot \left(1+\frac{1}{\alpha_T(1-r)}\right)
\end{align*}
Therefore, the optimal objective value can be written as:
$$c_0m+\left(c_0\cdot\varepsilon\cdot p\cdot\zeta_T\cdot c^{\alpha_T(1-r)}\cdot\alpha_T(1-r)\right)^{\frac{1}{\alpha_T(1-r)+1}}\cdot \left(1+\frac{1}{\alpha_T(1-r)}\right)m^{\alpha_{T+1}\cdot(1-r)}+ o\left(m^{\alpha_{T+1}\cdot(1-r)}\right)$$
This concludes the proof of Lemma~\ref{prob:ws_det}, hence of the theorem.
\hfill\Halmos

\subsection{Adaptive policies, and proof of Theorem~\ref{thm:adaptive}}
\label{app:adaptive}

\subsubsection*{Dynamic programming formulation.}

In this appendix, we first formalize the adaptive optimization problem via dynamic programming. The state variable includes the number of tentative facility openings at the beginning of each period $t$, that is, $N_{t-1}$, and the number successful facility openings, which we denote by $\overline{B}_{t-1}$. Mathematically, these variables are the cumulative sum of the variables in our model, that is $N_{t-1}=\sum_{s=1}^{t-1}A_s$ and $\overline{B}_{t-1}=\sum_{s=1}^{t-1}B_s$. The decision variable characterizes the number of tentative facility openings in the $t^{\text{th}}$ period, that is $A_t$. The transition function specifies that $N_t=N_{t-1}+A_t$ and $\overline{B}_t=\overline{B}_{t-1}+W_t$, where $W_t$ follows a binomial distribution with $A_t$ trials and success probability $1- \frac{\varepsilon\cdot p}{\left(\sum_{s=1}^{t-1} A_s +1\right)^r}-\varepsilon\cdot(1-p)$ (per Assumption~\ref{ass:classification}). The Bellman equation is then given as follows where $J_t(N_{t-1},\overline{B}_{t-1})$ denotes the cost-to-go function. Note that the terminal condition is applied at time $T$ rather than at time $T+1$ to reflect the chance constraint in the final-period decision.
\begin{align*}
    J_t(N_{t-1},\overline{B}_{t-1})=\min_{A_t\geq0}\ \ &\left\{A_t+\E J_{t+1}\left(N_{t-1}+A_t,\overline{B}_{t-1}+\text{Binom}\left(A_t,1- \frac{\varepsilon\cdot p}{\left(N_{t-1} +1\right)^r}-\varepsilon\cdot(1-p)\right)\right)\right\}\\
    J_T(N_{T-1},\overline{B}_{T-1})=\min_{A_T\geq0}\ \ & N_{T-1}+A_T\\\st\ \ &\P\left[\Bino\left(A_T, 1- \frac{\varepsilon\cdot p}{\left(N_{T-1} +1\right)^r}-\varepsilon\cdot(1-p)\right)\geq m-\overline{B}_{T-1}\right]\geq1-\delta
\end{align*}

Unfortunately, the dynamic program would be very hard to solve analytically, in particular due to the interactions between the dynamics over the first $T-1$ periods and the final-period decision based on the chance constraint. We therefore resort to characterizing the solution of an adaptive re-solving algorithm, which applies the static problem iteratively over the planning horizon.

\subsubsection*{Details on the adaptive algorithm.}
The adaptive algorithm is specified in Algorithm~\ref{alg:adaptive}. At each time period $t=1,\cdots,T-1$, it solves Problem~$\left(\calQ^*_{T-t+1}\left(m-\sum_{s=1}^{t-1}B^{\textsc{ad}}_s,\sum_{s=1}^{t-1} A^{\textsc{ad}}_s\right)\right)$. It then solves Problem~$(\calS^*\left(m-\sum_{s=1}^{T-1}B^{\textsc{ad}}_s,\sum_{s=1}^{T-1} A^{\textsc{ad}}_s\right)$ in the last period, which is defined as follows:
\begin{align}
\min\quad &\left\{A_T:\P\left[\Bino\left(A_T, 1- \frac{\varepsilon\cdot p}{\left(N_0 +1\right)^r}-\varepsilon\cdot(1-p) \right)\geq m\right] \geq 1-\delta,\ A_T\geq0\right\} \label{prob:lastperiod}\tag{$\calS^*(m,N_0)$}
\end{align}

\begin{algorithm} [h!]
\caption{Adaptive algorithm to solve Problem $(\calP^*)$.}\small
\label{alg:adaptive}
\begin{algorithmic}
\item Repeat, for $t=1,\cdots,T-1$:
\begin{itemize}
\item[] \textbf{1. Optimize.} Define $\beta_t=\alpha_T\left(1-r^{t-1}\right)$ and $\gamma_t$ such that $\sum_{s=1}^{t-1} A^{\textsc{ad}}_s=\gamma_t m^{\beta_t}+o\left(m^{\beta_t}\right)$. Solve Problem $\left(\calQ^*_{T-t+1}\left(m-\sum_{s=1}^{t-1}B^{\textsc{ad}}_s,\sum_{s=1}^{t-1} A^{\textsc{ad}}_s\right)\right)$. Retrieve solution $A^\dagger_1,\cdots,A^\dagger_{T-t+1}$.
\item[] \textbf{2. Commit immediate decision.} Implement the first-period decision: $A^{\textsc{ad}}_t=A^\dagger_1$.
\item[] \textbf{3. Observe realizations.} Sample $B^{\textsc{ad}}_t\sim \Bino\left(A^{\textsc{ad}}_t, 1- \frac{\varepsilon\cdot p}{\left(\sum_{s=1}^{t-1} A^{\textsc{ad}}_s +1\right)^r}-\varepsilon\cdot(1-p)\right)$.
\end{itemize}
\item $A^{\textsc{ad}}_T\gets\left(\calS^*\left(m-\sum_{s=1}^{T-1}B^{\textsc{ad}}_s,\sum_{s=1}^{T-1} A^{\textsc{ad}}_s\right)\right)$. Sample $B^{\textsc{ad}}_T\sim \Bino\left(A^{\textsc{ad}}_T, 1- \frac{\varepsilon\cdot p}{\left(\sum_{s=1}^{T-1} A^{\textsc{ad}}_s +1\right)^r}-\varepsilon\cdot(1-p)\right)$.
\end{algorithmic}
\end{algorithm}

At each iteration, the algorithm solves a variant of the problem with offline data, albeit with an increasing offline data sample over iterations reflected in $\sum_{s=1}^{t-1} A^{\textsc{ad}}_s=\gamma_t m^{\beta_t}+o\left(m^{\beta_t}\right)$ and $\beta_t=\alpha_T\left(1-r^{t-1}\right)$. Therefore, at each iteration, we can invoke Theorem~\ref{thm:offline} to elicit the $A^\dagger_1,\cdots,A^\dagger_{T-t+1}$ and the corresponding regret. Specifically, at each iteration, Problem~$\left(\calQ^*_{T-t+1}\left(m-\sum_{s=1}^{t-1}B_s,\sum_{s=1}^{t-1} A_s\right)\right)$ is solved via Equations $\calA\left(T-t+1,m-\sum_{s=1}^{t-1}B^{\textsc{ad}}_s,\gamma_t,\beta_t\right)$ and $\calL\left(T-t+1,m-\sum_{s=1}^{t-1}B^{\textsc{ad}}_s,\gamma_t,\beta_t\right)$, exploiting the fact that $\sum_{s=1}^{t-1} A_s=\Theta(\gamma_t m^{\beta_t})$. Note that $\gamma_t$ and $\beta_t$ change across iterations due to the increasing data sample, but the procedure to solve the problem remains identical across iterations, and also remains identical to the solution procedure used to solve the problem with offline data in Theorem~\ref{thm:offline}.

\subsubsection*{Proof of Theorem~\ref{thm:adaptive}.}

We prove by induction over $t=1,\cdots,T-1$ that $$A^{\textsc{ad}}_t=\Theta\left(m^{\alpha_T\left(1-r^t\right)}\right).$$

The result for $t=1$ follows directly from Theorem~\ref{thm:main}. Let us assume that it holds for $t-1\in\{1,\cdots,T-2\}$ and prove it for $t$. In period $t$, the decision-maker solves $\left(\calQ^*_{T-t+1}\left(m-\sum_{s=1}^{t-1}B^{\textsc{ad}}_s,\sum_{s=1}^{t-1} A^{\textsc{ad}}_s\right)\right)$, where $\sum_{s=1}^{t-1} A^{\textsc{ad}}_s=\Theta\left(m^{\alpha_T\left(1-r^{t-1}\right)}\right)$ per the induction hypothesis. Let us denote by $\beta_t=\alpha_T\left(1-r^{t-1}\right)$. Let us momentarily assume that  $\delta=o\left(m^{(1-\beta_t)\cdot\alpha_{T-t+1}\cdot(1-r)+\beta_t(1-r)-1}\right)$. Then, per Theorem~\ref{thm:offline}, we know that the solution of that problem, denoted by $A^\dagger_\tau$, satisfies 
$A^\dagger_\tau=\theta\left(m^{(1-\beta_t)\alpha_{T-t+1}\left(1-r^\tau\right)+\beta_t}\right)$ for $\tau=1,\cdots,T-t+1$. By construction, the adaptive policy follows the first-period solution in the problem solved at time $t$, so that $A^{\textsc{ad}}_t=A^\dagger_1$. Therefore: $A^{\textsc{ad}}_t=\Theta\left(m^{(1-\beta_t)\alpha_{T-t+1}(1-r)+\beta_t}\right)$. We conclude the induction as follows:
\begin{align*}
(1-\beta_t)\alpha_{T-t+1}(1-r)+\beta_t
&=\left(1-\alpha_T(1-r^{t-1})\right)\cdot\frac{1}{1-r^{T-t+1}}(1-r)+\alpha_T(1-r^{t-1})\\
&=\frac{1}{1-r^T}\left(\frac{(r^{t-1}-r^T)(1-r)}{1-r^{T-t+1}}+1-r^{t-1}\right)\\
&=\frac{1}{1-r^T}\left(r^{t-1}(1-r)+1-r^{t-1}\right)\\
&=\alpha_T(1-r^t)
\end{align*}

Similarly, the leading-order term in the regret expression does not change over the iterations of the adaptive algorithm. Indeed, per Theorem~\ref{thm:offline}, the algorithm achieves a regret of $\Theta\left(m^{(1-\beta_t)\cdot\alpha_{T-t+1}\cdot(1-r)+\beta_t(1-r)}\right)$ at each iteration, and
\begin{align*}
(1-\beta_t)\alpha_{T-t+1}(1-r)+\beta_t(1-r)
&=\frac{1}{1-r^T}\left(r^{t-1}(1-r)+(1-r^{t-1})(1-r)\right)
=\alpha_T(1-r)
\end{align*}
Note, also, that this property implies that $\delta=o\left(m^{(1-\beta_t)\cdot\alpha_{T-t+1}\cdot(1-r)+\beta_t(1-r)-1}\right)$, since by assumption $\delta=o\left(m^{\alpha_T(1-r)-1}\right)$, which ensures that the condition of Lemma~\ref{lem:QtmN_det} is satisfied. This concludes the proof that the algorithm maintains a regret of $\Theta\left(m^{\alpha_T\cdot(1-r)}\right)$ at each iteration.

\subsection{Semi-adaptive procedure with final-period adjustment}

Recall that the semi-adaptive procedure implements $A^\dagger_1,\cdots,A^\dagger_{T-1}$ from Algorithm~\ref{ALG}, and then solves Problem $\left(\calS^*\left(m-\sum_{s=1}^{T-1}B\dagger_s,\sum_{s=1}^{T-1} A\dagger_s\right)\right)$ in period $T$. This is detailed in Algorithm~\ref{alg:adaptive_simple}.

\begin{algorithm} [h!]
\caption{Semi-adaptive algorithm to solve Problem $(\calP^*)$.}\small
\label{alg:adaptive_simple}
\begin{algorithmic}
\item Repeat, for $t=1,\cdots,T-1$:
\begin{itemize}
    \item[] \textbf{1. Optimize.} Solve Problem $(\calP^*)$ using Algorithm 1 from the paper. Retrieve solution $A^\dagger_1,\cdots,A^\dagger_{T}$.
    \item[] \textbf{2. Commit $T-1$ decisions.} Implement decisions $A^\dagger_1,\cdots,A^\dagger_{T-1}$.
    \item[] \textbf{3. Observe realizations.} Sample $B_t\sim \Bino\left(A^\dagger_t, 1- \frac{\varepsilon\cdot p}{\left(\sum_{s=1}^{t-1} A_s +1\right)^r}-\varepsilon\cdot(1-p)\right)$, $t=1,\cdots,T-1$.
\end{itemize}
\item $A^\dagger_T\gets\left(\calS^*\left(m-\sum_{s=1}^{T-1}B^\dagger_s,\sum_{s=1}^{T-1} A^\dagger_s\right)\right)$. Sample $B^\dagger_T\sim \Bino\left(A^\dagger_T, 1- \frac{\varepsilon\cdot p}{\left(\sum_{s=1}^{T-1} A^\dagger_s +1\right)^r}-\varepsilon\cdot(1-p)\right)$.
\end{algorithmic}
\end{algorithm}

\section{Proof of statements from Section~\ref{sec:network}}
\label{app:network}

We will now focus on the case where the conditions does not hold. We will first prove the relationship between the solution of Algorithm~\eqref{prob:network_ALG} and the optimal solution of Problem~\eqref{prob:network} for any value of $p$ (Part 1 below). We then prove the relationship between the Problem~\eqref{prob:network} and the fully-learned benchmark when $p=1$ (Part 2).
\begin{itemize}
\item[--]  Part 1: Algorithm~\ref{ALG_network} yields a feasible solution to Problem~\eqref{prob:network} such that
$$m^\dagger(\varepsilon, p,T)\leq m^*(\varepsilon, p,T)+\begin{cases}
o\left(m^{\alpha_T \cdot (1-r)}\right)& \text{if }p=1 \text{ or } (p \neq 1 \text{ and } \alpha_T\cdot(1-r)>1/2)\\
\calO(m^{1/2}) & \text{if }p \neq 1\text{ and } \alpha_T\cdot(1-r)\leq 1/2        
\end{cases}$$
\item[--] Part 2: regret of the optimal stochastic solution against the fully-learned solution when $p=1$:
\[m^*(\varepsilon, p,T) = \widehat{m}(\varepsilon,p)+\Theta\left(m^{g(r)}\right)\]
\end{itemize}

The restriction to $p=1$ for Part 2 stems from the fact that the characterization of the fully-learned solution is no longer tractable when $p<1$, which prevents us from bounding the solutions to the stochastic problem and the fully-learned problem. When $p=1$, the "fully-learned" information benchmark is formulated as a (deterministic) discrete optimization problem (Problem~\eqref{prob:network_perfect}).

Throughout this proof, we denote by $y^*_{it}$, $y^\dagger_{it}$, and $\widehat{y}_{it}$ the solutions of the stochastic problem~\eqref{prob:network}, our algorithm~\eqref{prob:network_ALG}, and the fully-learned problem~\eqref{prob:network_perfect}, respectively. We denote by $s^*_{jT}$, $s^\dagger_{jT}$, and $\widehat{s}_{jT}$ the corresponding coverage variables. The optimal values are given as
\begin{align*}
m^*(\varepsilon, p,T)=\sum_{i=1}^n \sum_{t=1}^T y^*_{it}\qquad
m^\dagger(\varepsilon, p,T)=\sum_{i=1}^n \sum_{t=1}^T y^\dagger_{it}\qquad
\widehat{m}(\varepsilon,p)=\sum_{i=1}^n \sum_{t=1}^T \widehat{y}_{it}
\end{align*}
We define $\pi_{j}(\by)$ as the probability that customer $j=1,\sum_{j=1}^q$ is covered by a facility at the end of the horizon under the decision vector $\by=\{y_{it}\mid i=1,\cdots,n,\ t=1,\cdots,T\}$:
\begin{equation*}
\pi_{j}(\by)=\left(1-\prod_{\tau=1}^T \left(\frac{\varepsilon\cdot p}{\left(\sum_{s=1}^{\tau-1} \sum_{i=1}^n y_{is} +1\right)^r}+\varepsilon\cdot(1-p)\right)^{\sum_{i \in \calF_j} y_{i\tau}}\right)
\end{equation*}

We use $\calN(\by)$ to denote the set of facilities that are tentatively opened during the planning horizon, and $\calM(\by)$ to denote the set of customers with at least one adjacent facility that has been tentatively opened under the decision vector $\by=\{y_{it}\mid i=1,\cdots,n,\ t=1,\cdots,T\}$. Formally:
\begin{align*}
\calN(\by)&=\{i\in\{1,\cdots,n\} \mid \exists t\in\{1,\cdots,T\}:y_{it}=1\}\\
\calM(\by)&=\{j\in\{1,\cdots,q\} \mid \pi_{j}(\by)>0\}=\left\{j\in\{1,\cdots,q\}\mid\calF_j\cap\calN(\by)\neq\emptyset\right\}
\end{align*}

We assume that $\calF^\textsc{ctl}(d_\star)=\emptyset$ without loss of generality. Consider the graph $\calG'$ where all central nodes $\calF^\textsc{ctl}(d_\star)\cup \calC^\textsc{ctl}(d_\star)$ and all connecting edges are removed. Let the optimal values of the stochastic problem, our algorithm, and the fully-learned problem in $\calG'$ as $m_0^*(\varepsilon, T)$, $m_0^\dagger(\varepsilon, T)$, and $\widehat{m}_0(T)$ respectively. We can construct solutions to the original problems (with $\calG$) by adding all central facilities, so that
\begin{align*}
& m^*(\varepsilon, p,T)\leq m_0^*(\varepsilon, T)\leq m^*(\varepsilon, p,T)+|\calF^\textsc{ctl}(d_\star)|\\
& m^\dagger(\varepsilon, p,T)\leq m_0^\dagger(\varepsilon, T)\leq m^\dagger(\varepsilon, p,T)+|\calF^\textsc{ctl}(d_\star)|\\
& \widehat{m}(\varepsilon, T)\leq \widehat{m}_0(\varepsilon, T)\leq \widehat{m}(\varepsilon, T)+|\calF^\textsc{ctl}(d_\star)|
\end{align*}

Once we prove the theorem for any graph such that $\calF^\textsc{ctl}(d_\star)=\emptyset$, we get $m_0^*(\varepsilon, T)=\widehat{m}_0(T)+\Theta\left(m^{g(r)}\right)$ and $m_0^\dagger(\varepsilon, T)=m_0^*(\varepsilon, T)+o\left(m^{g(r)}\right)$ or $m_0^\dagger(\varepsilon, T)=m_0^*(\varepsilon, T)+\Theta\left(m^{1/2}\right)$. Since the minimum order of terms is strictly larger than $k$, this implies per Assumption~\ref{ass:degree} that $m^*(\varepsilon, p,T)=\widehat{m}(\varepsilon,p)+\Theta\left(m^{g(r)}\right)$ and $m^\dagger(\varepsilon, p,T)=m^*(\varepsilon, p,T)+o\left(m^{g(r)}\right)$ or $m^\dagger(\varepsilon, p,T)=m^*(\varepsilon, p,T)+\Theta\left(m^{1/2}\right)$.

We make use of several lemmas. The logic of the proof is summarized in Figure~\ref{fig:proof_network}. To prevent circular reasoning, we prove the lemmas in ascending order to derive a proof of Theorem~\ref{thm:network}.

\begin{figure}[h!]
\begin{center}
\begin{tikzpicture}[node distance=3cm, scale=0.3]
% Define styles for boxes
\tikzstyle{box} = [rectangle, draw=black, thick, minimum size=1cm, text centered]

% Nodes
\node (th) [box, minimum width=3cm] {Theorem~\ref{thm:network}};
\node (p0) [below of=th] {};
\node (p1) [box, left of=p0] {Part 1};
\node (p2) [box, right of=p0, fill=black!10] {Part 2};	
\node (l10) [box, below of=p1] {Lemma~\ref{lem:network_deter_stochstic}};
\node (l8) [box, below of=p2, fill=black!10] {Lemma~\ref{lem:network_deter_lower}};
\node (l5) [box, left of=l10] {Lemma~\ref{lem:hoeffding_gen_bound}};
\node (l6) [box, below of=l8, fill=black!10] {Lemma~\ref{lem:bound_loss_gen}};
\node (l9) [box, below of=l10] {Lemma~\ref{lem:prob_bound}};
\node (l4) [box, left of=l9] {Lemma~\ref{lem:sol_bound}};
\node (l7) [box, left of=l6] {Lemma~\ref{lem:gen_deter_sol_lower}};
\draw[thick,->] (p1) -- (th);
\draw[thick,->] (p2) -- (th);
\draw[thick,->] (l5) -- (l10);
\draw[thick,->] (l9) -- (l10);
\draw[thick,->] (l4) -- (l10);
\draw[thick,->] (l6) -- (l8);
\draw[thick,->] (l7) -- (l8);
\draw[thick,->] (l7) -- (l10);
\draw[thick,->] (l10) -- (p2);
\draw[thick,->] (l4.west) -- +(-2,0) |- (p1);
\draw[thick,->] (l4) -- (l5);
\draw[thick,->] (l5) -- (p1);
\draw[thick,->] (l8) -- (p2);
\draw[thick,->] (l10) -- (p1);
\end{tikzpicture}
\end{center}
\caption{Logic of the proof of Theorem~\ref{thm:network}. Arrows indicate dependencies between results. Grayed boxes indicate results obtained for $p=1$, white boxes indicate results obtained for any value of $p\leq1$.}
\label{fig:proof_network}
\end{figure}
\vspace{12pt}

\subsection{Statement and proof of Lemma~\ref{lem:sol_bound}}

Lemma~\ref{lem:sol_bound} shows that the number of facilities that need to be opened is on the order of $m$, both under the algorithm and in the optimal solution to the stochastic problem. The proof uses the degree assumption to identify sufficient feasibility conditions based on the \textit{number} of facilities that are tentatively opened, which enables to leverage the technical results from Section~\ref{sec:base}.

\begin{lemma}\label{lem:sol_bound}
For sufficiently large $m$, we have:
\begin{align*}
    \frac{m}{d_\star} \leq |\calN(\by^\dagger)|&\leq  d_\star c_0\left(m + \varepsilon\cdot p \cdot \zeta_T\cdot m^{\alpha_T\cdot(1-r)} +\Delta(m)+3\Delta_G(m)\right)\\
    \frac{m(1-\delta)}{d_\star} \leq \E[|\calN(\by^*)|]&\leq   d_\star c_0\left(m + \varepsilon\cdot p \cdot \zeta_T\cdot m^{\alpha_T\cdot(1-r)} + \Delta(m)\right)      
\end{align*}
\end{lemma}

\proof{Proof:} Let us prove the lower bounds and upper bounds separately.

\paragraph{Lower bounds.}

The lower bounds stem from the degree assumption: each facility covers up to $d_\star$ customers and $m$ customers need to be covered, so at least $m/d_\star$ facilities are required. Formally, $\pi_{j}(\by)=0$ if and only if $y_{it}=0$ for all $t=1,\cdots,T$ and for all $i\in\calF_j$:
$$|\calM(\by^\dagger)|\geq\sum_{j\in\calM(\by^\dagger)}\pi_{j}(\by^\dagger)=\sum_{j=1}^q \pi_{j}(\by^\dagger) \geq m+\Delta_G(m)\geq m$$
Moreover:
\begin{align*}
|\calM(\by^\dagger)|
&=\sum_{j=1}^q\bone(\calF_j\cap\calN(\by^\dagger)\neq\emptyset)\\
&\leq \sum_{j=1}^q\sum_{i=1}^n\bone(i\in\calF_j\cap\calN(\by^\dagger))\\
&=\sum_{i=1}^n\left(\bone\left(i\in\calN(\by^\dagger)\right)\times\sum_{j=1}^q\bone(i\in\calF_j)\right)\\
&\leq d_\star|\calN(\by^\dagger)|,
\end{align*}
due to the assumption that the degree is bounded by $d_\star$. We obtain that
$m\leq d_\star|\calN(\by)|$.

Similarly, for the stochastic solution, we know that with probability $1-\delta$ the solution satisfies $\sum_{j=1}^q s^*_{jT} \geq m$. Conditioned on this event, we use the result above to write:
\[\E\left[|\calN(\by^*)| \mid \sum_{j=1}^q s^*_{jT} \geq m\right] \geq \frac{m}{d_\star}\]
We conclude:
\[\E[|\calN(\by^*)|] \geq \frac{m(1-\delta)}{d_\star}\]

\paragraph{Upper bounds.}

Let us start with Problem~\eqref{prob:network}. Lemma~\ref{lem:main_feasible} provides an algorithm that tentatively opens $d_\star c_0\left(m + \varepsilon\cdot p \cdot\zeta_T\cdot m^{\alpha_T\cdot(1-r)} + \Delta(m)\right)$ facilities over $T$ periods, so that at least $d_\star m$ facilities are actually opened with probability at least $1-\delta$. Per our degree assumption, each customer is connected to at most $d_\star$ facilities. Using the same argument as above, any set of $d_\star m$ facilities is connected to at least $m$ customers. This means that the solution satisfies the chance constraint, and defines a feasible solution to problem~\eqref{prob:network}. We have derived a solution that satisfies:
\begin{align*}
&\sum_{t=1}^T \sum_{i=1}^n y_{it} =d_\star c_0 \left(m + \varepsilon\cdot p \cdot\zeta_T\cdot m^{\alpha_T\cdot(1-r)}+ \Delta(m)\right) \\
& \P\left[\sum_{t=1}^T \sum_{i=1}^n z_{it} \geq d_\star m\right] \geq 1-\delta\\
& \P\left[\sum_{j=1}^q s_{jT} \geq m\right] \geq 1-\delta
\end{align*}

Turning to Algorithm~\ref{ALG_network}, define $q=m+\Delta_G(m)$. Using Lemma~\ref{lem:main_bound}, there exists an algorithm such that tentatively opening the following number of facilities
$$\sum_{i=1}^n \sum_{t=1}^T y_{it}=d_\star c_0 \left(q + \varepsilon\cdot p \cdot\zeta_T\cdot q^{\alpha_T\cdot(1-r)} + \Delta(q)\right)$$
would result in the number of opened facilities in expectation such that:
\[\E\left[\sum_{i=1}^n \sum_{t=1}^T z_{it}\right]  \geq d_\star q,\]
for $m$ sufficiently large. Again, since each customer is connected to at most $d_\star$ facilities, any set of $d_\star q$ facilities is connected to at least $q$ customers. Thus, the expected number of connected customers must satisfy:
\[\E\left[\sum_{j=1}^q s_{jT} \geq m\right]\geq\frac{1}{d_\star}\E\left[\sum_{i=1}^n \sum_{t=1}^T z_{it}\right]\geq q\]\

Therefore, opening $d_\star c_0 \left(q + \varepsilon\cdot p \cdot\zeta_T\cdot q^{\alpha_T\cdot(1-r)} + \Delta(q) \right)$ defines a feasible solution for Problem~\eqref{prob:network_ALG}, hence an upper bound to $|\calN(\by^\dagger)|$. Note that the function within the brackets is concave in $q$ and its derivative is less than 3 for $q$ sufficiently large. Therefore, we obtain:
\begin{align*}
|\calN(\by^\dagger)|&\leq d_\star c_0 \left(q + \varepsilon\cdot p \cdot \zeta_T\cdot q^{\alpha_T\cdot(1-r)} + \Delta(q)\right)\\
&\leq d_\star c_0 \left(m + \varepsilon\cdot p \cdot\zeta_T\cdot m^{\alpha_T\cdot(1-r)} + \Delta(m)+3(q-m)\right)\\
&= d_\star c_0 \left(m + \varepsilon\cdot p \cdot\zeta_T\cdot m^{\alpha_T\cdot(1-r)}+ \Delta(m)+3\Delta_G(m)\right)
\end{align*}
This completes the proof of Lemma~\ref{lem:sol_bound}.
\hfill\Halmos

\subsubsection*{Statement and proof of Lemma~\ref{lem:hoeffding_gen_bound}.}

\begin{lemma}\label{lem:hoeffding_gen_bound}
Consider $\by=\{y_{it}\mid i=1,\cdots,n,\ t=1,\cdots,T\}$ such that $|\calN(\by)|\leq Km$ for some fixed constant $K$. Define the event:
\begin{equation}
\calE(\by)=\left\{\left|\sum_{j=1}^q s_{jT} - \pi_{j}(\by)\right|\leq \Delta_G(m)\right\}	\label{eq:Ek_gen}
\end{equation}
Then, $\P(\calE(\by^\dagger))\geq 1-\delta$ and $\P(\calE(\by^*))\geq 1-\delta$.
\end{lemma}

Lemma~\ref{lem:hoeffding_gen_bound} proves that the ``good'' events $\calE(\by^\dagger)$ and $\calE(\by^*)$ occur with high probability under the stochastic and deterministic solutions. This shows that the total number of customers that are covered is likely to be close to its expected value within a factor of $\Delta_G(m)$.

The key technical difficulty is that whether a customer is covered is not independent from whether another customer is covered, due to the underlying graph structure. This prevents the application of standard concentration inequalities. Instead, we use the concentration inequality in dependency graphs from \cite{janisch2024berry}, based on the fractional chromatic number.

\begin{definition}\label{def:chromatic}
Let $\calG=(\calN,\calA)$ define an undirected graph. A coloring of $\calG$ labels each vertex by one color each such that no two adjacent vertices have the same color. The chromatic number $\chi(\calG)$ is defined as the smallest number of colors defining a coloring of $\calG$.
\end{definition}

\begin{definition}\label{def:Fchromatic}
Let $\calG=(\calN,\calA)$ define an undirected graph. A $b$-fold coloring of $\calG$ labels each vertex with up to $b$ colors, such that no two adjacent vertices have any color in common. Its $b$-fold chromatic number $\chi_b(\calG)$ is defined as the smallest number of colors defining a $b$-fold coloring of $\calG$. The fractional chromatic number, denoted by $\chi^*(\calG)$, is given by: $\chi^*(\calG)=\lim\limits_{b\to\infty}\frac{\chi_b(\calG)}{b}$
\end{definition}

The following result extends the Berry-Esseen in the case of dependency graphs.
\begin{theorem}[\cite{janisch2024berry}]
Let $X_1,\cdots, X_n$ be random variables in $[0,c_i]$. Let $\calG$ denote the dependency graph of $X_1,\cdots, X_n$ and $\chi^*(\calD)$ denote its fractional chromatic number. Then we have:
\begin{equation*}
    \P\left(\left|\sum_{i=1}^n X_i - \E\left[\sum_{i=1}^n X_i\right]\right|\geq s \sqrt{\Var\left(\sum_{i=1}^n X_i\right)}\right)=2\Phi(-s) +\calO\left(\frac{\max_{i} c_i \cdot \chi^*(\calG)^2 \sum_{i=1}^n \E\left[(X_i-\E[X_i])^3\right]}{\left(\Var\left(\sum_{i=1}^n X_i\right)\right)^{3/2}}\right)
\end{equation*}
\end{theorem}

Let us prove Lemma~\ref{lem:hoeffding_gen_bound} with $\by = \by^\dagger$. The arguments proceed in the same way for $\by = \by^*$. We define $\calD=(\calN,\calA)$ as the dependency graph of the random variables $s_{jT}$, conditioned on the facilities tentatively opened under decision $\by$. Each node represents a customer. An edge connects two customers $j$ and $k$ if and only if $s_{jT}$ and $s_{kT}$ are not independent, i.e., if and only if a facility that can cover both has been tentatively opened: $(j,k)\in\calA\iff\calF_j\cap\calF_k\cap\calN(\by)\neq\emptyset$

By assumption, the degree of any node in the customer-facility graph is less than or equal to $d_\star$. Therefore, any customer can share an adjacent facility with up to $d_\star(d_\star-1)$ other customers.
By Brooks' Theorem \citep{brooks1941colouring}, the chromatic number of $\calD$ satisfies $\chi(\calD)\leq (d_\star)^2-d_\star+1\leq (d_\star)^2$. This implies that its fractional chromatic number satisfies $\chi^*(\calD)\leq \chi(\calD)\leq (d_\star)^2$.

We use this result to characterize $\left|\sum_{j=1}^q s_{jT} - \pi_{j}(\by^\dagger)\right|$. By definition, each variable $s_{jT}$ is bounded by 1. We also need to bound $\Var\left(\sum_{j=1}^q s^\dagger_{jT}\right)$. By definition of $\chi^*(\calD)$, $s^\dagger_{jT}$ is correlated to at most $\chi^*(\calD)$ other variables $s^\dagger_{iT}$. Using the inequality $\Var(\sum_{i=1}^n X_i)\leq n \sum_{i=1}^n \Var(X_i)$, we obtain
\[\Var\left(\sum_{j=1}^q s^\dagger_{jT}\right) \leq  \chi^*(\calD)\ \sum_{j=1}^q \Var(s^\dagger_{jT})\]

Now we separate the cases where $p=1$ and $p\neq 1$.

\underline{Case 1: $p\neq 1$.}
\begin{align*}
&\Var\left(\sum_{j=1}^q s^\dagger_{jT}\right)
\\\leq & \chi^*(\calD)\ \sum_{j=1}^q \Var(s^\dagger_{jT})
\\\leq & d_\star^2 \sum_{j \in \calM(\by^\dagger)} \Var(s^\dagger_{jT})
\\\leq & d_\star^2\sum_{j \in \calM(\by^\dagger)}  \prod_{\tau=1}^T \left(\frac{\varepsilon\cdot p}{\left(\sum_{s=1}^{\tau-1} \sum_{i=1}^n y^\dagger_{is} + 1\right)^r}+\varepsilon\cdot(1-p)\right)^{\sum_{i \in \calF_j} y^\dagger_{i\tau}}
\\= & d_\star^2\sum_{j \in \calM(\by^\dagger)}  \prod_{\tau:\sum_{i \in \calF_j} y^\dagger_{i\tau}>0 }  \prod_{i:i \in \calF_j, y^\dagger_{i\tau}>0 }  \left(\frac{\varepsilon\cdot p}{\left(\sum_{s=1}^{\tau-1} \sum_{i=1}^n y^\dagger_{is} + 1\right)^r}+\varepsilon\cdot(1-p)\right)^{ y^\dagger_{i\tau}}    
\\\leq & d_\star^2\sum_{j \in \calM(\by^\dagger)}  \prod_{\tau:\sum_{i \in \calF_j} y^\dagger_{i\tau}>0 } \sum_{i: i \in \calF_j,  y^\dagger_{i\tau}>0} \left(\frac{\varepsilon\cdot p}{\left(\sum_{s=1}^{\tau-1} \sum_{i=1}^n y^\dagger_{is} + 1\right)^r}+\varepsilon\cdot(1-p)\right)\frac{y^\dagger_{i\tau}}{|i:  y^\dagger_{i\tau}>0|}
\\\leq & d_\star^2\sum_{j \in \calM(\by^\dagger)}  \left(\prod_{\tau:\sum_{i \in \calF_j} y^\dagger_{i\tau}>0 } \left(\sum_{i: i \in \calF_j,  y^\dagger_{i\tau}>0} \left(\frac{\varepsilon\cdot p}{\left(\sum_{s=1}^{\tau-1} \sum_{i=1}^n y^\dagger_{is} + 1\right)^r}+\varepsilon\cdot(1-p)\right)\frac{y^\dagger_{i\tau}}{|i:  y^\dagger_{i\tau}>0|}\right)\right)^{1/|\tau:\sum_{i \in \calF_j} y^\dagger_{i\tau}>0 |}
\\\leq & d_\star^2\sum_{j \in \calM(\by^\dagger)}  \sum_{\tau:\sum_{i \in \calF_j} y^\dagger_{i\tau}>0 }\sum_{i: i \in \calF_j,  y^\dagger_{i\tau}>0}\left(\frac{\varepsilon\cdot p}{\left(\sum_{s=1}^{\tau-1} \sum_{i=1}^n y^\dagger_{is} + 1\right)^r}+\varepsilon\cdot(1-p)\right)\frac{y^\dagger_{i\tau}}{|\tau:\sum_{i \in \calF_j} y^\dagger_{i\tau}>0 ||i:  y^\dagger_{i\tau}>0|}    
\\\leq & d_\star^3\sum_{i \in \calN(\by^\dagger)} \sum_{\tau=1}^T \left(\frac{\varepsilon\cdot p}{\left(\sum_{s=1}^{\tau-1} \sum_{i=1}^n y^\dagger_{is} + 1\right)^r}+\varepsilon\cdot(1-p)\right)y^\dagger_{i\tau}
\\\leq & \varepsilon d_\star^4 c_0 (m + \varepsilon \cdot p \cdot \zeta_T \cdot m^{\alpha_T \cdot(1-r)} + \Delta(m)+3\Delta_G(m))
\\\leq& 2\varepsilon d_\star^4 c_0 m
\end{align*}
where in the fourth and sixth inequality, we are using the well-known AM-GM inequality twice. The third last inequality comes from the fact that each $i,\tau$ with $y^\dagger_{i\tau}>0$ has been counted at most $d_\star$ times in the triple sum; the second-to-last inequality comes from Lemma \ref{lem:sol_bound}; and the last inequality holds for sufficiently large $m$. Similarly, since $p\neq1$, the variance of each item satisfies  $\Var(s^\dagger_{jT})>(1-(1-p)\varepsilon)((1-p)\varepsilon)>0$, and therefore $\Var\left(\sum_{j=1}^q s^\dagger_{jT}\right)=\Theta(m)$. This implies:
\[\calO\left(\frac{\max_{i} c_i \cdot \chi^*(\calG)^2 \sum_{i=1}^n \E\left[(X_i-\E[X_i])^3\right]}{\left(\Var\left(\sum_{i=1}^n X_i\right)\right)^{3/2}}\right)=\calO\left(\frac{1}{\left(\Var\left(\sum_{i=1}^n X_i\right)\right)^{1/2}}\right)=o(1)\]

From \cite{janisch2024berry}, we have for sufficiently large $m$:
\[\P\left(\left|\sum_{j=1}^q s_{jT} - \pi_{j}(\by^\dagger)\right|\geq s\sqrt{2\varepsilon d_\star^4 c_0 m}\right)\leq2\Phi(-s) \]

By construction of $\Delta_G(m)$, we obtain:
\[\P\left(\left|\sum_{j=1}^q s_{jT} - \pi_{j}(\by^\dagger)\right|>\Delta_G(m)\right)=\P\left(\left|\sum_{j=1}^q s_{jT} - \pi_{j}(\by^\dagger)\right|> -\Phi^{-1}\left(\frac{\delta}{2}\right)\sqrt{2\varepsilon d_\star^4 c_0 m}\right)\leq\delta \]
\underline{Case 2: $p=1$.}
\begin{align*}
\Var\left(\sum_{j=1}^q s^\dagger_{jT}\right)
%\leq & \chi^*(\calD)\ \sum_{j=1}^q \Var(s^\dagger_{jT})
%\\\leq & d_\star^2 \sum_{j \in \calM(\by^\dagger)} \Var(s^\dagger_{jT})
%\\\leq & d_\star^2\sum_{j \in \calM(\by^\dagger)}  \prod_{\tau=1}^T \left(\frac{\varepsilon}{\left(\sum_{s=1}^{\tau-1} \sum_{i=1}^n y^\dagger_{is} + 1\right)^r}\right)^{\sum_{i \in \calF_j} y^\dagger_{i\tau}}
%\\\leq & d_\star^2\sum_{j \in \calM(\by^\dagger)}  \prod_{\tau=1}^T \sum_{i \in \calF_j} \left(\frac{\varepsilon}{\left(\sum_{s=1}^{\tau-1} \sum_{i=1}^n y^\dagger_{is} + 1\right)^r}\right)\frac{y^\dagger_{i\tau}}{|\calF_j|}
% \leq & d_\star^2\sum_{j \in \calM(\by^\dagger)}  \sum_{\tau=1}^T \sum_{i \in \calF_j} \left(\frac{\varepsilon}{\left(\sum_{s=1}^{\tau-1} \sum_{i=1}^n y^\dagger_{is} + 1\right)^r}\right)\frac{y^\dagger_{i\tau}}{|\calF_j|}    
\leq & d_\star^3\sum_{i \in \calN(\by^\dagger)} \sum_{\tau=1}^T \left(\frac{\varepsilon}{\left(\sum_{s=1}^{\tau-1} \sum_{i=1}^n y^\dagger_{is} + 1\right)^r}\right)y^\dagger_{i\tau}
\\\leq & d_\star^4 \varepsilon\left(\zeta_T \cdot m^{\alpha_T \cdot (1-r)} + o\left(m^{\alpha_T \cdot (1-r)} \right)\right)
\end{align*}
where the first inequality follows the same logic as above, and the last inequality can be derived following the same procedure as Lemma \ref{lem:main_feasible}. As in the proof of Lemma \ref{lem:main_feasible}, we also have $\Var\left(\sum_{j=1}^q s^\dagger_{jT}\right)=\Omega(m^{1-\alpha_T\cdot r(1-r)})$. From \cite{janisch2024berry}, we have for sufficiently large $m$:
\[\P\left(\left|\sum_{j=1}^q s_{jT} - \pi_{j}(\by^\dagger)\right|\geq s\sqrt{2\varepsilon d_\star^4 \zeta_T \cdot m^{\alpha_T \cdot (1-r)} }\right)\leq2\Phi(-s) \]
By construction of $\Delta_G(m)$, we obtain
\[\P\left(\left|\sum_{j=1}^q s_{jT} - \pi_{j}(\by^\dagger)\right|> \Delta_G(m)\right)=\P\left(\left|\sum_{j=1}^q s_{jT} - \pi_{j}(\by^\dagger)\right|> -\Phi^{-1}\left(\frac{\delta}{2}\right)\sqrt{2\varepsilon d_\star^4 \zeta_T \cdot m^{\alpha_T \cdot (1-r)} }\right)\leq\delta \]
This completes the proof.
\hfill\Halmos

\subsection{Statement and proof of Lemmas~\ref{lem:gen_deter_sol_lower},~\ref{lem:prob_bound} and~\ref{lem:network_deter_stochstic}}

Lemma~\ref{lem:gen_deter_sol_lower} provides a sensitivity result providing lower and upper bounds for how much the solution of Problem~\eqref{prob:network_ALG} changes with the coverage target. Then, Lemma~\ref{lem:prob_bound} indicates the (conditional) feasibility of the solution of the deterministic problem, Lemma~\ref{lem:network_deter_stochstic} bounds the difference between the deterministic solution of Algorithm~\ref{ALG_network} and the optimal solution to the stochastic problem~\eqref{prob:main}, proving Part 1 of the theorem.

Let us define $\by^\dagger(m-\Delta)$ as the solution to Problem~\eqref{prob:network_ALG} with the constraint that $m-\Delta$ facilities must open, so the solution to Problem~\eqref{prob:network_ALG} can be written as $\by^\dagger\left(m+\Delta_G(m)\right)$:
\begin{align*}
\by^\dagger(m-\Delta)\in\argmin\left\{\sum_{t=1}^T \sum_{i=1}^n y_{it}\ \mid\ \sum_{j=1}^q \pi_{j}(\by) \geq m-\Delta;\ \by \in \Z^{N \times T}\right\}
\end{align*}

\begin{lemma}\label{lem:gen_deter_sol_lower}
The following holds, as $m,\Delta\to \infty$:
\begin{align*}
   \sum_{t=1}^T \sum_{i=1}^n y_{it}^\dagger(m)-d_\star c_0\left(\Delta+\calO\left(\Delta^{g(r)}\right)\right)\leq\sum_{t=1}^T \sum_{i=1}^n y_{it}^\dagger(m-\Delta)\leq \sum_{t=1}^T \sum_{i=1}^n y_{it}^\dagger(m)-\frac{c_0\Delta}{d_\star}
\end{align*}
\end{lemma}
\proof{Proof:}
Lemma~\ref{lem:main_bound} provides an algorithm that tentatively opens $d_\star c_0 \left(\Delta+\calO\left(\Delta^{g(r)}\right)\right)$ facilities and that opens at least $d_\star\Delta$ facilities on average, for sufficiently large $\Delta$. Again, due to our degree assumption, this implies that the solution will serve at least $\Delta$ customers.

Let us consider the solution $\by^\dagger(m-\Delta)$ and add $d_\star c_0\left(\Delta+\calO\left(\Delta^{g(r)}\right)\right)$ more facilities that connect to customers separate from those that are served in the current solution $\by^\dagger(m-\Delta)$ (those exist per the assumption of Theorem \ref{thm:network}). This new solution serves $\Delta$ more customers than the solution $\by^\dagger(m-\Delta)$ in expectation. By proceeding as in the proof of Lemma~\ref{lem:sol_bound}, we can prove that it is a feasible solution to Problem~\eqref{prob:network_ALG} with a coverage target of $m$. We get the following bound:
\[\sum_{t=1}^T \sum_{i=1}^n y_{it}^\dagger(m)\leq \sum_{t=1}^T \sum_{i=1}^n y_{it}^\dagger(m-\Delta)+d_\star c_0 \left(\Delta+\calO\left(\Delta^{g(r)}\right)\right)\]

Similarly, let us consider the solution $\by^\dagger(m)$ with a coverage target of $m$, and subtract $\frac{c_0 \Delta}{d_\star}$ facilities. By definition of $c_0$, every facility has at most $\frac{1}{c_0}$ chance of succeeding, and therefore, on average at most $\frac{\Delta}{d_\star}$ facilities can succeed. Per our degree assumption, each facility is connected to at most $d_\star$ customers, so $\frac{\Delta}{d_\star}$ facilities are connected to at most $\Delta$ customers. Therefore, the new solution covers at least equal as many customers as $\by^\dagger(m)$ minus $\Delta$ customers. The new solution is feasible for Problem~\eqref{prob:network_ALG} with a coverage target of $m-\Delta$. We obtain:
\[\sum_{t=1}^T \sum_{i=1}^n y_{it}^\dagger(m-\Delta)\leq \sum_{t=1}^T \sum_{i=1}^n y_{it}^\dagger(m)-\frac{c_0 \Delta}{d_\star}\]
This completes the proof of the lemma.
\hfill\Halmos

Lemma~\ref{lem:prob_bound} shows that the optimal stochastic solution satisfies the coverage constraint with high probability conditionally to the ``good'' event $\calE(\by^*)$ (defined in Equation~\eqref{eq:Ek_gen}).

\begin{lemma}\label{lem:prob_bound}
The following holds:
\[\P\left(\sum_{j=1}^q s^*_{jT}\geq m\mid \calE(\by^*) \right)\geq 1-o\left(m^{\alpha_T\cdot(1-r)-1}\right)\]
\end{lemma}
\proof{Proof:}
Note that the optimal solution of Problem~\eqref{prob:network} must satisfy the constraint:
\begin{equation}
\P\left(\sum_{j=1}^q s^*_{jT}\geq m\right)\geq 1-\delta = 1-o\left(m^{\alpha_T\cdot(1-r)-1}\right)\label{eq:Bmdelta_gen}
\end{equation}
Therefore, we have that:
\begin{align*}
\P\left(\sum_{j=1}^q s^*_{jT} \geq m \mid \calE(\by^*)\right)
&=\frac{\P\left(\sum_{j=1}^q s^*_{jT}\geq m\right)-\P\left(\sum_{j=1}^q s^*_{jT}\mid \calE^c(\by^*) \right)\P(\calE^c(\by^*))}{\P(\calE(\by^*))}\\
&\geq1-o\left(m^{\alpha_T\cdot(1-r)-1}\right)-\delta\quad\text{(Equation~\eqref{eq:Bmdelta_gen} and Lemma~\ref{lem:hoeffding_gen_bound})}\\
&=1-o\left(m^{\alpha_T\cdot(1-r)-1}\right)
\end{align*}
This completes the proof of Lemma~\ref{lem:prob_bound}.\hfill\Halmos

Then, Lemma~\ref{lem:network_deter_stochstic} bounds the suboptimality of the deterministic approximation solved in Algorithm~\ref{ALG_network} with a sub-linear term against the stochastic solution.

\begin{lemma}\label{lem:network_deter_stochstic}
The following holds:
\begin{equation*}
m^*(\varepsilon, p,T)
\geq m^\dagger (\varepsilon, T)- \begin{cases}
    \calO\left(m^{1/2}\right) & \text{if }p \neq 1\\
    o\left(m^{\alpha_T\cdot(1-r)}\right) & \text{if }p = 1
\end{cases} 
\end{equation*}
\end{lemma}
\proof{Proof:}
We proceed by conditioning on the ``good'' event $\calE(\by^*)$, as follows:
\begin{align*}
m^*(\varepsilon, p,T)
&=\E\left[\sum_{t=1}^T \sum_{i=1}^n  y^*_{it} \mid \calE(\by^*)\right]\P(\calE(\by^*))+\E\left[\sum_{t=1}^T \sum_{i=1}^n  y^*_{it} \mid \calE^c(\by^*)\right]\P(\calE^c(\by^*))\\
&\geq \E\left[\sum_{t=1}^T \sum_{i=1}^n  y^*_{it} \mid \calE(\by^*)\right]\left(1-\delta\right)\quad\text{(Lemma~\ref{lem:hoeffding_gen_bound})}\\
&=\E\left[\sum_{t=1}^T \sum_{i=1}^n  y^*_{it} \mid \calE(\by^*)\right] + o(m^{\alpha_T\cdot(1-r)})\ \text{(Lemma~\ref{lem:sol_bound})}
\end{align*}

Conditioned on the good event, we have:
\begin{align*}
\sum_{j=1}^q s^*_{jT}&\leq \sum_{j=1}^q \pi_{j}(\by^*)+\left|\sum_{j=1}^q (s^*_{jT}-\pi_{j}(\by^*))\right|\leq\sum_{j=1}^q \pi_{j}(\by^*)+\Delta_G(m)
\end{align*}

By further conditioning on the value of $s^*_{jT}$ and defining $A^*_t = \sum_{i=1}^n y^*_{it}$, we obtain:
\begin{align*}
m^*(\varepsilon, p,T)
&\geq\sum_{\sigma=0}^\infty \E\left[\sum_{t=1}^T A^*_t \mid \calE(\by^*), \sum_{j=1}^q s^*_{jT}=\sigma\right]\P\left(\sum_{j=1}^q s^*_{jT}=\sigma \mid \calE(\by^*) \right) + o(m^{\alpha_T\cdot(1-r)})\\
&\geq\sum_{\sigma=m}^\infty \E\left[\sum_{t=1}^T A^*_t \mid \calE(\by^*), \sum_{j=1}^q s^*_{jT}=\sigma\right]\P\left(\sum_{j=1}^q s^*_{jT}=\sigma \mid \calE(\by^*) \right) + o(m^{\alpha_T\cdot(1-r)})\\
&\geq\E\left[\sum_{t=1}^T A^*_t \mid \calE(\by^*), \sum_{j=1}^q s^*_{jT}=m\right]\P\left(\sum_{j=1}^q s^*_{jT}\geq m \mid \calE(\by^*) \right) + o(m^{\alpha_T\cdot(1-r)})\\
&= \E\left[\sum_{t=1}^T A^*_t \mid \calE(\by^*), \sum_{j=1}^q s^*_{jT}=m, \sum_{j=1}^q \pi_{j}(\by^*)\geq m-\Delta_G(m)\right]\P\left(\sum_{j=1}^q s^*_{jT}\geq m\mid \calE(\by^*) \right) + o(m^{\alpha_T\cdot(1-r)})
\end{align*}
The first term defines a feasible solution to Problem~\eqref{prob:network_ALG} with a right-hand side of $m-\Delta_G(m)$. Per Lemma~\ref{lem:gen_deter_sol_lower}, the following holds, defining again $q=m+\Delta_G(m)$:
\begin{align*}
\sum_{t=1}^T \sum_{i=1}^n y_{it}^\dagger\left(m -  \Delta_G(m)\right)
\geq  &\begin{cases}
\sum_{t=1}^T \sum_{i=1}^n y_{it}^\dagger(m)- \Theta(m^{1/2}) & \text{if }p \neq 1\\
\sum_{t=1}^T \sum_{i=1}^n y_{it}^\dagger(m)- \Theta(m^{1/2\cdot \alpha_T\cdot(1-r)}) & \text{if }p =1 
\end{cases}
\\= & \begin{cases}
\sum_{t=1}^T \sum_{i=1}^n y_{it}^\dagger(q)- \Theta(m^{1/2}) & \text{if }p \neq 1\\
\sum_{t=1}^T \sum_{i=1}^n y_{it}^\dagger(q)- \Theta(m^{1/2\cdot \alpha_T\cdot(1-r)}) & \text{if }p =1 
\end{cases}
\end{align*}
In fact, $\sum_{t=1}^T \sum_{i=1}^n y_{it}^\dagger(q)$ is exactly the objective value achieved with Algorithm~\ref{ALG_network}, that is, $\sum_{t=1}^T \sum_{i=1}^n y_{it}$. Combining this observation with Lemma~\ref{lem:prob_bound}, we obtain:
\begin{align*}
m^*(\varepsilon, p,T)
&\geq \begin{cases}
    \left(\sum_{t=1}^T \sum_{i=1}^n y_{it}^\dagger -  \Theta(m^{1/2}) \right) \times \left(1-o\left(m^{\alpha_T\cdot(1-r)-1}\right)\right)+ o(m^{\alpha_T\cdot(1-r)}) & \text{if }p \neq 1\\
    \left(\sum_{t=1}^T \sum_{i=1}^n y_{it}^\dagger -  \Theta(m^{1/2\cdot \alpha_T\cdot(1-r)})\right) \times \left(1-o\left(m^{\alpha_T\cdot(1-r)-1}\right)\right)+ o(m^{\alpha_T\cdot(1-r)}) & \text{if }p =1
\end{cases}
\end{align*}
Therefore, we conclude:
\begin{align*}
m^*(\varepsilon, p,T)
&\geq m^\dagger (\varepsilon, T)- \begin{cases}
    \calO\left(m^{\max\{\alpha_T\cdot (1-r),1/2\}}\right) & \text{if }p \neq 1\\
     o\left(m^{\alpha_T\cdot(1-r)}\right) & \text{if }p = 1
\end{cases} 
\end{align*}

This completes the proof.
\hfill\Halmos

\subsubsection*{Statement and proof of Lemmas~\ref{lem:bound_loss_gen}~ and~\ref{lem:network_deter_lower}}\

In this section, we assume that $p=1$ (see Figure~\ref{fig:proof_network}). We first provide in Lemma~\ref{lem:bound_loss_gen} lower and upper bounds for the sum of the probabilities of customer coverage under the optimal solution of the fully-learned problem (Problem~\eqref{prob:network_perfect}), which will imply it is ``close'' to the right-hand side target in Problem~\eqref{prob:network_ALG}. Combining this result with Lemma~\ref{lem:gen_deter_sol_lower} , Lemma~\ref{lem:network_deter_lower} relates the solution of Algorithm~\ref{ALG_network} (Problem~\eqref{prob:network_ALG}) to the fully-learned solution (Problem~\eqref{prob:network_perfect}). In particular, the bounds on both directions are important to derive exact regret expressions.

\begin{lemma}\label{lem:bound_loss_gen}
Consider an optimal solution $\widehat{\by}$ to Problem~\eqref{prob:network_perfect}. We have:
\begin{align*}
\sum_{j=1}^q\pi_{j}(\widehat{\by})&\geq m-\sum_{\tau=1}^T \sum_{i =1}^n \left(\frac{\varepsilon}{\left(\sum_{s=1}^{\tau-1}\sum_{i=1}^n \widehat{y}_{is}+1\right)^r}+\right)\widehat{y}_{i\tau}\cdot d_\star\\
\sum_{j=1}^q\pi_{j}(\widehat{\by})&\leq m-\sum_{\tau=1}^T\sum_{i=1}^n\left(\frac{\varepsilon}{\left(\sum_{s=1}^{t-1}\sum_{i=1}^n \widehat{y}_{is}+1\right)^r}\right)\widehat{y}_{i\tau}
\end{align*}
\end{lemma}

\proof{Proof:}

Under solution $\widehat{y}_{it}$, every facility is opened at most once throughout the planning horizon (since $p=1$ and we know the success indicators). This implies that: $\sum_{t=1}^T\widehat{y}_{it}\leq 1, \;\;\forall i=1,\cdots,n$. Therefore, for the facilities $i\in\calN(\widehat{\by})$, we can define the time $t(i)$ as the unique time period where facility $i$ was opened, i.e., $\widehat{y}_{i,t(i)}=1$ and $\widehat{y}_{i\tau}=0$ for all $\tau\neq t(i)$. By convention, $t(i)=0$ and $\widehat{y}_{i,t(i)}=0$ if $i\notin\calN(\widehat{\by})$. The lower bound is obtained as follows:
\begin{align*}
\sum_{j=1}^q\pi_{j}(\widehat{\by})
 =&\sum_{j=1}^q \left[1-\prod_{\tau=1}^T \left(\frac{\varepsilon}{\left(\sum_{s=1}^{\tau-1}\sum_{i=1}^n \widehat{y}_{is}+1\right)^r}\right)^{\sum_{i \in \calF_j} \widehat{y}_{i\tau}}\right]\\
 =&\sum_{j=1}^q \left[1-\prod_{i \in \calF_j} \left(\frac{\varepsilon}{\left(\sum_{s=1}^{t(i)-1}\sum_{i=1}^n \widehat{y}_{is}+1\right)^r}\right)^{\widehat{y}_{i,t(i)}}\right]\\
 =&\sum_{j=1}^q \left[1-\prod_{i \in \calF_j\cap\calN(\widehat{\by})} \frac{\varepsilon}{\left(\sum_{s=1}^{t(i)-1}\sum_{i=1}^n \widehat{y}_{is}+1\right)^r}\right]\\
 =&\sum_{j\in \calM(\widehat{\by})} \left[1-\prod_{i\in\calF_j\cap\calN(\widehat{\by})} \frac{\varepsilon}{\left(\sum_{s=1}^{t(i)-1}\sum_{i=1}^n \widehat{y}_{is}+1\right)^r}\right]\\
 \geq& \sum_{j\in \calM(\widehat{\by})} \left[1-\sum_{i\in\calF_j\cap\calN(\widehat{\by})} \frac{\varepsilon}{\left(\sum_{s=1}^{t(i)-1}\sum_{i=1}^n \widehat{y}_{is}+1\right)^r}\right]\\
 =&\sum_{j\in \calM(\widehat{\by})} \left[1-\sum_{i \in \calF_j} \frac{\varepsilon\cdot \widehat{y}_{i,t(i)}}{\left(\sum_{s=1}^{t(i)-1}\sum_{i=1}^n \widehat{y}_{is}+1\right)^r}\right]\\
 =&|\calM(\widehat{\by})|- \sum_{i=1}^n \sum_{j \in \calC_i\cap\calM(\widehat{\by})}  \frac{\varepsilon\cdot \widehat{y}_{i,t(i)}}{\left(\sum_{s=1}^{t(i)-1}\sum_{i=1}^n \widehat{y}_{is}+1\right)^r}y\\
 \geq& m-\sum_{i =1}^n \frac{\varepsilon\cdot \widehat{y}_{i,t(i)}\cdot d_\star}{\left(\sum_{s=1}^{t(i)-1}\sum_{i=1}^n \widehat{y}_{is}+1\right)^r}\\
 =& m-\sum_{\tau=1}^T \sum_{i =1}^n \frac{\varepsilon\cdot \widehat{y}_{i\tau}\cdot d_\star}{\left(\sum_{s=1}^{\tau-1}\sum_{i=1}^n \widehat{y}_{is}+1\right)^r},
\end{align*}
where the first inequality stems from the fact that $ab\leq a+b$ for any $a,b\in[0,1]$, and the second inequality stems from the feasibility of the solution $\widehat{\by}$ (which implies that $|\calM(\widehat{\by})|\geq m$) and from the fact that $|\calC_i|\leq d_\star$ per our degree assumption.

We denote by $\calM_t(\widehat{\by})$ the set of customers first tentatively covered in period $t$:
\begin{align*}
\calM_t(\widehat{\by})&=\left\{j\in\{1,\cdots,q\}:\exists i\in\calF_j:y_{i1}=\cdots=y_{i,t-1}=0,\ y_{it}=1,\ y_{k\tau}=0,\ \forall \tau\leq t,\forall k\in\calF_j\setminus\{i\}\right\}
\end{align*}
Since $\widehat{\by}$ is the optimal solution to the fully-learned problem, every opened facility must serve at least one customer that is not served by any other facility (otherwise, this facility can be removed, contradicting the optimality of the solution). Thus, if $\widehat{y}_{it(i)}=1$, we can denote such customer by $\widehat{j}(i)$. Then, we have:
\begin{align*}
\sum_{j=1}^q \pi_{j}(\widehat{\by})
=&\sum_{j=1}^q\left[1-\prod_{\tau=1}^T \left(\frac{\varepsilon}{\left(\sum_{s=1}^{\tau-1}\sum_{i=1}^n \widehat{y}_{is}+1\right)^r}\right)^{\sum_{i \in \calF_j} \widehat{y}_{i\tau}}\right]\\
=&\sum_{t=1}^T\sum_{j \in \calM_t(\widehat{\by})} \left[1-\prod_{\tau=t}^T \left(\frac{\varepsilon}{\left(\sum_{s=1}^{\tau-1}\sum_{i=1}^n \widehat{y}_{is}+1\right)^r}\right)^{\sum_{i \in \calF_j} \widehat{y}_{i\tau}}\right]\\
\leq &\left(\sum_{t=1}^T\sum_{j \in \calM_t(\widehat{\by})} 1\right)-\left(\sum_{t=1}^T\sum_{j \in \calM_t(\widehat{\by}), \exists k\;s.t.\;j=\widehat{j}(k)}\prod_{\tau=t}^T \left(\frac{\varepsilon}{\left(\sum_{s=1}^{\tau-1}\sum_{i=1}^n \widehat{y}_{is}+1\right)^r}\right)^{\sum_{i \in \calF_j} \widehat{y}_{i\tau}}\right]\\
\overset{{(*)}}{=}& m-\left(\sum_{t=1}^T\sum_{j \in \calM_t(\widehat{\by}), \exists k\;s.t.\;j=\widehat{j}(k)} \left(\frac{\varepsilon}{\left(\sum_{s=1}^{t-1}\sum_{i=1}^n \widehat{y}_{is}+1\right)^r}\right)\widehat{y}_{kt}\right)\\
=&m-\left(\sum_{i=1}^n \frac{\varepsilon\cdot \widehat{y}_{i,t(i)}}{\left(\sum_{s=1}^{t(i)-1}\sum_{i=1}^n \widehat{y}_{is}+1\right)^r}\right)\\
=&m-\sum_{\tau=1}^T\sum_{i=1}^n\frac{\varepsilon\cdot \widehat{y}_{i\tau}}{\left(\sum_{s=1}^{t-1}\sum_{i=1}^n \widehat{y}_{is}+1\right)^r}
\end{align*}

The starred equality comes from two observations. First, the coverage constraint is binding (i.e., $|\calM(\widehat{\by})|=m$) due to the optimality of the fully-learned solution. Second, for $j\in \calM_t(\widehat{\by})$ (i.e., customer $j$ is first tentatively covered at time $t$) such that there exists $k$ such that $j=\widehat{j}(k)$ (i.e., customer $j$ is only covered by facility $k$), the following holds: (i) $\widehat{y}_{kt}=1$; (ii) $\widehat{y}_{it} =0,\ \forall i\in\calF_{\widehat{j}(k)}\setminus\{k\}$; and (iii) $\widehat{y}_{i\tau} =0,\ \forall i\in\calF_{\widehat{j}(k)},\ \forall \tau\neq t$. This concludes the proof of the lemma.
\hfill\Halmos

We can now turn to Lemma~\ref{lem:network_deter_lower}, which proves that the deterministic approximation solved in Algorithm~\ref{ALG_network} achieves sub-linear regret against the fully-learned optimum.

\begin{lemma}\label{lem:network_deter_lower}
The following holds if $p=1$, as $m\to\infty$: $m^\dagger(\varepsilon, p,T) =  \widehat{m}(\varepsilon,p) + \Theta\left(m^{g(r)}\right)$.
\end{lemma}

\proof{Proof:}
Let $\by^{\textsc{L-OPT}}$ denote an optimal solution to Problem~\eqref{prob:network_perfect}. Recall that each facility is tentatively opened at most once in that solution, i.e., $\sum_{t=1}^T{y}^{\textsc{L-OPT}}_{it}\in\{0,1\}$ for all $i=1,\cdots,n$. Moreover, the success of each facility is time-invariant, i.e., $S_{it}=S_i$ for all $t=1,\cdots,T$. Therefore, any facility can be opened in any time period in the fully-learned solution. Out of all optimal solutions to Problem~\eqref{prob:network_perfect}, we select one that solves the following problem:
\begin{align*}
\min_{\by\in\{0,1\}^{T \times N}} \quad & \sum_{i=1}^n\sum_{\tau=1}^T\frac{\varepsilon\cdot y_{it}}{\left(\sum_{s=1}^{\tau-1}\sum_{i=1}^n y_{is}+1\right)^r} \nonumber\\
\st\quad& \sum_{t=1}^Ty_{it}=\sum_{t=1}^T{y}^{\textsc{L-OPT}}_{it},\ \forall i=1,\cdots,n\nonumber
\end{align*}
By construction, the solution is feasible in Problem~\eqref{prob:network_perfect} because of time invariance. Also by construction, the solution achieves the same cost as $\by^{\textsc{L-OPT}}$. It therefore remains an optimal solution to Problem~\eqref{prob:network_perfect}. We denote it by $\widehat{\by}$.

We operate the change of variables $A^{\textsc{L-OPT}}_t = \sum_{i=1}^ny^{\textsc{L-OPT}}_{it}$ and $\widehat{A}_t = \sum_{i=1}^n \widehat{y}_{it}$. Then, $\widehat{\bA}$ is the optimal solution to the following problem:
\begin{align}
\max_{\mathbf{A}\in\{0,1\}^T} \quad & \sum_{t=1}^TA_t- \sum_{\tau=1}^T\frac{\varepsilon\cdot A_t}{\left(\sum_{s=1}^{\tau-1}A_s+1\right)^r} \label{prob:network_perfect_ties} \nonumber\\
\st\quad& \sum_{t=1}^TA_t=\sum_{t=1}^T A^{\textsc{L-OPT}}_{it},\ \forall i=1,\cdots,n\nonumber
\end{align}

This problem now has an equivalent format to  Problem~\eqref{prob:main_det} in Section~\ref{sec:base}. Following the same procedure as Lemma~\ref{lem:main_det}, we can derive a solution such that $\sum_{t=1}^TA_t=m+K(T,\varepsilon,p)\cdot m^{g(r)}+o\left(m^{g(r)}\right)$ and $\sum_{t=1}^TA_t- \sum_{\tau=1}^T\left(\frac{\varepsilon}{(\sum_{s=1}^{\tau-1}A_s+1)^r}\right)A_t\geq\sum_{t=1}^T A^{\textsc{L-OPT}}_{it}\geq m$, so that
\begin{align*}
\sum_{\tau=1}^T\frac{\varepsilon\cdot \widehat{A}_t}{\left(\sum_{s=1}^{\tau-1}\widehat{A}_s+1\right)^r}
&= \sum_{i=1}^n\sum_{\tau=1}^T\frac{\varepsilon\cdot \widehat{y}_{it}}{\left(\sum_{s=1}^{\tau-1}\sum_{i=1}^n \widehat{y}_{is}+1\right)^r}
=K(T,\varepsilon,p)\cdot m^{g(r)} + o\left(m^{g(r)}\right)
\end{align*}
for some constant $K(T,\varepsilon,p)>0$ that is absolutely bounded above by a constant $K_0<\infty $.

By Lemma~\ref{lem:bound_loss_gen}, we derive:
$$m -d_\star K(T,\varepsilon,p)m^{g(r)}+o\left(m^{g(r)}\right)\leq \sum_{j=1}^q\pi_{j}(\widehat{\by})\leq m -K(T,\varepsilon,p)m^{g(r)}+o\left(m^{g(r)}\right)$$

Therefore, the fully-learned solution $\widehat{\by}$ is feasible in Problem~\eqref{prob:network_ALG} with right-hand side $m - d_\star K(T,\varepsilon,p)\cdot m^{\alpha_T \cdot (1-r)}+o\left(m^{\alpha_T\cdot(1-r)}\right)$, that is:
$$\sum_{t=1}^T \sum_{i=1}^n y_{it}^\dagger\left(m - d_\star K(T,\varepsilon,p)\cdot m^{g(r)}+o\left(m^{g(r)}\right)\right)\leq \widehat{m}(\varepsilon,p)$$

Let us again denote $q=m+\Delta_G(m)$. Lemma~\ref{lem:gen_deter_sol_lower} implies:
\begin{align*}
&\sum_{t=1}^T \sum_{i=1}^n y_{it}^\dagger\left(m - d_\star K(T,\varepsilon,p)\cdot m^{g(r)}+o\left(m^{g(r)}\right)\right)\\
&\geq \sum_{t=1}^T \sum_{i=1}^n y_{it}^\dagger(m)-d_\star^2 K(T,\varepsilon,p) m^{g(r)} + o\left(m^{g(r)}\right)\\
&\geq \sum_{t=1}^T \sum_{i=1}^n y_{it}^\dagger(q)-d_\star \left((q-m)+\calO\left((q-m)^{g(r)}\right)\right)-d_\star^2  K(T,\varepsilon,p) m^{g(r)} + o\left(m^{g(r)}\right)\\
&=	 \sum_{t=1}^T \sum_{i=1}^n y_{it}^\dagger(q)-d_\star^2 K'(T,\varepsilon,p) m^{g(r)} + o\left(m^{g(r)}\right) 
\end{align*}
with
$$K'(T,\varepsilon,p)=\begin{cases}
K(T,\varepsilon,p)& \text{if }p=1 \text{ or } (p \neq 1 \text{ and } \alpha_T\cdot(1-r)>1/2)\\
K'(T,\varepsilon,p)+1 & \text{if }p \neq 1\text{ and } \alpha_T\cdot(1-r)\leq 1/2        
\end{cases} $$
Combining these observations, we obtain:
$$\widehat{m}(\varepsilon,p)\geq\sum_{t=1}^T \sum_{i=1}^n y_{it}^\dagger(q)-\d_\star^2 K'(T,\varepsilon,p)\cdot m^{g(r)}+ o\left(m^{g(r)}\right)$$

Moreover, the fully-learned solution $\widehat{\by}$ is optimal in Problem~\eqref{prob:network_ALG} with right-hand side $\sum_{j=1}^q \pi_{j}(\widehat{\by})$. Therefore, we have:
\begin{align*}
\widehat{m}(\varepsilon,p)&\leq \sum_{t=1}^T \sum_{i=1}^n y_{it}^\dagger\left(\sum_{j=1}^q \pi_{j}(\widehat{\by})\right)
\leq \sum_{t=1}^T \sum_{i=1}^n y_{it}^\dagger\left(m-K(T,\varepsilon,p)\cdot m^{g(r)}+ o\left(m^{g(r)}\right)\right)
\end{align*}

Again, using Lemma~\ref{lem:gen_deter_sol_lower}, we have:
\begin{align*}
&\sum_{t=1}^T \sum_{i=1}^n y_{it}^\dagger\left(m - K(T,\varepsilon,p)\cdot m^{g(r)}+o\left(m^{\alpha_T\cdot(1-r)}\right)\right)\\
&\leq \sum_{t=1}^T \sum_{i=1}^n y_{it}^\dagger(m)-\frac{1}{d_\star}K(T,\varepsilon,p)\cdot m^{g(r)} + o\left(m^{g(r)}\right)\\
&\leq \sum_{t=1}^T \sum_{i=1}^n y_{it}^\dagger(q)-\frac{1}{d_\star}\left((q-m)+\calO\left((q-m)^{g(r)}\right)\right)-\frac{1}{d_\star}K(T,\varepsilon,p)\cdot m^{g(r)} + o\left(m^{g(r)}\right)\\
&= \sum_{t=1}^T \sum_{i=1}^n y_{it}^\dagger(q)-\frac{1}{d_\star}K'(T,\varepsilon,p)\cdot m^{g(r)} + o\left(m^{g(r)}\right)
\end{align*}

Combining these observations, we obtain:
$$\widehat{m}(\varepsilon,p)\leq\sum_{t=1}^T \sum_{i=1}^n y_{it}^\dagger(q)-\frac{1}{d_\star}K'(T,\varepsilon,p)\cdot m^{g(r)} + o\left(m^{g(r)}\right)$$
We conclude that:
\begin{align*}
m^\dagger(\varepsilon, p,T)
&= \widehat{m}(\varepsilon,p)+\Theta\left(m^{g(r)}\right)
\end{align*}
This completes the proof of Lemma~\ref{lem:network_deter_lower}. \hfill\Halmos

\subsubsection*{Proof of Theorem~\ref{thm:network}}

\paragraph{Part 1.}

We prove that Algorithm~\ref{ALG_network} yields a feasible solution to~\eqref{prob:network}. By definition of Problem~\eqref{prob:network_ALG}:
\[\sum_{j=1}^q \pi_{j}(\by^\dagger) \geq m + \Delta_G(m)\]

Conditioned on $\calE(\by^\dagger)$, we have:
\begin{align*}
\sum_{j=1}^q s^\dagger_{jT}&\geq \sum_{j=1}^q \pi_{j}(\by^\dagger)-\left|\sum_{j=1}^q (s^\dagger_{jT}-\pi_{j}(\by^\dagger)\right|\geq\sum_{j=1}^q \pi_{j}(\by^\dagger)-\Delta_G(m)\geq m
\end{align*}

This implies from Lemma \ref{lem:hoeffding_gen_bound} that $\P\left(\sum_{j=1}^q s^\dagger_{jT}\geq m\right)\geq \P(\calE(\by^\dagger))\geq 1-\delta$. Therefore, the decision vector $\by^\dagger$ defines a feasible solution to Problem~\eqref{prob:network}. From Lemma~\ref{lem:network_deter_stochstic}, we also have
$$m^\dagger(\varepsilon, p,T)\leq m^*(\varepsilon, p,T)+\begin{cases}
o\left(m^{\alpha_T \cdot (1-r)}\right)& \text{if }p=1 \text{ or } (p \neq 1 \text{ and } \alpha_T\cdot(1-r)>1/2)\\
\calO(m^{1/2}) & \text{if }p \neq 1\text{ and } \alpha_T\cdot(1-r)\leq 1/2        
\end{cases} $$

\paragraph{Part 2.} This is obtained from Lemmas~\ref{lem:network_deter_stochstic} and~\ref{lem:network_deter_lower}, when $p=1$:
\begin{align*}
   & m^\dagger(\varepsilon, p,T)\leq m^*(\varepsilon, p,T)+o\left(m^{\alpha_T\cdot(1-r)}\right)\\
   & m^\dagger(\varepsilon, p,T) =  \widehat{m}(\varepsilon,p) +\Theta(m^{g(r)}) =  \widehat{m}(\varepsilon,p) +\Theta(m^{\alpha_T(1-r)})
\end{align*}
This proves Part 2 of the theorem: $m^*(\varepsilon, p,T) =  \widehat{m}(\varepsilon,p) +\Theta(m^{\alpha_T(1-r)}) =  \widehat{m}(\varepsilon,p)+\Theta(m^{g(r)})$.
\hfill\Halmos

\subsection{Proof of Proposition~\ref{prop:network_nolearn}}

As in Lemma~\ref{lem:gen_deter_sol_lower}, let us define $\by^{\mathrm{NL}}(m-\Delta)$ as the solution to Problem~\eqref{prob:network_nolearn} with the constraint that $m-\Delta$ facilities must open with probability $1-\delta$:
\begin{align*}
\by^{NL}(m-\Delta)\in\argmin \quad & \sum_{t=1}^T \sum_{i=1}^n y_{it} \\
\st\quad&\P\left[\sum_{j=1}^q s_{jT} \geq m-\Delta\right] \geq 1-\delta\\
\quad& s_{jT}\sim \Ber\left(1-\prod_{\tau=1}^T \varepsilon^{\sum_{i \in \calF_j} y_{i\tau}}\right),\quad \forall j\in\calC\\
\quad& y_{it}\in\{0,1\},\ \forall i\in \calF, \ \forall t\in\calT
\end{align*}

By the same argument as in Lemma~\ref{lem:gen_deter_sol_lower}, we have:
\[\sum_{t=1}^T \sum_{i=1}^n y_{it}^{\mathrm{NL}}(m-\Delta) \leq\sum_{t=1}^T \sum_{i=1}^n y_{it}^{NL}(m) - \frac{c_0\Delta}{d_\star} \]

Recall that $\sum_{t=1}^T\widehat{y}_{it}\leq 1, \;\;\forall i=1,\cdots,n$ and that $|\calM(\widehat{\by})|=m$. The expected customer coverage of the fully-learned solution in the no-learning problem satisfies:
\begin{align*}
\E\left[\sum_{j=1}^q s_{jT}\right]
&= \sum_{j=1}^q\left(1-\prod_{\tau=1}^T \varepsilon^{\sum_{i \in \calF_j} \widehat{y}_{i\tau}} \right)\\
&=\sum_{j \in \calM(\widehat{\by})} \left(1 - \prod_{\tau=1}^T \varepsilon^{\sum_{i \in \calF_j} \widehat{y}_{i\tau}}\right) \\
&=\sum_{j \in \calM(\widehat{\by})} \left(1 - \varepsilon^{\sum_{i \in \calF_j} \sum_{\tau=1}^T\widehat{y}_{i\tau}}\right) \\
&\leq\sum_{j \in \calM(\widehat{\by})} \left(1 - \varepsilon^{d_\star}\right) \\
&=  m - m \varepsilon^{d_\star}
\end{align*}

By the same argument as Lemma \ref{lem:network_deter_lower},
\[\widehat{m}(\varepsilon,p) \leq \sum_{t=1}^T \sum_{i=1}^n \by^{\mathrm{NL}}(m-m \varepsilon^{d_\star})\leq \sum_{t=1}^T \sum_{i=1}^n \by^{\mathrm{NL}}(m) - \frac{c_0m \varepsilon^{d_\star}}{d_\star}=m^{\mathrm{NL}}(\varepsilon, T) - \frac{c_0m \varepsilon^{d_\star}}{d_\star}.\]
This proves that $m^{\mathrm{NL}}(\varepsilon, T)\geq \widehat{m}(\varepsilon,p)+\Omega(m)$.
\hfill\Halmos

\section{Computational Algorithm to Solve Problem~\eqref{prob:network_ALG}}
\label{app:network_ALG}

Recall that the deterministic approximation in Algorithm~\ref{ALG_network} involves a mixed-integer non-convex optimization problem (Problem~\eqref{prob:network_ALG}), given as follows:
\begin{align*}
\min \quad & \sum_{t=1}^T \sum_{i=1}^n y_{it}\\
\quad& \sum_{j=1}^q \left(1-\prod_{\tau=1}^T \left(\frac{\varepsilon\cdot p }{\left(\sum_{s=1}^{\tau-1} \sum_{i=1}^n y_{is} +1\right)^r}+\varepsilon\cdot(1-p)\right)^{\sum_{i \in \calF_j} y_{i\tau}}\right) \geq m+\Delta_G(m),\\
\quad& y_{it}\in\{0,1\},\ \forall i\in \calF, \ \forall t\in\calT
\end{align*}

We propose in this appendix a computational algorithm to solve it efficiently. The algorithm proceeds by opening facilities by increasing order of expected incremental coverage: it ignores the least impactful facilities, initially targets moderately impactful ones for learning-based exploration, and then opens the most impactful ones for optimization-based exploitation. We show that the solution is optimal in star graphs and provides a principled heuristic in general bipartite graphs.

\subsection{Optimal algorithm in star graphs.}

Proposition~\ref{prop:small_first} characterizes the algorithm's solution.

\begin{proposition}\label{prop:small_first}
Let $y^\dagger_{it}$ be the solution of Algorithm~\ref{ALG_network} and $\pi_t(\by^\dagger)=\frac{\varepsilon\cdot p}{\left(\sum_{s=1}^{\tau-1} \sum_{i=1}^n y^\dagger_{is} +1\right)^r}+\varepsilon\cdot(1-p)$. Consider $i_1\neq i_2$ such that $y^\dagger_{i_1,t}=y^\dagger_{i_2, \tau}=1$ and $y^\dagger_{i_1,\tau}=y^\dagger_{i_2, t}=0$, with $t<\tau$. Then:
\begin{align*}
    &\sum_{j \in \calC_{i_1} \setminus \calC_{i_2}}^q \prod_{\tau=1}^T \pi_t(\by^\dagger)^{\sum_{i \in \calF_j} y^\dagger_{iv}-\mathbf{1}\{v=t\}}
\leq \sum_{j \in \calC_{i_2} \setminus \calC_{i_1}}^q \prod_{\tau=1}^T \pi_t(\by^\dagger)^{\sum_{i \in \calF_j} y^\dagger_{iv}-\mathbf{1}\{v=\tau\}}
\end{align*}
Similarly, if $y^\dagger_{i_2,\tau}=1$, $y^\dagger_{i_2,t}=0,\ \forall t\neq\tau$ and $y^\dagger_{i_1,t}=0,\ \forall t=1,\cdots,T$ then:
\begin{align*}
    &\sum_{j \in \calC_{i_1} \setminus \calC_{i_2}}^q \prod_{\tau=1}^T \pi_t(\by^\dagger)^{\sum_{i \in \calF_j} y^\dagger_{iv}}
\leq \sum_{j \in \calC_{i_2} \setminus \calC_{i_1}}^q \prod_{\tau=1}^T \pi_t(\by^\dagger)^{\sum_{i \in \calF_j} y^\dagger_{iv}-\mathbf{1}\{v=\tau\}}
\end{align*}
\end{proposition}

This result sheds light on the exploration-exploitation trade-off to balance the optimization and learning objectives. Specifically, the decision-maker disregards facilities with low marginal expected coverage, to maximize the impact of opened facilities. Then, they prioritize facilities with moderate coverage early on to enable exploration but still contribute toward the coverage target if successful. The facilities with the largest incremental coverage are opened later on for exploitation purposes once uncertainty is mitigated. Corollary~\ref{cor:small_first} characterizes the exact solution in star graphs, when every customer is connected to exactly one facility. This star structure eliminates interdependencies across facilities, leading to a threshold-based solution: a facility gets selected if and only if its degree is larger than $n_0$; and selected facilities then get selected by non-decreasing degree order.

\begin{corollary}\label{cor:small_first}
Assume that every customer is connected to one facility, i.e., $|\calF_j|=1,\ \forall j\in\calC$, and that every facility is opened at most once, i.e., $\sum_{t=1}^T y_{it}^\dagger \leq 1 \;\;\forall i$. Without loss of generality, arrange the set of facilities in increasing degree order (i.e. $\deg(i_1)\leq \deg(i_2)$ for all $i_1\leq i_2$). There exist $0\leq n_0\leq n_1 \leq \cdots \leq n_T=n$ such that the optimal solution $y_{it}^\dagger$ is of the form:
\begin{align*}
    y_{it}=\begin{cases}
        1 & n_{t-1}<i \leq n_{t} \\
        0 & \text{otherwise}
    \end{cases}
\end{align*}
\end{corollary}

In star graphs, Problem~\eqref{prob:network_ALG} therefore involves finding the largest possible threshold $n_0$ such that there exist subsequent thresholds $n_0\leq n_1 \leq \cdots \leq n_T=n$ satisfying
\begin{align*}
\sum_{i=n_0+1}^n \deg(i)-\sum_{t=1}^T\sum_{i=n_{t-1}+1}^{n_t}\left(\frac{\varepsilon \deg(i)}{\left(n_{t}-n_0 +1\right)^r}+\varepsilon\cdot(1-p)\right)\geq m+\Delta_G(m)
\end{align*}

For any value of $n_0$, we therefore solve the following subproblem:
\begin{equation}\label{eq:SP}
\min\quad\left\{\sum_{t=1}^T\sum_{i=n_{t-1}+1}^{n_t}\left(\frac{\varepsilon\deg(i)}{\left(n_{t}-n_0 +1\right)^r}+\varepsilon\cdot(1-p)\right)\ \bigg|\ n_0\leq n_1 \leq \cdots \leq n_T=n\right\}\tag{$\calP_0^\dagger$}.
\end{equation}

Problem~\eqref{eq:SP} can be cast as a shortest path problem over $(n-n_0+1)(T-1)+2$ nodes. The network representation, shown in Figure~\ref{fig:SP}, includes one source and a sink, and one node for each value of $n_t\in\{n_0,\cdots,n\}$ for each period $t\in\{1,\cdots,T-1\}$. For $t\leq T-2$, each node $n_t$ is connected to the nodes $n_{t+1}\geq n_t$, with an arc cost of $\sum_{i=n_{t-1}+1}^{n_t}\left(\frac{\varepsilon\deg(i)}{\left(n_{t}-n_0 +1\right)^r}+\varepsilon\cdot(1-p)\right)$. Finally, each node $n_{T-1}$ is connected to the sink, with an arc cost of $\sum_{i=n_{T-1}+1}^{n}\left(\frac{\varepsilon\deg(i)}{\left(n-n_0 +1\right)^r}+\varepsilon\cdot(1-p)\right)$. This shortest path problem can be solved in polynomial time in $n$, hence in polynomial time in $m$.

\begin{figure}[h!]
\centering
\begin{tikzpicture}
\node[draw, circle] (source) at (-4,3) {\color{white}0};
\node[draw, circle] (n1t1) at (0,5) {3};
\node[draw, circle] (n2t1) at (0,4) {4};
\node[draw, circle] (n3t1) at (0,3) {5};
\node[draw, circle] (n4t1) at (0,2) {6};
\node[draw, circle] (n5t1) at (0,1) {7};
\node[draw, circle] (n1t2) at (4,5) {3};
\node[draw, circle] (n2t2) at (4,4) {4};
\node[draw, circle] (n3t2) at (4,3) {5};
\node[draw, circle] (n4t2) at (4,2) {6};
\node[draw, circle] (n5t2) at (4,1) {7};
\node[draw, circle] (n1t3) at (8,5) {3};
\node[draw, circle] (n2t3) at (8,4) {4};
\node[draw, circle] (n3t3) at (8,3) {5};
\node[draw, circle] (n4t3) at (8,2) {6};
\node[draw, circle] (n5t3) at (8,1) {7};
\node[draw, circle] (sink) at (12,3) {\color{white}0};
\node at (0,0) {$t=1$};
\node at (4,0) {$t=2$};
\node at (8,0) {$t=3$};
\draw[->] (source) to (n1t1);
\draw[->,ultra thick,color=myred] (source) to (n2t1);
\draw[->] (source) to (n3t1);
\draw[->] (source) to (n4t1);
\draw[->] (source) to (n5t1);
\draw[->] (n1t1) to (n1t2);
\draw[->] (n1t1) to (n2t2);
\draw[->] (n1t1) to (n3t2);
\draw[->] (n1t1) to (n4t2);
\draw[->] (n1t1) to (n5t2);
\draw[->,ultra thick,color=myred] (n2t1) to (n2t2);
\draw[->] (n2t1) to (n3t2);
\draw[->] (n2t1) to (n4t2);
\draw[->] (n2t1) to (n5t2);
\draw[->] (n3t1) to (n3t2);
\draw[->] (n3t1) to (n4t2);
\draw[->] (n3t1) to (n5t2);
\draw[->] (n4t1) to (n4t2);
\draw[->] (n4t1) to (n5t2);
\draw[->] (n5t1) to (n5t2);
\draw[->] (n1t2) to (n1t3);
\draw[->] (n1t2) to (n2t3);
\draw[->] (n1t2) to (n3t3);
\draw[->] (n1t2) to (n4t3);
\draw[->] (n1t2) to (n5t3);
\draw[->] (n2t2) to (n2t3);
\draw[->] (n2t2) to (n3t3);
\draw[->,ultra thick,color=myred] (n2t2) to (n4t3);
\draw[->] (n2t2) to (n5t3);
\draw[->] (n3t2) to (n3t3);
\draw[->] (n3t2) to (n4t3);
\draw[->] (n3t2) to (n5t3);
\draw[->] (n4t2) to (n4t3);
\draw[->] (n4t2) to (n5t3);
\draw[->] (n5t2) to (n5t3);
\draw[->] (n1t3) to (sink);
\draw[->] (n2t3) to (sink);
\draw[->] (n3t3) to (sink);
\draw[->,ultra thick,color=myred] (n4t3) to (sink);
\draw[->] (n5t3) to (sink);
\end{tikzpicture}
\caption{Shortest-path representation of Problem~\eqref{eq:SP}, with $n=7$, $n_0=3$ and $T=4$. The path highlighted in red corresponds to $n_1=4$, $n_2=4$, $n_3=6$, and $n_4=7.$}\label{fig:SP}
\end{figure}

If the optimum of Problem~\eqref{eq:SP} is less than $\frac{1}{\varepsilon}\left(\sum_{i=n_0+1}^n \deg(i)-m-\Delta_G(m)\right)$, then the solution to Problem~\eqref{eq:SP} is feasible for Problem~\eqref{prob:network_ALG}. Otherwise, $n_0$ needs to be increased to guarantee feasibility. We proceed by binary search over $n_0$ (Algorithm~\ref{alg:network_simp}). Combined with the polynomial-time shortest path subproblem, this algorithm solves Problem~\eqref{prob:network_ALG} in star graphs to optimality in polynomial time in $n$, hence in polynomial time in $m$ (Proposition~\ref{prop:exact}).

\begin{algorithm} [h!]
\caption{Algorithm to solve Problem~\eqref{prob:network_ALG} in star graph $\calG$: \texttt{STAR}($\calG$).}\small
\label{alg:network_simp}
\begin{algorithmic}[1]
\item \textbf{Initialization:} $\underline{n}_0=0,\overline{n}_0=n, n_0=\left\lfloor\frac{\overline{n}_0+\underline{n}_0}{2}\right\rfloor$.
\item Iterate between Steps 1-3, until $\overline n_0-\underline n_0\leq 1$.
\begin{itemize}
\item[] \textbf{Step 1.} Solve Problem~\eqref{eq:SP} as a shortest path problem; retrieve solution $n_1^*(n_0),\cdots, n_{T-1}^*(n_0)$.
\item[] \textbf{Step 2.} If $\displaystyle\sum_{i=n_0+1}^n \deg(i)-\sum_{t=1}^T\sum_{i=n_{t-1}^*(n_0)+1}^{n_t}\left(\frac{\varepsilon \deg(i)}{\left(n_{t}^*(n_0)-n_0 +1\right)^r+\varepsilon\cdot(1-p)}\right)\geq m+\Delta_G(m)$, update $\overline{n}_0 = n_0$; otherwise, update $\underline{n}_0 = n_0+1$.
\item[] \textbf{Step 3.} Update $n_0 = \left\lfloor\frac{\overline{n}_0 + \underline{n}_0}{2}\right\rfloor$.
\end{itemize}
\item \textbf{Output:} Set $n_0^*\gets\overline{n}_0$; return $n-n_0^*$ as the optimal objective value of Problem~\eqref{prob:network_ALG}; use $n_0^*$, $n_1^*(n_0^*),\cdots, n_{T-1}^*(n_0^*)$ to reconstruct an optimal solution of Problem~\eqref{prob:network_ALG} as given in Corollary~\ref{cor:small_first}.
\end{algorithmic}
\end{algorithm}

\begin{proposition}\label{prop:exact}
Algorithm~\ref{alg:network_simp} returns the optimal solution of Problem~\eqref{prob:network_ALG} in polynomial time in $n$ (hence, in polynomial time in $m$) if $\calG$ is a star graph (i.e., if $|\calF_j|\leq1,\ \forall j\in\calC$).
\end{proposition}

\subsection{A heuristic in general graphs.}

We leverage the exact algorithm in star graphs to devise a principled heuristic for Problem~\eqref{prob:network_ALG} in general bipartite graphs. This approach decomposes any bipartite graph $\calG$ with maximum degree $d_\star$ into a weighted star graph $\calG_w$ by duplicating every customer $j\in\calC$ into $|\calF_j|$ artificial customers---once per connected facility, as shown in Figure~\ref{fig:graph_decompose}. By design, each artificial customer is connected to exactly one facility, so the decomposition recovers a star graph structure at the expense of ignoring interdependencies across facility-customer pairs. Thus, we weigh each connection in the star graph by $1/d_\star\leq w \leq 1$, so that each successful facility contributes a fraction $w$ per customer toward total coverage. The resulting problem is formulated as Problem~\eqref{prob:network_ALG}, except that the degree function is modified as $\text{deg}(i)=w|\calC_i|$.

\begin{figure}[h!]
\centering
\begin{tikzpicture}
% Original bipartite graph on the left
\node[draw, circle] (j) at (-2,0) {$j$};
\node[draw, circle] (i1) at (-4,2) {$i_1$};
\node[draw, circle] (i2) at (-2,2) {$i_2$};
\node[draw, circle] (i3) at (0,2) {$i_k$};

\draw (i1) -- (j);
\draw (i2) -- (j);
\draw (i3) -- (j);

\node at (-2,-1) {Original graph $\calG$};

% Decomposed bipartite graph on the right
\node[draw, circle] (j1) at (4,0) {$j_1$};
\node[draw, circle] (j2) at (6,0) {$j_2$};
\node[draw, circle] (j3) at (8,0) {$j_3$};

\node[draw, circle] (di1) at (4,2) {$i_1$};
\node[draw, circle] (di2) at (6,2) {$i_2$};
\node[draw, circle] (di3) at (8,2) {$i_3$};

\draw (di1) -- (j1) node[midway, right] {$w$};
\draw (di2) -- (j2) node[midway, right] {$w$};
\draw (di3) -- (j3) node[midway, right] {$w$};

\node at (6,-1) {Decomposed graph $\calG_w$};

% Connecting original and decomposed graphs
\draw [dashed, ->] (j) to [out=345,in=195] (j1);
\draw [dashed, ->] (j) to [out=345,in=195] (j2);
\draw [dashed, ->] (j) to [out=345,in=195] (j3);
\end{tikzpicture}
\caption{Decomposition of a general graph to a star graph.}\label{fig:graph_decompose}
\end{figure}

At one extreme, a weight of $w=1$ provides an aggressive approximation, in which the formulation allows each customer $j\in\calC$ to be covered $|\calF_j|$ times. Vice versa, a weight of $w=1/d_\star$ provides a conservative approximation in which each customer $j\in\calC$ can be covered at most once. 
\begin{corollary}\label{cor:heuristic}
Let $m^S(w)$ be the optimum of Problem~\eqref{prob:network_ALG} in the star graph $\calG_w$ with weights $\text{deg}(i)=w|\calC_i|$, and $m^\dagger$ the one of Problem \eqref{prob:network_ALG}. Then $m^S(1) \leq m^\dagger \leq m^S(1/d_\star)$.
\end{corollary}

In between, the weight provides a degree of freedom to control the error resulting from the star graph approximation. We select the weight $w$ via one-dimensional line search between $1/d_\star$ and 1:
\begin{equation}\label{eq:line}
\max\quad\left\{\texttt{STAR}(\calG_w)\ \mid\ 1/d_\star \leq w \leq 1\right\}\tag{$\calP_w^\dagger$}.
\end{equation}

Finally, we prune the resulting graph to further improve the solution. Let $w^*$ denote the optimal weight from Problem~\eqref{eq:line} and let $\by^\dagger(w^*)$ denote the corresponding optimal solution. We perform a greedy search to prune binary decisions $y_{it}^{\dagger}(w^*)$, by removing facilities one at a time as long as the coverage constraint in Problem~\eqref{prob:network_ALG} is satisfied. The full algorithm is detailed in Algorithm \ref{alg:network_full}. 

\begin{algorithm} [h!]
\caption{Algorithm to solve Problem~\eqref{prob:network_ALG} in a general graph $\calG$.}\small
\label{alg:network_full}
\begin{algorithmic}
\State Solve Problem~\eqref{eq:line} by line search; retrieve optimal solution $w^*$ and solution $\by^*\gets\by^\dagger(w^*)$.
\State Repeat Steps 1 -2, until breaking condition is met.
\begin{itemize}
\item[] \textbf{Step 1.} Select $i_0\in\calF$ and $t_0\in\calT$ such that $y^*_{i_0,t_0}=1$. Define candidate solution
$$\widetilde y_{it}^\dagger=\begin{cases}
    y_{it}^*&\text{for all $(i,t)\in\calF\times\calT\setminus\{(i_0,t_0)\}$}\\
    0&\text{if $i=i_0$ and $t=t_0$}
\end{cases}$$
\item[] \textbf{Step 2.} Update $y^*_{i_0,t_0}\gets0$ if the following condition is met. Otherwise, break.
$$\sum_{j=1}^q \left(1-\prod_{\tau=1}^T \left(\frac{\varepsilon\cdot p}{\left(\sum_{s=1}^{\tau-1} \sum_{i=1}^n \widetilde y_{is}^\dagger +1\right)^r}+\varepsilon\cdot(1-p)\right)^{\sum_{i \in \calF_j} \widetilde y_{i\tau}^\dagger}\right) \geq m+\Delta_G(m)$$
\end{itemize}
\item \textbf{Output:} $\by^*$
\end{algorithmic}
\end{algorithm}

\subsection{Proof of Proposition~\ref{prop:small_first}}
\label{app:small_first}

For notational ease, we focus on the case where $p=1$, but all arguments trivially to the case with $p\in[0,1]$. The proof proceeds by swapping decisions $y^\dagger_{i_1,t} \to y^\dagger_{i_1,\tau}$ and $y^\dagger_{i_2,\tau} \to y^\dagger_{i_2,t}$. This swap does not change the coverage of any customer other than those in the neighborhood of the two facilities. It also leaves the objective function of Problem~\eqref{prob:network_ALG} unchanged. Denote the new decisions as $\widetilde{\by}^\dagger$. Given that $y^\dagger_{it}$ is the optimal solution to Problem~\eqref{prob:network_ALG}, it is without loss of generality to assume that the left-hand side of the constraint is at least as large under $y^\dagger_{it}$ as under $\widetilde{y}^\dagger_{it}$, i.e.:
\[\sum_{j \in \calC_{i_1} \cup \calC_{i_2}}^q \left(1-\prod_{\tau=1}^T \left(\frac{\varepsilon}{\left(\sum_{s=1}^{\tau-1} \sum_{i=1}^n y^\dagger_{is} +1\right)^r}\right)^{\sum_{i \in \calF_j} y^\dagger_{iv}}\right)\geq \sum_{j \in \calC_{i_1} \cup \calC_{i_2}}^q \left(1-\prod_{v=1}^T \left(\frac{\varepsilon}{\left(\sum_{s=1}^{v-1} \sum_{i=1}^n \widetilde{y}^\dagger_{is} +1\right)^r}\right)^{\sum_{i \in \calF_j} \widetilde{y}^\dagger_{iv}}\right)\]

We can simplify this expression by noting that $\sum_{i \in \calF_j} \widetilde{y}^\dagger_{iv}=\sum_{i \in \calF_j} y^\dagger_{iv}$ for all $j\in\calC\setminus(\calC_{i_1}\cup\calC_{i_2})$ and for all $v\in\calT$:
\begin{align*}
&\sum_{j \in \calC_{i_1} \setminus \calC_{i_2}}^q \prod_{\tau=1}^T \left(\frac{\varepsilon}{\left(\sum_{s=1}^{\tau-1} \sum_{i=1}^n y^\dagger_{is} +1\right)^r}\right)^{\sum_{i \in \calF_j} y^\dagger_{iv}-\mathbf{1}\{v=t\}} \left(\left(\frac{\varepsilon}{\left(\sum_{s=1}^{t-1} \sum_{i=1}^n y^\dagger_{is} +1\right)^r}\right)-\left(\frac{\varepsilon}{\left(\sum_{s=1}^{\tau-1} \sum_{i=1}^n y^\dagger_{is} +1\right)^r}\right)\right)
\\+&\sum_{j \in \calC_{i_2} \setminus \calC_{i_1}}^q \prod_{\tau=1}^T \left(\frac{\varepsilon}{\left(\sum_{s=1}^{\tau-1} \sum_{i=1}^n y^\dagger_{is} +1\right)^r}\right)^{\sum_{i \in \calF_j} y^\dagger_{iv}-\mathbf{1}\{v=\tau\}} \left(\left(\frac{\varepsilon}{\left(\sum_{s=1}^{\tau-1} \sum_{i=1}^n y^\dagger_{is} +1\right)^r}\right)-\left(\frac{\varepsilon}{\left(\sum_{s=1}^{t-1} \sum_{i=1}^n y^\dagger_{is} +1\right)^r}\right)\right)\\
&\leq 0
\end{align*}
Further simplification results in:
\begin{align*}
&\sum_{j \in \calC_{i_1} \setminus \calC_{i_2}}^q \prod_{\tau=1}^T \left(\frac{\varepsilon}{\left(\sum_{s=1}^{\tau-1} \sum_{i=1}^n y^\dagger_{is} +1\right)^r}\right)^{\sum_{i \in \calF_j} y^\dagger_{iv}-\mathbf{1}\{v=t\}}\\
&-\sum_{j \in \calC_{i_2} \setminus \calC_{i_1}}^q \prod_{\tau=1}^T \left(\frac{\varepsilon}{\left(\sum_{s=1}^{\tau-1} \sum_{i=1}^n y^\dagger_{is} +1\right)^r}\right)^{\sum_{i \in \calF_j} y^\dagger_{iv}-\mathbf{1}\{v=\tau\}}
\leq 0
\end{align*}
We use a similar switching argument to show that those facilities that do not get selected have a smaller expected incremental coverage than those that do get selected.\hfill\Halmos

\subsection{Proof of Corollary~\ref{cor:small_first}}

Consider two facilities $i_1\neq i_2$ such that $y^\dagger_{i_1,t}=y^\dagger_{i_2, \tau}=1$, with $t<\tau$. In the case of a star graph, we know that $\calC_{i_1}\cap\calC_{i_2}=\emptyset$ and that $|\calF_j|=1$ for all $j\in\calC$. Denote by $N_{t-1}$ the number of samples available at time $t=1,\cdots,T$. Then, the condition from Proposition~\ref{prop:small_first} reduces to:
$$\sum_{j \in \calC_{i_1}}\left(\frac{\varepsilon\cdot p}{\left(N_{t-1} +1\right)^r}+\varepsilon\cdot(1-p)\right)\leq\sum_{j \in \calC_{i_2}}\left(\frac{\varepsilon\cdot p}{\left(N_{t-1} +1\right)^r}+\varepsilon\cdot(1-p)\right),$$
i.e.:
$$\left(\frac{\varepsilon\cdot p}{\left(N_{t-1} +1\right)^r}+\varepsilon\cdot(1-p)\right)\text{deg}(i_1)\leq\left(\frac{\varepsilon\cdot p}{\left(N_{t-1} +1\right)^r}+\varepsilon\cdot(1-p)\right)\text{deg}(i_1)$$
Since $t<\tau$, we know that $N_{t-1}\leq N_{\tau-1}$, which implies that $\text{deg}(i_1)\leq\text{deg}(i_2)$. This proves that facilities are tentatively opened by increasing degree order. We proceed similarly to show that those facilities that do not get selected have a smaller degree than those that do get selected.
\hfill\Halmos

\subsection{Proof of Corollary~\ref{cor:heuristic}}

With $\text{deg}(i)=|\calC_i|$, we are effectively solving the same problem as formulated as Problem~\eqref{prob:network_ALG}, except that the left-hand side is written as:
\[\sum_{j=1}^q \sum_{i \in \calF_j} \left(1-\prod_{\tau=1}^T \left(\frac{\varepsilon\cdot p }{\left(\sum_{s=1}^{\tau-1} \sum_{i=1}^n y_{is} +1\right)^r}+\varepsilon\cdot(1-p)\right)^{y_{i\tau}}\right)\]
Since $1-\prod_{\tau=1}^T \left(\frac{\varepsilon\cdot p }{\left(\sum_{s=1}^{\tau-1} \sum_{i=1}^n y_{is} +1\right)^r}+\varepsilon\cdot(1-p)\right)^{y_{i\tau}}<1$, this is larger than or equal to
\begin{align*}
\sum_{j=1}^q 1-\prod_{\tau=1}^T \left(\frac{\varepsilon\cdot p }{\left(\sum_{s=1}^{\tau-1} \sum_{i=1}^n y_{is} +1\right)^r}+\varepsilon\cdot(1-p)\right)^{\sum_{i \in \calF_j} y_{i\tau}}  
\end{align*}
Therefore, any solution that is feasible for the original problem is feasible for the modified one, with the same objective value. This implies that $m^S(1)\leq m^\dagger$.

Similarly, for $\text{deg}(i)=w|\calC_i|$ ans $w=1/d_\star$, we effectively solve the same problem as formulated as Problem~\eqref{prob:network_ALG}, except the left side is written as:
\begin{align*}
&\sum_{j=1}^q \sum_{i \in \calF_j} \frac{1}{d_\star}\left(1-\prod_{\tau=1}^T \left(\frac{\varepsilon\cdot p }{\left(\sum_{s=1}^{\tau-1} \sum_{i=1}^n y_{is} +1\right)^r}+\varepsilon\cdot(1-p)\right)^{y_{i\tau}}\right)
\\\leq &\sum_{j=1}^q \sum_{i \in \calF_j} \frac{1}{|\calF_j|}\left(1-\prod_{\tau=1}^T \left(\frac{\varepsilon\cdot p }{\left(\sum_{s=1}^{\tau-1} \sum_{i=1}^n y_{is} +1\right)^r}+\varepsilon\cdot(1-p)\right)^{y_{i\tau}}\right)
 \\ =&q-\sum_{i \in \calF_j} \frac{1}{|\calF_j|}\prod_{\tau=1}^T \left(\frac{\varepsilon\cdot p }{\left(\sum_{s=1}^{\tau-1} \sum_{i=1}^n y_{is} +1\right)^r}+\varepsilon\cdot(1-p)\right)^{y_{i\tau}}
  \\ \leq &q-\prod_{\tau=1}^T \left(\frac{\varepsilon\cdot p }{\left(\sum_{s=1}^{\tau-1} \sum_{i=1}^n y_{is} +1\right)^r}+\varepsilon\cdot(1-p)\right)^{\sum_{i \in \calF_j} y_{i\tau}}    
\end{align*}
The last inequality follows from the fact that, for all $b>0$ and $x_i>0$ we have per the AM-GM inequality:
\[\frac{1}{k} \sum_{i=1}^k b^{x_i} \geq b^{\frac{1}{k}\sum_{i=1}^k x_i},\]
hence, for all $0<b\leq 1$ and any $x_i>0$,
\[\frac{1}{k} \sum_{i=1}^k b^{x_i} \geq b^{\sum_{i=1}^k x_i},\]
Thus, any feasible solution for $w=1/d^\star$ is feasible for the original problem with the same objective value. Thus $m(1/d^\star)\geq m^\dagger$.
\end{document}